# A topological solution to object segmentation and tracking

Thomas Tsao[1] and Doris Y. Tsao[2, 3, *]

1. OpticArray Technologies, 1917 Dundee Road, Rockville, MD 20850
2. Division of Biology and Biological Engineering, Caltech, Pasadena CA 91125
3. Howard Hughes Medical Institute, Pasadena, CA 91125

**\*Corresponding author:**
Doris Y. Tsao
Email: tsao.doris@gmail.com

**Abstract**
The world is composed of objects, the ground, and the sky. Visual perception of objects requires solving two fundamental challenges: segmenting visual input into discrete units, and tracking identities of these units despite appearance changes due to object deformation, changing perspective, and dynamic occlusion. Current computer vision approaches to segmentation and tracking that approach human performance all require learning, raising the question: can objects be segmented and tracked without learning? Here, we show that the mathematical structure of light rays reflected from environment surfaces yields a natural representation of persistent surfaces, and this surface representation provides a solution to both the segmentation and tracking problems. We describe how to generate this surface representation from continuous visual input, and demonstrate that our approach can segment and invariantly track objects in cluttered synthetic video despite severe appearance changes, without requiring learning.

**Introduction**
Through a process of perceptual organization that is still not well understood, the primate visual system transforms visual input consisting of a stream of retinal images into a percept of stable, discrete objects. This process has traditionally been broken down into two separate problems: the "segmentation problem," which addresses how visual pixels can be grouped into distinct objects within a single image (1), and the "tracking problem," which addresses how objects can be identified across images despite changing appearance (2).

Both problems are highly challenging. Segmentation is difficult because distant pixels of different color/texture can belong to the same object while neighboring pixels of the same color/texture can belong to different objects (**Fig. 1A**). Tracking is difficult because the appearance of the same object can change drastically due to object deformation, changing perspective, or dynamic occlusion (**Fig. 1B**). The segmentation problem has classically been tackled through intensity-, color-, and texture-based region-growing approaches relying upon properties extracted from single images (3), and more recently through deep learning approaches. The tracking problem has been approached through probabilistic dynamical modeling (4) or "tracking by detection" (5-8), with recent methods incorporating deep learning (9-15). While earlier learning approaches to segmentation and tracking were supervised (10, 16), requiring large labeled training sets, more recently unsupervised approaches have emerged (15, 17). In this paper we explore the *computational origin* of the ability to segment and invariantly track objects and show that this problem can in principle be solved without learning, supervised or unsupervised.

Complementing these image-based approaches to segmentation and tracking, a geometry-based approach considers vision as an inverse graphics problem (18). In this framework, the visual system infers 3D surfaces from images by inverting a 3D graphics model. However, because the third dimension is lost during perspective projection onto the retina, this inverse inference process is considered to be not fully constrained (19), implying that extensive learning from experience is necessary. In this paper, we show that the problem of inferring 3D surfaces from images is in fact fully constrained, if the input is in the form of a sequence of images of a scene in which either the observer or objects are moving. We demonstrate through both mathematical analysis and computational experiments that with only two natural assumptions, namely, i) the world is composed of objects possessing smooth, textured surfaces with locally constant lighting, and ii) animals view the world from a moving observation point, it is possible to solve the problem of segmenting and invariantly tracking each discrete surface in the environment without requiring learning. Our computational experiments are limited to synthetic video and we assume access to high-quality images, but as we argue below, our approach should be readily extendable to natural conditions.

Our paper is essentially a mathematical translation of the "ecological approach to visual perception" developed by the psychologist J.J. Gibson (20). Gibson pointed out that the key to understanding human vision is to insert between the 3D environment and the eye a new item, the *field of ambient optic arrays*. The ambient optic array at one point in space consists of the 2D distribution of light rays passing the point from illuminated surfaces in the environment (**Fig. 1C**). Gibson pointed out that the field of ambient optic arrays is governed by a set of laws which he dubbed "ecological optics", and these laws can explain much of visual perception: "*Instead of making the nervous system carry the whole burden of explaining perception, I wish to assign part of this burden to light itself. Ecological optics is my way of doing so*" (20). In the decades since Gibson proposed his ecological optics approach to vision, this important concept has remained in the language of psychology, without mathematical elaboration.

We explain how Gibson's theory can be formulated in precise mathematical terms and be implemented computationally. Mathematical analysis shows that object surface information is redundantly represented by the field of ambient optic arrays through two of its topological structures: the pseudogroup of stereo diffeomorphisms and the set of infinitesimal accretion borders. Formulated in terms of ecological optics, vision is a fully constrained, well-posed problem. Complete information for perception of objects as discrete, persistent units is contained in the visual environment itself within the field of ambient optic arrays.

The paper has three main parts. In the first part we give a broad overview of our approach. In the second part we present the mathematical theory of ecological optics (this part heavily references the Supplementary Material and may be skipped without loss of comprehension to the remainder of the paper). In the third part we show how to exploit ecological optics computationally to solve the segmentation and invariant tracking problems.

**I. Surface representation: overview**

Unlike taste and touch, vision allows an animal to experience the environment without immediate contact. In vision, the link between the distal stimulus (objects in the environment) and proximal stimulus (light impinging on the retina) is the light reflected from environment surfaces, which at each point of observation forms what Gibson called the "optic array." We will prove in the next section that information sufficient to both segment and track surfaces is faithfully represented in the field of optic arrays by transformations between visual images across a sequence of observation points. We will then demonstrate in section three how to compute these transformations and use them to perform object segmentation and tracking. While understanding the proofs requires a basic understanding of differential topology, the essential ideas, which we summarize in this section, are highly intuitive.

Given a complex scene containing multiple objects (**Fig. 2A**), the goal of segmentation is to identify object boundaries. An efficient way to approach this is to start with a map of all the edges in the image (**Fig. 2B**), since object boundaries should be a subset of these edges. The key difficulty is that some edges are "texture edges" (e.g., the sticker in **Fig. 2A**), while others are true object edges (e.g., the edge of the apple in **Fig. 2A**). We prove that information in the transformation between nearby perspectives of a scene can be used to distinguish these two types of edges. Specifically, *if a region of space contains a patch of surface*, then two image patches taken from nearby observation points will be *diffeomorphic* to each other, i.e., *one can register them by stretching and warping like a rubber sheet* (**Fig. 2C**). Furthermore, we show how to compute this diffeomorphism computationally through an iterative optimization scheme in which a set of local Gabor receptive fields dynamically undergo affine transformation to cancel the transformation between the two image patches (see **Fig. 5** below). However, *if an image patch contains an object edge* (e.g., the edge of the apple), then on one side of the edge the image patches will be diffeomorphic, but on the other side they will not, because there will be a piece of the background visible from one perspective but not the other, leading to a one-sided breakdown in diffeomorphism (**Fig. 2D**). In visual psychophysics, this phenomenon has been referred to as "da Vinci stereopsis" (21). This provides an effective way to distinguish texture edges from true object borders: for each edge element, determine the existence of diffeomorphism on each side of the edge. Object borders are accompanied by diffeomorphism on only one side. Moreover, we can identify this as the side that owns the edge (**Fig. 2E**). By repeating this process across the entire image, we can convert an edge map into a truly informative map of object borders (**Fig. 2F**).

Once segmentation has been framed in this surface representation framework, the solution to the invariant tracking problem, which has been considered one of the hardest problems in vision (22, 23), becomes almost trivial. How can we know if two discrete patches (e.g., the two patches shown in **Fig. 2G**, or the front and back views of a horse) belong to the same invariant surface? Our insight is that we can determine this if the two patches are not separated by any object edge, i.e., if they are connected through a series of overlapping surface patches (**Fig. 2H, I**). Thus, in the surface representation framework, an invariant object constitutes an equivalence class of surface patches, where the equivalence relation is defined by surface overlap. Importantly, the same diffeomorphism machinery for solving segmentation also allows us to compute these surface overlaps, and thus to connect (i.e., track) different views of the same surface over time. Even if a surface undergoes a drastic transformation in appearance (e.g., the front and back views of a horse), as long as successive views are related by local diffeomorphisms, then the tracking process can readily link the views.

## II. Surface representation: mathematical theory

In this section, we express the laws of ecological optics mathematically. We show that the data for solving the segmentation and invariance problems, and more generally, for obtaining a representation of visual surfaces, is sufficiently and redundantly available in the animal's proximal visual environment. We formulate the problems of segmentation and invariance as follows: is it possible to determine whether two image patches (seen simultaneously through stereopsis or from different observation points) belong to the same physical surface? For the case of a single view, this corresponds to the segmentation problem; for the case of a continuous series of views over time, this corresponds to the tracking problem. Our solution to these problems, already summarized in the previous section, is categorically different from those proposed previously. It relies on a key property, surface contiguity, that is *topological* and not *image-based*, and computed from pairs of images taken from different perspectives rather than from single images.

We introduce two topological spaces: one for describing the 3D objects in the environment (the distal stimulus), and one for describing the light rays reflected from these objects and converging at each observation point in the environment (the proximal stimulus). We study the mapping between these two spaces and prove that information about the topological organization of objects in the former space is faithfully represented in the latter space. In other words, we prove that visual perception of invariant objects is possible.

Specifically, we prove that the property of local surface contiguity is specified by the existence/non-existence of a particular type of mapping between pairs of images taken from different perspectives, namely, "stereo diffeomorphism"; this provides the key to topological image segmentation (**Fig. 3A-C**). We further prove that two surface representations are of the same object if they each contain a part related by stereo diffeomorphism; this global topological property provides the key to invariance (**Fig. 3D**).

*The geometry of light rays in an environment containing objects and multiple observation points.* Let $O$ be a potential observation point in the medium. We call the set $S(O)$ of all rays starting at the point $O$ the *ray space* based at $O$ (**Fig. 3A**). The space containing all rays with their base point located in a domain of the observation space $\Omega$ (i.e., set of all potential observation points) is the space $VS(\Omega) = S \times \Omega$. We call $VS(\Omega)$ the *visual space* on the *observation domain* $\Omega$. The space spanned by two ray spaces $S(O)$ and $S(O')$ represents all possible pairings of rays taken from $S(O)$ with rays taken from $S(O')$. We call $S(O) \times S(O')$ the transition space based at $(O, O')$.

We have two types of topological spaces: the 3D Euclidean space of ordinary points for describing the spatial structure of the objects and their surfaces; and the ray spaces, transition spaces, and visual space for describing the spatial structure of light rays converging on every possible observation point. There are mapping relations between "points" of the two different types of topological spaces.

We use the term "environment" to refer to all the surfaces, the ordinary surfaces of 3D objects and the ground, and the sky which is considered a surface with each point at an infinite distance. A further mathematical assumption is that environment surfaces are piecewise smooth. The mapping from a point in the 3D Euclidean space to a ray space is given by point projection: let $P$ be a point in the 3D Euclidean space, $O$ a point in a domain of observation, $P \neq O$. We call the ray $r \in S(O)$ the image of $P$ if $P$ is a point on $r$ (**Fig. 3A**). A point of the visual environment is *visible from a point* of observation if the line segment connecting these two points does not intersect any other points on an ordinary surface. We assume brightness constancy, i.e., different rays coming from the same point have the same intensity.

A *perspective projection* from a surface to a ray space is a map generated by applying point projection to every visible point on the surface. The analytical structure of perspective projection from a general 2D manifold to another 2D manifold is the subject matter of differential topology. In particular, according to a theorem by Whitney (24), upon perspective projection of environment surfaces to ray spaces, the points in each ray space are divided into two sets: the set of *regular* values, where the perspective projection is one-to-one continuous and differentiable, and the set of *critical* values occurring at the boundaries of regular domains, where this relation breaks down. This insight provides a means to compute surface contiguity and separation from information available in the proximal ray spaces.

In a rigid environment, the perspective changes with change of the point of observation (achieved by having two eyes or by physically displacing one eye). We call a pair of perspectives a *perspective transition*. We call the image of 3D Euclidean space in the 4D transition space the *stereo space*. Each of its elements is called a *stereo pair*. The stereo space constitutes the subset of ray pairs in the transition space that intersect in a point in 3D Euclidean space (**Fig. 3B**). A *stereo diffeomorphism* is a diffeomorphism between domains (*stereo neighborhoods*) of two ray spaces, such that each ray and its image form a stereo pair.

*Coding local surface contiguity.* Let $S$ be a patch of surface in the environment visible from two observation points $O_1$ and $O_2$ (**Fig. 3B**). We prove that the images of $S$ under perspective projection in $S(O_1)$ and $S(O_2)$ are related by *stereo diffeomorphism* $g$ (Existence of Stereo Diffeomorphism, Supplementary Material); conversely, given a stereo diffeomorphism $g$ from a domain $S_1$ in one ray space to $S_2$ in another ray space, the stereo pairs in the transition space satisfying the constraint $g$ specify a 2D manifold in the 3D stereo space (Surfaciness Theorem, Supplementary Material).

We call $h=(g, S_1, S_2)$ a *mapping triple*. Both $S_1$ and $S_2$ are diffeomorphic to the 2D surface patch specified by $h$, and therefore each qualifies as a topological representation of this patch. The mapping triple specifies not only the existence of a contiguous surface patch, but also its metric properties of distance and curvature (Shape from Perspective Mapping Theorem, Supplementary Material).

*Coding surface spatial separation.* In the previous section, we showed how local surface contiguity is encoded in pairs of ray spaces by mapping triples. Points in domains of mapping triples are regular values of perspective projection (Local Stability of the Regular Value Set Theorem, Supplementary Material). A perspective projection can also have critical values, and these turn out to be the key to encoding surface spatial separation.

Whitney proved that there are only two types of singularities of a smooth mapping from a 2D manifold to a 2D manifold: *folds* and *cusps* (24) (see **Fig. S10**). We call images of fold singularities in a ray space under perspective projection *occluding contours* (**Fig. 3C**, vertical magenta segment).

Occluding contours carry rich information about surface spatial separation and continuation. First, we prove that the two sides of an occluding contour represent spatially *separated* local surfaces; second, each of the local surfaces continues in a particular manner, with the owner side folding back, and the non–owner side extending behind the surface of the owner side (Separated Surface Continuation Theorem, Supplementary Material). The brain appears to use this information effectively: Gestalt psychologists observed that the presence of occluding contours can remove the boundary of a figure and make it "incomplete" and thus trigger a process of "amodal completion" behind the occluding contour (25).

Occluding contours are defined as singularities of the mapping from the 3D Euclidean space to the visual space. But the visual system only has access to data in the visual space. What information in the visual space is available to detect occluding contours? The key insight is that an occluding contour is a border of infinitesimal accretion: there always exists a small domain of observation, such that the occluding contour is the border of accretion of a perspective transition within this domain (**Fig. 3C,** vertical green region; Border of Accretion Criteria for Occluding Contours Theorem, Supplementary Material). Furthermore, the owner side can be computed as the side opposite that which undergoes perceivable accretion, while it itself remains topologically invariant, i.e., is only subject to a diffeomorphic transformation. Note that for a smooth surface like a sphere, there can be accretion on both sides. However, we can prove for shift of observation point of magnitude $\varepsilon$ (where $\varepsilon$ is small enough that $\varepsilon^2 \ll \varepsilon$), the width of the accretion on the owner-side goes as $\varepsilon^2$, while the width of the accretion on the background goes as $\varepsilon$ (Sided Division of Accretion Track Theorem, Supplementary Material). Because the amount of accretion on the owner side is so much smaller, it is easy to differentiate the two sides computationally.

*Coding global surfaces in a single perspective: ad hoc surface representation.* Under perspective projection, a tuple $T = (C, B)$ of a regular component (i.e., maximal connected set) $C$ and its surrounding occluding contours $B$ gives a representation of the global surface at a point of observation in the following sense: component $C$ represents a visible part of the global surface, and $B$ represents the rest of the global surface. We call such a tuple an *ad hoc representation* of the global surface.

*Coding global invariant surfaces across perspectives.* Once the machinery for generating an ad hoc surface representation in a single perspective through occluding contours is in place, extracting a globally invariant surface representation (i.e., a representation in which the same surface is identified across perspectives) is essentially trivial: the local contiguous surface components in each ad hoc representation can be simply stitched together through partial overlaps.

How can we identify representations of the same surface across perspectives? First we define an equivalence relation among domains of regular values in the visual space: domains $S_1$ and $S_2$ are equivalent if there is a perspective mapping triple $(g, S_1, S_2)$ for domains $S_1$ and $S_2$ in the ray spaces of some pair of observation points. This equivalence relation divides the whole set of domains of regular values in the visual space defined on an observation domain $\Omega$ into different equivalence classes, each called an $MS(\Omega)$-equivalence class, where $MS(\Omega)$, the *mapping structure* on $\Omega$, is the total set of perspective mapping triples $(g, S_1, S_2)$. The mapping structure forms a pseudogroup (26) on the visual space and provides the conceptual foundation for understanding visual invariance. Each of these equivalence classes represents a local surface patch of the environment invariant to perspective.

From these $MS(\Omega)$-equivalence classes, we construct a perspective-invariant representation of a global surface as follows: if a pair of domains from two different $MS(\Omega)$-equivalence classes have non-void intersection, we call the two $MS(\Omega)$-equivalence classes *connected*. We call two $MS(\Omega)$-equivalence classes *chain connected* if there is a chain of consecutively connected domains linking these two classes. Chain connectedness defines an equivalence relation. Each chain-connected $MS(\Omega)$-equivalence class, denoted a $CC(\Omega)$-equivalence class, represents the perspective-invariant global surface of a 3D object (Fixed Owner Theorem, Supplementary Material).

Finally, we are ready to answer the question: how can ecological optics represent an invariant global surface across different perspectives? In each perspective, an ad hoc representation of the global surface is available if it is partially visible. Let $T_1 = (C_1, B_1)$ and $T_2 = (C_2, B_2)$ be two ad hoc representations in a perspective transition from observation point $O_1$ to observation point $O_2$. These two representations are perceived as encoding the same global surface if $C_1$ and $C_2$ are $CC$-equivalent (**Fig. 3D**).

To summarize, we set out to understand whether it is possible to determine that two patches $S_1$, $S_2$ of ambient optic arrays in ray spaces at different observation points $O_1$, $O_2$ are perspective images of the same physical surface or not. We first sketched a mathematical framework: light from object surfaces is mapped to ray spaces at each observation point in the environment through perspective projection, defining a mapping from the distal Euclidean space to the proximal visual space $VS(\Omega)$. We then searched for a stereo diffeomorphism between $S_1$ at $O_1$ and an image patch in a nearby observation point, and likewise for $S_2$ at $O_2$. The existence of these stereo diffeomorphisms means that $S_1$ and $S_2$ each represent some local surface patch (not necessarily belonging to the same global surface). Next, at both $O_1$ and $O_2$, we extend these local surface patches to ad hoc representations of global surfaces by identifying the occluding contours bounding $S_1$ at $O_1$, and $S_2$ at $O_2$. Finally, we determine whether these ad hoc representations are of the same global surface by testing for chain connectedness between $S_1$ and $S_2$.

Thus the laws governing the optical projection of the visual environment to the visual space give rise to a topological representation of persistent environment surfaces in terms of (1) the ad hoc representation in each perspective defined by the set of regular and critical values, and (2) the invariants across perspectives defined by equivalence relations given by the pseudogroup of stereo diffeomorphisms and chain connectivity. This persistent surface representation sets the stage upon which object perception functions.

We note that in computer vision, Koenderink and van Doorn were the first to try to explore the singularity structure of images for the purpose of understanding invariant perception of objects (27), but their goals were very different from ours. They observed that i) self-induced movements of an observer generate motion-parallax fields, ii) the singularities of these fields correspond to folds and cusps (following Whitney (24)), which are stable for most vantage points and provide information about invariant object shape, and iii) at unstable points, these singularities can change in a specific number of possible ways to reveal new shapes (e.g., a hill transforming into a hyperbolic intrusion). Their main focus was extracting information about *solid shape,* while our main focus is segmentation and object tracking *independent of shape*; in neuroscience terms, this can be considered a distinction between "what" and "where" stream functions. Critically, in our theory, the invariance of surfaces is based upon the equivalence relation of partial overlap, not on "stability of singularities."

### III. Surface representation: algorithmic implementation and computational experiment

So far, we have presented a theory of ecological optics. In the same way that geometric optics describes how points on an object are carried by light to points in the image plane, ecological optics describes how topologically important structures of object surfaces in 3D Euclidean space (i.e., properties such as contiguity, spatial separation, partial overlap, etc.) are carried by light to topological structures of rays in the visual space: regular components, perspective mappings, occluding contours, accretion/deletions around occluding contours, $MS(\Omega)$-equivalence classes, and $CC(\Omega)$-equivalence classes. The theory of ecological optics presented in the previous section describes the physical reality of the animal's visual environment and does not depend in any way on the presence of a visual system. In this section, we demonstrate how a visual system that moves through the environment can computationally exploit the topological structures of rays in the visual space to perceive the topology of the visual environment, i.e., to perceive discrete, invariant units.

*Algorithmic method for segmentation and invariant object tracking.* Given a sequence of video frames of a scene in which either the observer or objects are moving, our goal is to segment each frame according to surface contiguity and assign the same label to surface components corresponding to the same object across frames.

We first find intensity edges using a standard edge detection algorithm, e.g., the Canny edge detector (28) (**Fig. 4A**); here, we are assuming that in natural viewing conditions occluding contours are mostly associated with intensity edges. This assumption is due to the fact that images of borders between spatially separated surfaces likely have different intensity. We then randomly select a set of neighborhoods of the identified edges for further topological analysis. Importantly, these neighborhoods are taken in pairs from successive frames (**Fig. 4B**). The next, crucial step is to classify edge segments as texture edges or occluding edges, based on diffeomorphism detection between successive frames performed separately on each side of the segment (**Fig. 4B-D**), and then to identify the owner of each occluding edge. Following the mathematical theory, at texture edges, diffeomorphisms computed on either side are the same, while at object edges, the neighborhood on the side that owns the edge is diffeomorphic to its counterpart in the next frame, but the neighborhood on the opposite side is not due to accretion/deletion. The specific method we use to determine existence/non-existence of diffeomorphism is described in detail in the next section as well as in Section 4 of the Supplementary Material and in (29, 30).

Once texture edges have been distinguished from object edges, owners of object edges have been identified, and diffeomorphisms have been computed between successive frames at each neighborhood, we are then ready to perform object segmentation and tracking. We start by computing a "super segmentation" map that assigns a different label to each contour-bounded component (31) (**Fig. 4D**, left). Then, to compute the segmentation map, we simply erase texture edges by re-assigning the label of any pure texture region (i.e., a region that abuts a texture edge but is never a one-sided owner) to that of its two-sided partner (**Fig. 4D**, middle). Finally, once segmentation is complete, the last step of computing the object tracking map becomes trivial: we determine *persistent surfaces*--components of the object segmentation map containing a patch diffeomorphic to a one-sided owner or texture patch from the previous frame (**Fig. 4D**, middle)-- and assign each persistent surface the same label as that in the previous frame (**Fig. 4D**, right). Note that here we are re-using the diffeomorphism detections performed during the segmentation stage.

In broad terms, the steps for scene segmentation and tracking just presented can be organized into three major groups of steps: i) edge extraction and computation of a super segmentation map, ii) computation of correspondence, and iii) relabeling of components of the super segmentation map using correspondence information. Below, we elaborate on the key computational workhorse in this scheme, detection of diffeomorphisms.

*Extraction of diffeomorphisms in a perspective transition.* Distinguishing texture from object edges requires determination of existence/non-existence of diffeomorphism. A diffeomorphism $g$ can locally be approximated by its first order Taylor expansion, which is a shift of the center point of the domain and a linear correction term for points in its vicinity, an affine transformation with six parameters. If the diffeomorphism is a stereo diffeomorphism (i.e., a diffeomorphism arising from viewing the same rigid object from two perspectives), then the transformation is constrained to only three parameters (Rigid Shape from Perspective Mappings, Supplementary Material). Let $p$ be a ray from the ray space $S(O)$ and $p'=g(p)$ its conjugate pair in $S(O')$. Let $U(p)$ and $U'(p')$ denote the set of all rays in local neighborhoods of $p$ and $p'$, respectively, and $f_O$ and $f_{O'}$ functions that map each ray at $O$ and $O'$, respectively, to a brightness value. The image patches taken at two locations $O$ and $O'$, $f_O(U(p))$ and $f_{O'}(U'(p'))$, are said to be $g$-related if

$$p' = g(p), U'(p') = g(U(p)) = g \circ U(p), \tag{1}$$

and

$$f_O(U(p)) = f_{O'}(g \circ U(p)). \tag{2}$$

Our goal is to compute, for any two image patches $f_O$ and $f_{O'}$, whether there exists a 6-parameter affine transform $g$ that satisfies (2). If so, then we conclude that the two image patches are related by diffeomorphism $g$. Equation (1) expresses the fact that light rays projected from a surface patch to ray spaces at two observation points are related by diffeomorphism (**Fig. 3B**), while equation (2) expresses the brightness constancy constraint, namely, that the brightness of every light ray originating from the same point is the same (we only need this to hold locally).

Our general approach is as follows: we project both image patches onto a set of Gabor receptive fields of varying orientation and spatial frequency. Importantly, we make these receptive fields *dynamic,* such that they can undergo affine transforms. We then set up an energy minimization process to find the affine transform of the receptive field that exactly cancels the affine transform of the image patch. If we succeed, then we conclude the two image patches are related by diffeomorphism, and the affine approximation to this diffeomorphism is given by the parameters of the identified receptive field transform. For example, if image patch 2 is shifted relative to image patch 1, then our energy minimization process identifies the precise amount of shift in receptive field such that

$$\langle receptive\ field, image\ patch\ 1 \rangle = \langle shifted\ receptive\ field, image\ patch\ 2 \rangle.$$

Formally, the real-valued function $f_O(U(p))$ on the image plane (i.e., the image patch) can be thought of as a vector in an infinite dimensional Hilbert space (i.e., a complete space with an inner-product), and equation (2) is an abbreviation of an infinite system of equations. Given image patch $f_{O'}(U(p))$, the trajectory of $f_{O'}(g \circ U(p))$ in the Hilbert space of images on the second image plane under the affine Lie transformation group is a 6D sub-manifold in the Hilbert space. The power of considering an image patch as a vector in Hilbert space is that we can then represent an affine transform of the image patch as a *conjugate affine transform* in a dual vector space of differentiable receptive field functions. This allows us to use a gradient-based optimization approach to identify the conjugate affine transform that exactly cancels the affine transform of the image patch.

Projecting $f_O(U(p))$ on a subspace spanned by $n$ differently-oriented Gabor "receptive field" functions $F_i(U)$, $i = 1, 2, ..., n,$ gives smoothed and band-pass filtered vector-valued signals:

$\gamma_O^i(p) = \langle F_i(U), f_O(U(p)) \rangle$, $i = 1, 2, ..., n$.

Equation (2) implies

$$\gamma_O^i(p) = \langle F_i(U), f_O(U(p)) \rangle = \langle F_i(U), f_{O'}(g \circ U(p)) \rangle, \ i = 1,2,\dots,n. \tag{3}$$

The $n$-tuple $\vec{\gamma}_O(p) = (\gamma_O^1(p), \gamma_O^2(p), \dots, \gamma_O^n(p))$ in the n-dimensional signal space $R^n$ is called a *Gabor place token*. Notice the pullback via mapping $g: U \to U'$ and $g: (x, y) \mapsto (x', y')$, $T(g): f' \mapsto f, f \in L^2, f' \in L^2$, where $(T(g) \circ f_{O'})(U(p)) = f_{O'}(g \circ U(p))$ is a linear transformation on the Hilbert space of $L^2$ functions. Let $T^*(g)$ be the *conjugate* of the Hilbert space transformation of $T(g)$ with respect to the $L^2$ *inner* product (see Methods), from equation (3) we have:

$\langle F_i(U), f_O(U(p)) \rangle = \langle F_i(U), (T(g) \circ f_{O'})(U(p)) \rangle, \ i = 1,2,\dots,n.$

$\langle F_i(U), f_O(U(p)) \rangle = \langle (T^*(g) \circ F_i)(U), f_{O'}(U(p)) \rangle, \ i = 1,2,\dots,n,$

Let $g^*$ be the image domain affine transformation with $T^*(g)$ as its pullback image transformation

$$\langle F_i(U), f_O(U(p))\rangle = \langle F_i(g^* \circ U), f_{O'}(U(p))\rangle, \ i = 1,2,\ldots,n. \tag{4}$$

Let $g = g(\vec{a})$ be an affine transformation of six parameters $\vec{a} = (a_{11}, a_{12}, a_{21}, a_{22}, t_x, t_y)$ at location $p$, and define

$$\gamma_{O'}^i(p,\vec{a}) = \langle F_i(g^* \circ U), f_{O'}(U(p))\rangle, \ i = 1,2,\ldots,n.$$

Equation (4) implies that the Gabor-place token at point $p$ is invariant to affine transformation in the sense that the Gabor place token extracted by *conjugate affine-distorted* Gabor receptive fields from the *affine distorted-image* at a place on an image, $\gamma_{O'}^i(p,\vec{a})$, equals the Gabor place token at the same place, $\gamma_O^i(p)$. Thus we can define an energy function

$$E(\vec{a}, O, O') = ||\vec{\gamma}_{O'}(p,\vec{a}) - \vec{\gamma}_O(p)||^2.$$

Since this energy function is an analytical function of the affine parameters $\vec{a} = (a_{11}, a_{12}, a_{21}, a_{22}, t_x, t_y)$ defining $g$ in equation (2), we can solve for $\vec{a}$ using a gradient dynamical system (**Fig. 5A**).

Images taken from different views are almost always subject to compounded distortion involving rotation, scale, and skew, and our method is the only one we know of that can handle such compounded distortion in a principled manner; in contrast, the popular SIFT (scale invariant feature transform) approach of Lowe (32, 33) can only handle scale and orientation changes. Other correspondence methods such as FlowNet (34) generate a dense optic flow map at each point and do not directly inform about the existence of a diffeomorphism within a local neighborhood.

Equipped with this dynamic receptive field method for extracting diffeomorphisms, we can readily distinguish texture from occluding edges and identify the owners of the latter, using the fact that the six affine parameters extracted from computing correspondence between the left and right sides of a texture edge are identical (**Fig. 5B**, left), while those extracted from computing correspondence between the left and right sides of an occluding edge are different (**Fig. 5B**, right). Moreover, we can readily identify the owner of the occluding edge by determining which side is diffeomorphic to its counterpart in the next frame (**Fig. 5C**, right). Overall, our method for diffeomorphism detection provides a principled way to compute the key signal necessary for topological segmentation and tracking, surface correspondence. From these local correspondence signals, surfaces can then be stitched together across space and time to endow the visual world with global, symbolic structure.

*Results on synthetic video containing severe appearance changes due to object deformation, changing perspective, and dynamic occlusion.* To test our system, we generated a video sequence consisting of 160 frames of a dynamic scene with four objects. The objects underwent severe deformation, perspective change, and partial occlusion, and furthermore, each contained an internal texture contour to challenge the segmentation process (see https://youtu.be/eu_aJNo3R5I for a movie of the stimulus sequence). **Fig. 6A** shows the results of our topological approach applied to this dataset: we are readily able to segment and track the four objects despite drastic appearance changes.

Following a feedforward sweep across all frames, we obtain a complete *scene graph* whose vertices comprise super segmentation components across space/time and whose edges correspond to connectedness between these surface components across space/time (see also (35)). The distinct components of this scene graph correspond to distinct invariant objects (**Fig. 6B**). Equipped with this scene graph, we can then re-traverse the frames and assign the same label to each surface in the segmentation map that belongs to the same connected component in the scene graph. This allows distinct surface components to be identified as part of the same object across splits and joins over time (**Fig. 6A**, bottom row).

We underscore the severity of appearance changes that our method can handle (**Fig. 6B**, bottom), due to the fact that information for grouping is not tied to image features within single frames, but rather to topological relationships between successive frames. Moreover, the approach is robust because information for surface representation is *redundantly available on a massive scale*: i) the number of distinct objects in the environment is much less than the number of local neighborhoods available for diffeomorphism detection, and ii) most objects persist over time. Assuming the occluding contour of a typical object runs through 100 neighborhoods, and persists through 100 stimulus frames (i.e., 1 s for a 0.1 kHz visual system), this would generate 10,000 independent diffeomorphism measurements; in our simulation, we found that ~5 diffeomorphism measurements were sufficient to correctly segment a surface component. Thus, even though our demonstration was for synthetic video of textured surfaces observed without noise, it is not implausible that the approach could be adapted to natural video where these assumptions no longer hold.

In our computational test, in which 100 neighborhoods/frame were sampled across 160 frames, the segmentation process made a total of 13 mistakes (representing an error rate of 2%, since if one simply copied the super segmentation map as the segmentation map, this would result in 640 errors). **Fig. S1** presents a detailed analysis of one such mistake. An internal texture component was incorrectly segmented as a separate unit in one frame (**Fig. S1A, B**), but was nevertheless correctly tracked due to correct surface connectivity information across frames (**Fig. S1A, C**). Indeed, in the same way, all 13 segmentation mistakes vanished after object tracking. This illustrates how redundancy in information for surface representation leads to robustness.

We conclude this section on computational results with a simple demonstration of how our topological surface representation mechanism could significantly augment the capabilities of current deep neural networks trained to classify objects. Such networks rely heavily on texture (36) and can be fooled by small amounts of strategically-placed noise imperceptible to humans (37). Furthermore, they are highly sensitive to training distribution (38). Indeed, if we take four images corresponding to different stages of tracking, which each carry different color/texture information, and present them to various deep networks trained to classify images, we get three or four different answers (**Fig. 6C**; to the human eye, it is evident all four frames contain a bear with varying amounts of occlusion and varying surface texture); only one, in response to segmented and tracked input, is reliably correct. Thus, through topological segmentation and tracking, we can transform cluttered visual input that is unrecognizable to a classification-trained deep network (images 1-3 in **Fig. 6C**) into a representation of object surfaces that is readily recognizable (image 4 in **Fig. 6C**).

**Discussion**
The essential conceptual advance of this article is to show how generation of a visual surface representation turns the problem of segmentation and invariance from an ill-posed challenge, requiring ad hoc tricks or black box deep learning, to a readily solvable problem. The world is composed of objects possessing smooth textured surfaces, and animals view the world from a moving observation point. With only these two natural assumptions, we proved it is possible to solve the problem of segmenting and invariantly tracking each discrete surface in the environment. Our theory explains how a surface representation, i.e., a topological labeling of contiguous surface components together with a geometric description of their shapes and positions, can be extracted from perspective projections of the environment in a manner that is invariant to changing perspective and occlusion. We prove that segmentation of an image into separate surfaces can be accomplished through detection of occluding contours (which carry information about spatial separation of visible surfaces), and tracking of invariant surfaces in an image sequence can be accomplished by detection of diffeomorphisms (which carry information about overlap relations between surfaces visible from different views). Furthermore, we not only prove the validity of our approach mathematically, but demonstrate its computational efficacy for object segmentation and invariant tracking of synthetic video.

It is a common belief that in an image there is in reality no occlusion, no surface, no contour, but only an assemblage of pixels, and the goal of perception is to "interpret" these sensory data. Our work shows how the visual system can perceive topological structures (occlusions, surfaces, contours, etc.) in a true and original sense. The perception of these topological structures does not require observer-dependent interpretations, but can result from extraction of information directly specifying these topological objects and their relations in a rigorous mathematical sense. To achieve this, it is necessary to regard perspective projection from a new angle. Perspective projection is generally considered as a mapping from a point in 3D space to a point in the image plane. However, to understand segmentation and invariant tracking of real, curved objects, we show that it is essential: (1) to regard perspective projection as a mapping from a 2D surface of an object to a 2D ray space, and (2) to further enlarge the focus, from how a 2D surface is projected to a single ray space, to how it is projected to a *field* of ray spaces. This mathematical construction enables us to use differential topology to reach statements about surfaces as global entities: perspective projections are now 2D to 2D diffeomorphisms on regular domains, which are separated by critical points that take the form of fold contours, and these critical points are encoded by accretion/deletions in mappings between neighboring ray spaces. Without this construction, we can only speak of points.

Our theory was presaged by Gibson's theory of surface perception (20). Gibson observed that surface contiguity is specified by order-preserving transformations ("*the available information in the optic array for continuity could be described as the preservation of adjacent order*"), and related occluding contours to accretion/deletion events ("*It is called a transition, not a transformation, since elements of structure were lost or gained and one-to-one correspondence was not preserved…Deletion always caused the perception of covering, and accretion always caused the perception of uncovering*"). Nakayama and colleagues further developed the concept of surface representation and incisively demonstrated its importance to human vision through ingenious psychophysical experiments (39). In particular, they discovered the astonishing psychophysical phenomenon that accretion/deletion in stereograms is sufficient to produce the percept of surface separation. They termed this form of 3D perception "da Vinci stereopsis", to contrast it with "Wheatstone stereopsis," which concerns the perception of the depth of binocularly visible points (21, 39). Both da Vinci and Wheatstone stereopsis have been formulated in terms of matching between *points* in a pair of images. But the problems of segmentation and object tracking essentially require grouping of *neighborhoods* of points. Thus to make these two problems mathematically and therefore computationally tractable, we had to replace the geometric optics used to explain da Vinci and Wheatstone stereopsis with an ecological optics based on differential topology.

These topological concepts from ecological optics shine new light on many classic ideas in vision research. For example, an occluding contour is typically regarded simply as an intensity discontinuity due to a surface 3D distance discontinuity. Our definition, on the other hand, does not even involve "intensity". In our framework, an occluding contour is simply a singularity in the perspective projection, with the associated property of being an infinitesimal accretion border; this new concept of occluding contour lies at the foundation of our formulation of image segmentation. As another example, invariance has conventionally been regarded as an issue related to object learning. In our framework, invariance is mathematically formulated as an equivalence relation between perspective images of surfaces; the critical equivalence relation is surface overlap, and the machinery for computing equivalence is local diffeomorphism detection. Importantly, the new mathematical formalism carries with it enormous computational power, which we discuss next in relation to computer vision.

*Implications for computer vision*

The theory of topological surface representation has significant implications for computer vision. The theory underscores the importance of equipping artificial vision systems with an explicit surface representation intermediate between pixels and object labels. Furthermore, the theory clarifies that surface overlap is the key mathematical property enabling object tracking. In contrast, most computer vision algorithms for tracking assume that the tracked object should be 'similar' between frames (with 'similar' defined in various ad hoc ways).

Current computer vision methods for video segmentation can be broadly divided into three approaches. One approach ("tracking-by-detection") relies on first segmenting individual objects within single frames and then linking the segmented object instances across frames via some similarity measure (10, 40-43). The fundamental insufficiency of tracking-by-detection as an account of human perception was recognized by Bela Julesz more than sixty years ago (44): human perception of physical reality is first and foremost determined by perspective transformations between images and not by forms within single images (**Fig. S2**). A second approach attempts to perform video segmentation by directly using optical flow as input (11, 12, 15, 45-48). Finally, a third approach in the era of deep learning is end-to-end trained deep networks that take a video as input and output per-frame object detections (e.g., (49, 50)).

While some of these computer vision approaches, especially the ones that use optical flow as input, have kinship to the theory of topological surface representation presented here, their implementations often rely on i) ad hoc assumptions (e.g., that objects constitute clusters of pixels with similar motion patterns, which is invalid for non-rigid objects) or ii) black-box deep learning approaches that do not leverage the principles enabling optical flow to generate object labels. Nevertheless, existing approaches have achieved impressive performance on benchmarks for tracking objects in real-world video (10, 48, 51, 52) and gained valuable insights into how to incorporate learning to build robust segmentation and tracking systems (10-12, 15). We believe such systems may become even more powerful by incorporating a mathematically-grounded surface representation framework *ab initio*. Below, we give four specific arguments why this is advantageous.

*1. Surface representation clarifies what needs to be learned.* Ecological optics breaks the problem of object perception into two halves: i) how surfaces in space are projected into ray spaces and how the diffeomorphisms and breakdowns in diffeomorphisms within the ray spaces encode the surfaces, and ii) how to compute these diffeomorphisms from images. The first half of this problem is a mathematically exact "encoding" problem. The second half is a detection problem, which faces issues of noise and ambiguity. The conceptual insight is that the first step greatly simplifies the problem of vision. The organization of a scene into surfaces is defined by a one-dimensional set of occluding contours, and information to detect these contours is *highly redundantly* available through movements of the observation point, making the detection problem readily solvable (as demonstrated by the fact that in real life, we actually don't encounter many ambiguous visual situations).

We do not underestimate the magnitude of the second half of the problem and the amount of engineering necessary to transform our current algorithm, which works on synthetic video without any noise, into a robust system that works on real-world data. For this purpose, learning will almost certainly be essential: to generate high-quality super segmentation maps that provide the essential input for our system; to handle objects that lack enough pixels to compute border-ownership at edges accurately (e.g., thin shapes like chair legs); and most importantly, to intelligently combine local signals about surface organization into a coherent scene narrative. This last task will require knowledge of natural scene statistics to add breaks or links to the object graphs computed by the bottom-up surface segmentation and tracking mechanism (e.g., for the purpose of "re-identification" (53), in which object identity is preserved even after complete occlusion). Importantly, surface representation vastly simplifies the problem, since statistical knowledge supporting inference can now be expressed in terms of surfaces, which constitute a low-dimensional symbolic representation (**Fig. 6B**).

While it is certainly the case that our system cannot handle real-world video without further engineering, it is equally the case that existing segmentation and tracking systems have fundamental insufficiencies compared to our system, and to the human visual system: we note that if we apply a recent multi-object tracking system to our synthetic video, the results are extremely disappointing (**Fig. S3**).

*2. Surface representation enables self-supervised learning of object recognition from spatiotemporal contiguity in a principled manner.* An influential conceptual framework for object recognition suggests that it constitutes a process of manifold untangling (23), with the essential challenge to untangle tangled manifolds. We suggest that there is an even more fundamental and prior challenge: *finding connected paths along distinct tangled manifolds*. This is precisely what topological surface segmentation and tracking accomplishes. The theory makes the concept of "spatiotemporal contiguity," which has previously been suggested to play an important role in unsupervised learning (54-56), precise. For example, one technique used for the latter is contrastive learning of image views, in which a network learns to make the representations of two different views of the same scene agree and the representations of two views of different scenes disagree (57, 58). However, as noted by Hinton, for a scene with multiple objects, one doesn't want to learn a representation that makes the entire scene in one frame similar to the entire scene in the next frame; rather, one wants to encourage similar representations only for representations of the same objects (59). Topological surface representation provides machinery to achieve this: the tracking mechanism provides a large set of labeled object examples (**Fig. 6A**, bottom row) to pre-train a subsequent invariant recognition module in a self-supervised manner. Thus a visual system initially equipped with diffeomorphism-based surface representation machinery can learn much more effectively than a tabula rasa.

*3. Surface representation may benefit from specialized front-end hardware.* Our topological solution to segmentation and tracking depends on accurate computation of correspondence. While recent emphasis in building intelligent vision systems has focused on developing better learning algorithms and more powerful data sets, for solving the correspondence problem, faster front-end hardware can also make a critical difference. In particular, "event-based" cameras built on the principles of biological retinas to *detect changes* can operate at an effective frame rate of ~50 KHz rather than the typical video rates of 30 Hz, while maintaining low sensing and computing throughput (60). Such cameras would make the correspondence problem significantly easier due to smaller changes from frame to frame and elimination of image blur. Artificial vision systems thus equipped could exploit topological surface representation with maximal efficiency. Together with parallelization of correspondence detection using GPUs (61), we envision that our segmentation and tracking system could operate in real time.

*4. Surface representation unifies segmentation, tracking, and 3D surface reconstruction into a coherent framework.* In computer vision, object segmentation/tracking and 3D surface reconstruction have largely been pursued through distinct paths (for review of the latter, see (62)). In the current paper, we have focused on the former: how diffeomorphisms computed at object edges enable identification of occluding contours and stitching of overlapping surface patches over time. Importantly, diffeomorphisms specify not only the existence of a contiguous surface patch, but also its metric properties of distance and curvature (Shape from Perspective Mapping Theorem, Supplementary Material). Thus the same machinery for diffeomorphism computation, when carried out across the image, should enable accurate surface reconstruction.

### *Implications for biological vision*

We believe our results have important implications not only for building new artificial vision systems but also for understanding biological vision. We currently know a lot about the neural mechanisms for very early image processing such as edge detection (63) and motion detection (64), as well as mechanisms for very high-level object recognition such as face recognition (65). What is missing are the steps in between, which explain how an object first arises: how a set of edges can be transformed into a set of object contours invariantly associated to specific objects. The mathematical and computational solution to this fundamental problem presented here outlines a path for neuroscience research to go beyond search for simple neural correlates of perceptual grouping, to discover the detailed workings of visual surface representation.

The computations we describe for solving segmentation and invariant tracking are necessarily local and therefore likely accomplished in retinotopic visual areas. The invariant label for each object is propagated across the object through local diffeomorphisms between different perspectives—*local threads* (the edges of the scene graph, **Fig. 6B**) create global objects (the connected components of the scene graph and their associated symbolic labels). We believe purely local processes in retinotopic visual areas must generate a representation akin to an object graph, and this object graph structure must already be in place by the output stage of a retinotopic visual area (possibly area V4 or a retinotopic region within posterior parietal cortex (66)). To create an object graph, an essential neural mechanism is needed to represent the *linkages* within the graph. What this binding signal consists of remains unknown and constitutes, in our view, one of the biggest known unknowns in systems neuroscience. Notably, a recent study suggests that the machinery for invariant visual surface representation may be unique to primates (67). One piece of physiological evidence for the existence of topological surface representation in the primate brain is the finding of "border-ownership cells" that show selectivity for side-of-owner of contours (68), a critical topological feature which we show how to compute (**Fig. 5C**). Our theory predicts that the output of border-ownership cells should be integrated over time to generate invariant object labels (**Fig. 4D**), effecting the fundamental transformation of visual information from sensory to symbolic.

The theory of ecological optics presented here is not an arbitrary new model of vision but a mathematical necessity. And each part of the theory maps onto a computational goal and mechanism. The essential simplicity and necessity of the theory set a new direction for vision research to understand in detail how surface representation is accomplished in the brain.

**Acknowledgements**
This work was supported by HHMI and the Kavli Foundation. We are grateful to Xuemei Cheng, Janis Hesse, Michael Maire, Nicholas Masse, Mason McGill, Jennifer Sun, and Albert Tsao for discussions and comments on the manuscript, and to Susan Chang (1947-2014), our wife and mother, for her unwavering love and support in this journey.

**Author contributions**
The ideas were originally conceived by TT and subsequently developed by TT and DYT. Computational simulations were programmed and performed by DYT. The paper was written by TT and DYT.

**Conflict of Interest Statement**
TT is a cofounder and employee of OpticArray Technologies Inc., a company which is focused on applying the topological surface representation concept within machine vision systems, and is an inventor on U.S. Patent 9,087,381, related to these methods.

**Methods**

*Generating a synthetic dataset.* To perform the computational test, we generated a synthetic 160-frame video sequence as follows: we started with an initial 5745 x 5745 image containing four objects bounded by bright contours, each possessing an additional internal texture contour to make segmentation challenging; the texture of both object surfaces and the background consisted of 1/f noise. We then applied four independent sequences of affine transforms to each of the four objects. See https://youtu.be/eu_aJNo3R5I for a movie of the stimulus sequence. In addition to this video sequence, we also generated a corresponding edge map and super segmentation map for each frame.

*Computational implementation of topological segmentation and tracking.* We began by identifying a set of neighborhoods in each super segmentation frame evenly sampling the different types of surface adjacencies (i.e., pairs of distinct, abutting components). To do this, in each frame, for each edge point, we computed four numbers (n, s, e, w) corresponding to the value of the super segmentation map 20 pixels above, below, to the left, and to the right of the edge point. We identified unique neighborhoods as unique combinations of (n, s, e, w). For each frame, we identified 100 random edge points that evenly represented different unique two-sided neighborhoods (i.e., neighborhoods containing just two super segmentation labels). Then, at each point, on each side of the edge, we computed the affine transform (using our dynamic receptive field method for diffeomorphism extraction) between frame i and frame i+1. The end result of this process is that for each frame, at each of 100 edge points, we have two sets of 6 affine parameters, $\vec{p}$ and $\vec{p}'$, corresponding to the affine transforms computed for the two sides of the edge. If the two are the same (determined by the criterion that $|(p_1 - p_1'\ p_2 - p_2'\ p_3 - p_3'\ p_4 - p_4')| < 0.1$ and $|(p_5 - p_5'\ p_6 - p_6')| < 4$), then we classified the edge element as a texture edge, otherwise we classified it as an occluding contour. For edge elements classified as occluding contours, we further determined the owner side by applying the affine transforms computed for the left- and right-hand sides to the left and right parts of the image patch in frame i, respectively, and comparing these to the left and right parts of the image patch in frame i+1 (**Fig. 5C**); comparison was done after projection onto Gabor receptive fields. The side with smaller difference was taken as the owner side. For each unique two-sided neighborhood type, we used winner-take-all across all neighborhoods of that type to determine the consensus classification.

Thus for each frame of the super segmentation map, each unique neighborhood type was classified as one-sided (i.e., occluding edge) or two sided (i.e., texture edge), and for occluding edges the owner side was identified. Furthermore, for i) the owner side of an occluding edge and ii) for both sides of a texture edge, the diffeomorphic super segmentation component(s) in the next frame were identified (**Fig. 4D**, column 2): these constitute the persistent surface components. In cases where a single component splits in two, such that the same surface component in frame i is diffeomorphic to two or more different target surfaces in frame i+1, we picked the label of the modal target in frame i+1 as the one that persists, while others were assigned new labels. Finally, we identified the background label in each super segmentation map as the region for which the difference between the pixel value in the current frame and that in the previous frame was most often 0.

To create the object segmentation map, for each frame, we started with the super segmentation map. We then ran through all the labels in this map, and if a label was part of two-sided neighborhood but was never a one-sided owner, then we reset its label to that of its two-sided partner. This effectively erases all texture edges (**Fig. 4D**, column 1 → column 2).

To create the backward object tracking map, for each frame, we started with the segmentation map. We then set the label of each surface identified as persistent to the label of that surface in the previous frame (**Fig. 4D**, column 2 → column 3). In addition, we set the label of the background to 0.

Finally, to create the post-hoc object tracking map (**Fig. 6A**, row 5), we first built, during the process of forward traversal through frames just described, a scene graph containing a node for every distinct component of the super segmentation map in every frame, and an edge between any two nodes identified as connected, either due to separation by a texture edge, or due to persistency across frames. We then identified the distinct components of this scene graph, and assigned a unique label to each; for our synthetic dataset, there were five labels in total, one for each of the objects plus one for the background. Equipped with this scene graph, we then re-traversed the frames, assigning to each component of the super segmentation map a new invariant object label according to its membership in one of the scene graph components.

*Dynamic receptive field method for diffeomorphism extraction (29, 30).* We projected the input image onto Gabor receptive fields defined on 200 x 200 image patch, at $\theta = 0, 30, 60, 90, 120, 150$ and $\lambda = 10, 20, 30$:

$$g_i = e^{-\frac{x_\theta^2 + y_\theta^2}{2\sigma^2}} \cos\left(\frac{2\pi}{\lambda} x_\theta\right), \begin{pmatrix} x_\theta \\ y_\theta \end{pmatrix} = \begin{pmatrix} \cos\theta & \sin\theta \\ -\sin\theta & \cos\theta \end{pmatrix} \begin{pmatrix} x \\ y \end{pmatrix}, i = 1 \ldots 18$$

We then used Newton's method (**Fig. 5A**) with 10 iterations to find the value of the six affine parameters $\vec{p}$ constituting the zero of the equation

$$E_i = |<A^*(\vec{p}) \circ g_i, I_2> - <g_i, I_1>|^2 \qquad (1)$$

where $A^*(\vec{p})$ is the conjugate of the affine transform $A(\vec{p})$,

$$A(\vec{p}) \circ I(x,y) = I(\begin{pmatrix} p_1 & p_2 \\ p_3 & p_4 \end{pmatrix} \begin{pmatrix} x \\ y \end{pmatrix} + \begin{pmatrix} p_5 \\ p_6 \end{pmatrix}).$$

The rationale for (1) is as follows. By the brightness constancy constraint,

$$E = 0 = |\iint \left[ e^{-\frac{(x_\theta^2 + y_\theta^2)}{2\sigma^2}} \cos\left(\frac{2\pi}{\lambda} x_\theta\right) \right] [I2(u(x,y), v(x,y)) - I1(x,y)] dxdy|^2$$

where $\begin{pmatrix} u \\ v \end{pmatrix} = A \begin{pmatrix} x \\ y \end{pmatrix} + \begin{pmatrix} p_5 \\ p_6 \end{pmatrix}$, $A = \begin{pmatrix} p_1 & p_2 \\ p_3 & p_4 \end{pmatrix}$, assuming I1 and I2 are two image patches related by diffeomorphism.

Thus

$$\overbrace{< g_i, A(\vec{p}) \circ I_2 >}$$

$$E = 0 = |\iint \left[ e^{-\frac{(x_\theta^2 + y_\theta^2)}{2\sigma^2}} \cos\left(\frac{2\pi}{\lambda} x_\theta\right) \right] I2(u(x,y), v(x,y)) dxdy - \cdots |^2$$

$$= |\iint \left[ e^{-\frac{(x_\theta^2 + y_\theta^2)}{2\sigma^2}} \cos\left(\frac{2\pi}{\lambda} x_\theta\right) \right] J_{(u,v)}^{(x,y)} I2(u,v) dudv - \cdots |^2, J_{(u,v)}^{(x,y)} = \det(A^{-1}) = \frac{1}{p_1 p_4 - p_2 p_3}$$

$$\underbrace{\qquad\qquad\qquad\qquad\qquad\qquad\qquad}_{< A^*(\vec{p}) \circ g_i, I_2 >} \quad \underbrace{\qquad}_{\text{Jacobian}}$$

Note that prior to projection of one-sided neighborhoods onto Gabor receptive fields, image patches were shifted to the centroid of the to-be-projected image patch (e.g., image patches shown in **Fig. 4C** were shifted to produce the centered image patches shown in **Fig. 5B**). This ensured that a sufficient non-zero region was available for diffeomorphism detection.

For each neighborhood, we computed the affine transform with starting $\vec{p} = (1\ 0\ 0\ 1\ x\_shift\ y\_shift)$, where x_shift and y_shift ran from -20 to 20 pixels in steps of 5 pixels; this ensured convergence for our dataset. We also computed the value of $E = \sum_i E_i$ at the end of the 10-iteration Newton process for each neighborhood. We then choose the $\vec{p}$ with minimal $E$.

This algorithm assumes that illumination is locally (but not globally) constant (i.e., ambient) and furthermore that surfaces are opaque. Handling transparency and specularity would require further investigation (possibly, effects due to transparency and specularity could initially be treated as noise in a first-pass estimate of surface parameters, and subsequently incorporated in a refined model).

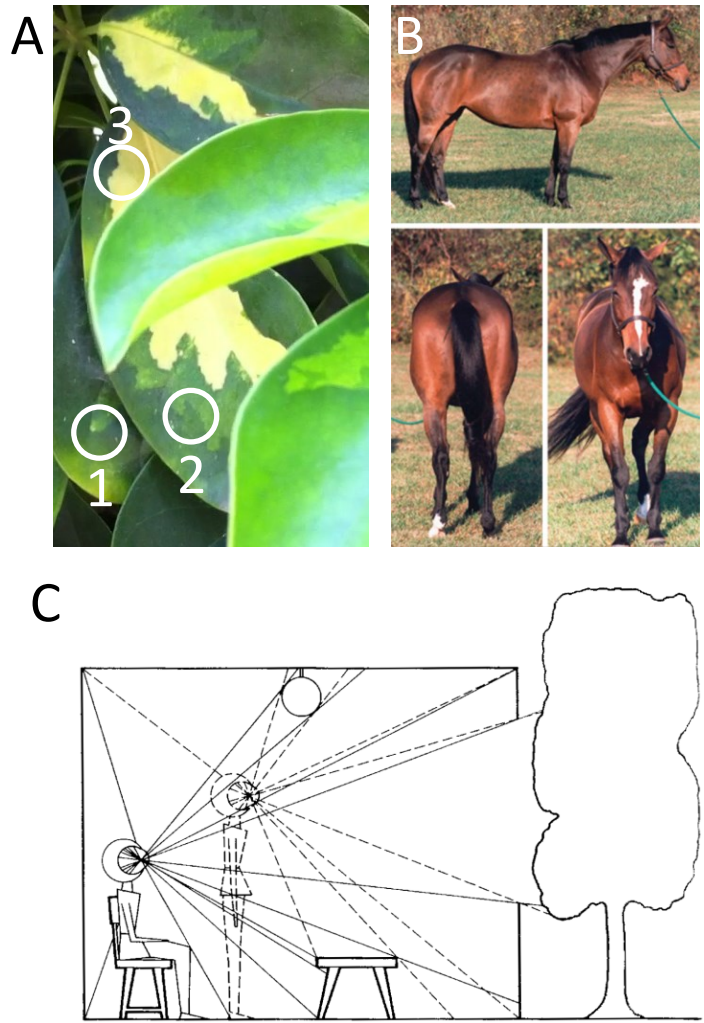

Figure 1

**Figure 1.** The challenge of object segmentation and tracking and Gibson's proposed framework for solution. (**A**) The challenge of segmentation: points 1 and 2 are nearby and have the same color, but belong to different objects, while points 2 and 3 are distant and different in color, but belong to the same object. (**B**) The challenge of invariant tracking: the three views of the horse are very different in shape and pixel composition, yet represent the same object. (**C**) Gibson's ecological approach to visual perception. An array of light rays from objects in the environment is sensed at each point in the observation space (two are illustrated). Gibson asserted that transformations between these arrays contain all the information necessary to solve the segmentation and invariant tracking problems (reproduced from (20)).

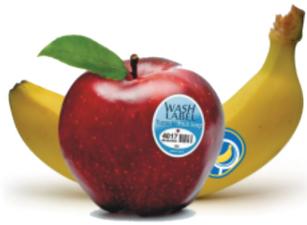
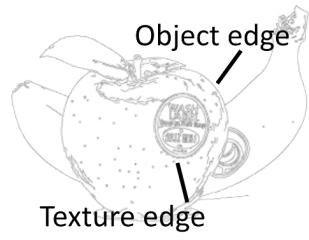
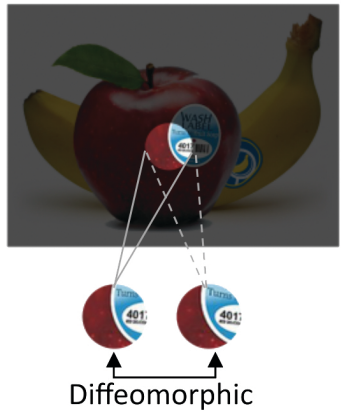
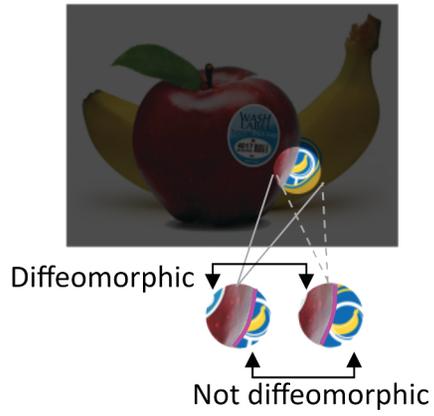
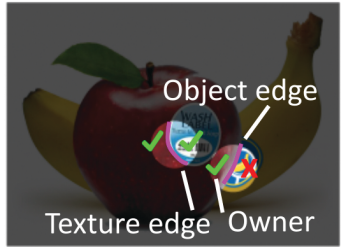
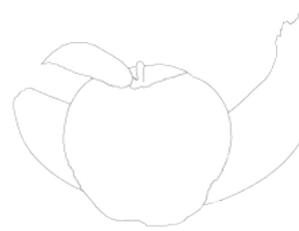
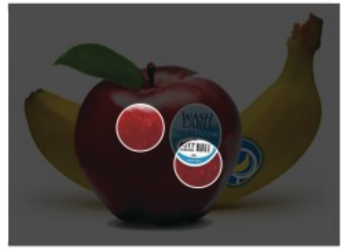
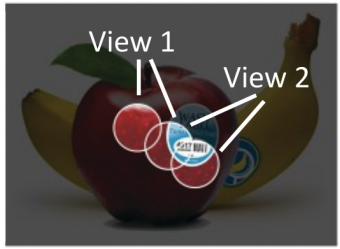
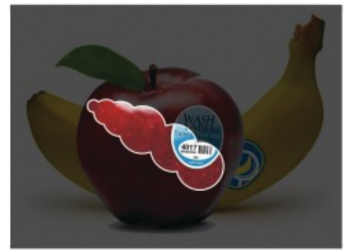

Figure 2

**Figure 2.** Topological solution to object segmentation and tracking: overview. (**A**) An example scene. (**B**) Edge map of the scene. (**C**) The projections of a region of space containing a contiguous surface patch to two observation points are diffeomorphic to each other. (**D**) The projections of a region of space containing an object edge to two observation points are diffeomorphic on one side (namely, the side that owns the edge), but non-diffeomorphic on the other side. (**E**) Two- versus one-sided diffeomorphism provides an effective criterion for distinguishing texture edges from true object edges. (**F**) Removal of texture edges produces a map of true object boundaries. (**G**) The invariance problem posed within the surface representation framework: how can one determine that the two distinct patches belong to the same surface? (**H**) Solution to the invariance problem: If one can identify a third patch overlapping both patches, then all three belong to the same surface. In this way, partial views (View 1, View 2) can be connected through overlaps. Thus the same diffeomorphism computation used to solve segmentation (C, D) can be used to solve tracking. (**I**) Through the equivalence relation of partial surface overlap, all possible views of an object can be identified.

**A** Projection of a point to a ray space

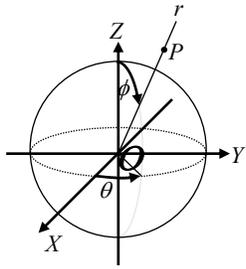

**B** Projection of a surface patch to a pair of ray spaces

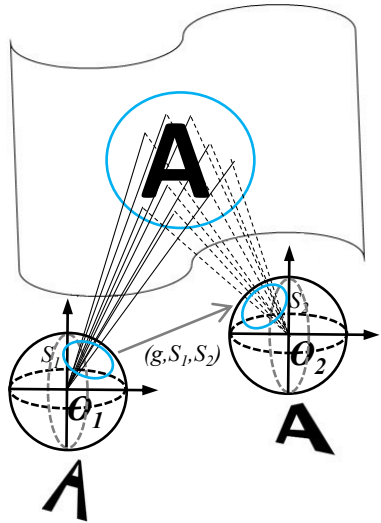

**C** Projection of an object edge to a pair of ray spaces

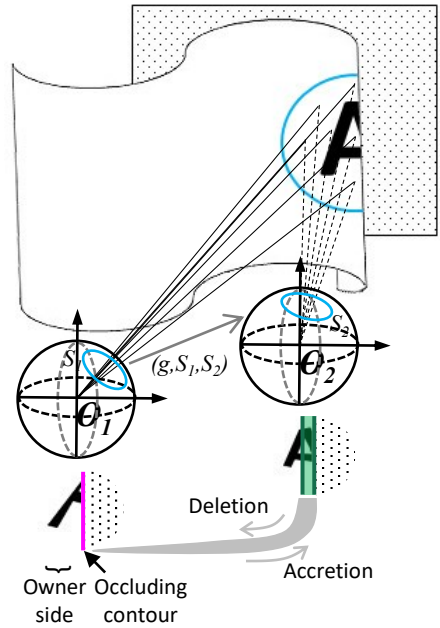

**D** Projection of a connected chain of surface patches to a sequence of ray spaces

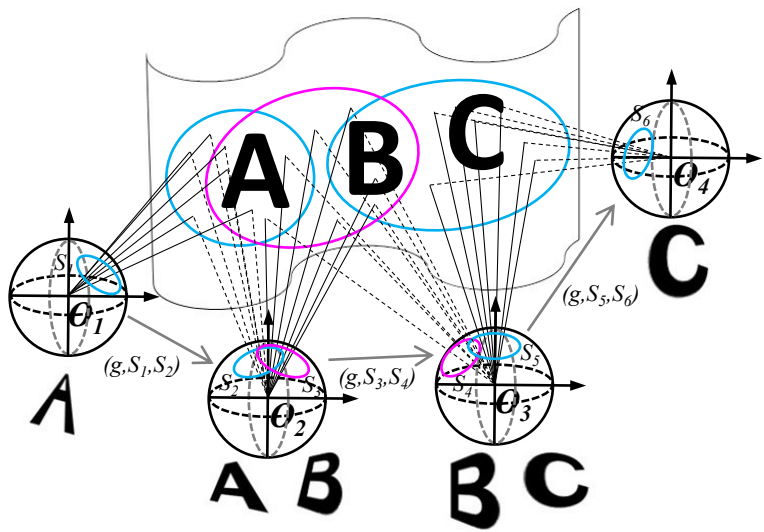

Figure 3

**Figure 3.** Coding local and global surface contiguity through stereo diffeomorphisms. (**A**) Point *P* projects to ray space *S(O)* with value ray *r = (θ,φ)* in a polar coordinate system. (**B**) Surface contiguity (e.g., the surface patch containing the letter "A") in the distal Euclidean space is faithfully encoded in the proximal visual space. If a neighborhood of a point is the perspective projection of a contiguous *local surface patch* in the environment, then a stereo diffeomorphism can be found from this neighborhood to a neighboring ray space. The pairs of intersecting rays correspond to stereo pairs in the transition space $S(O_1) \times S(O_2)$. (**C**) Surface discreteness (e.g., the border of a leaf) in the distal Euclidean space is faithfully encoded in the proximal visual space. No diffeomorphism can be found between a neighborhood in a ray space containing points of an occluding contour to a neighboring ray space. In the figure, an occluding contour segment in the ray space at $O_1$ is marked by the short vertical magenta line; it is the ray space image of an object fold under perspective projection, and constitutes a *border of infinitesimal accretion* because following any change in observation point away from the owner side, e.g., to $O_2$, there is accretion, i.e., the two side images of the border (the two dark green vertical lines) are now regular and have no intersection point. The owner of the occluding contour is specified by the side opposite the accretion (see **Fig. SM14**, Supplementary Material, for more details). Occluding contours provide a compact and complete representation of environment surfaces because all points in a ray space that are not in an occluding contour possess a neighborhood representing a local surface patch. (**D**) Surface persistence in distal Euclidean space is faithfully encoded in the proximal visual space. Image patches A and C in ray spaces based at $O_1$ and $O_4$ represent parts of the same contiguous environment surface because they are connected by a chain of overlapping stereo neighborhoods, i.e., they are *CC(Ω)*-equivalent. In detail: image patch A at $O_1$ and image patch A at $O_2$ are *MS(Ω)*-equivalent, as are image patches B at $O_2$ and B at $O_3$, and image patches C at $O_3$ and C at $O_4$. Image patches A at $O_2$ and B at $O_2$ are overlapping, as are image patches B at $O_3$ and C at $O_3$. Thus the *MS(Ω)*-equivalence class containing image patch A at $O_1$ is connected to the *MS(Ω)*-equivalence class containing image patch B at $O_3$, and the latter is further connected to the *MS(Ω)*-equivalence class containing image patch C at $O_4$. Thus image patch A at $O_1$ is *CC(Ω)*-equivalent to image patch C at $O_4$. This scheme allows extremely different views of the same global surface (e.g., the three views of the horse in **Fig. 1B**) to be perceived as belonging to the same global persistent surface.

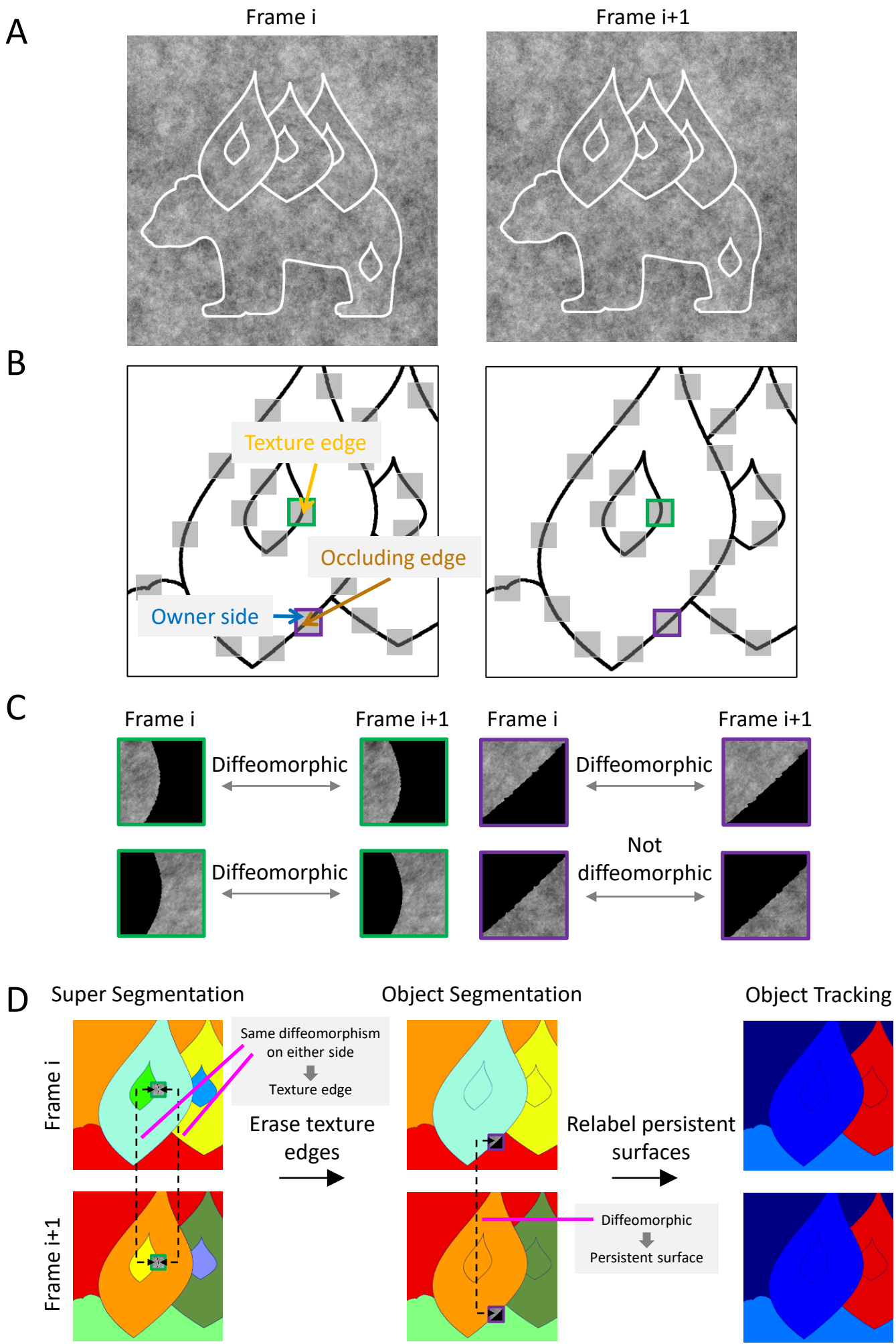

Figure 4

**Figure 4.** A computational implementation of topological segmentation and tracking. (**A**) A scene composed of four objects (a bear and three leaves) viewed from two neighboring observation points (frame i, frame i+1); each object contains an internal texture contour. (**B**) Edge map corresponding to (A), enlarged over one part of the image. The raw edge map includes both texture and occluding edges. To distinguish texture from occluding contours, we first randomly select a set of neighborhoods of edge elements (gray squares. (**C**) For each neighborhood, local diffeomorphism detection is independently performed on the left and right sides between successive frames. Existence of the same diffeomorphism on both sides implies a texture edge (left, green), while existence of a diffeomorphism on only one side (right, purple) implies an occluding edge; moreover, the owner side is specified by the side possessing diffeomorphism (here, the side to the left of the edge). (**D**) Workflow for computing object segmentation maps and object tracking maps. *Left:* Super segmentation map for frames i and i+1 assigning a different label to each contour-bounded component (31); note that texture edges and object edges are treated the same at this stage. *Middle:* Object segmentation maps produced by identifying and erasing texture edges by resetting the label of any pure texture region (i.e., a region that abuts a texture edge but is never a one-sided owner) to that of its two-sided partner. *Right:* Object tracking maps computed by determining persistent surfaces (i.e., in frame i+1, components of the object segmentation map in frame i+1 containing a patch diffeomorphic to a one-sided owner/texture patch from frame i) and assigning them the same label as that in the previous frame.

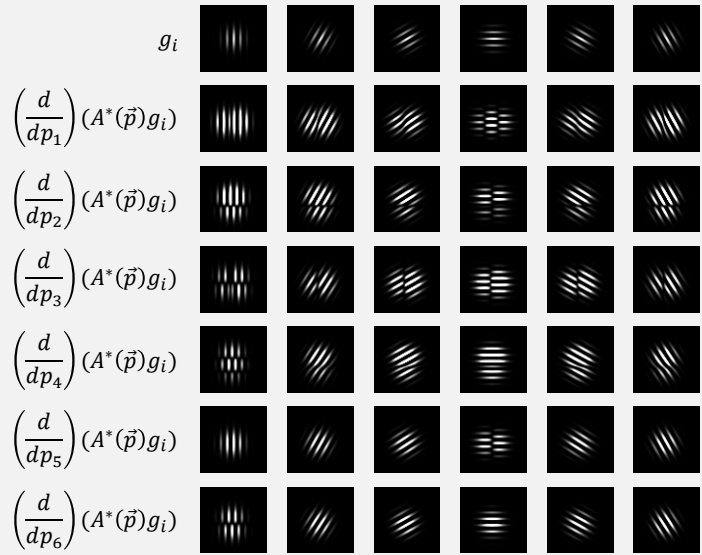
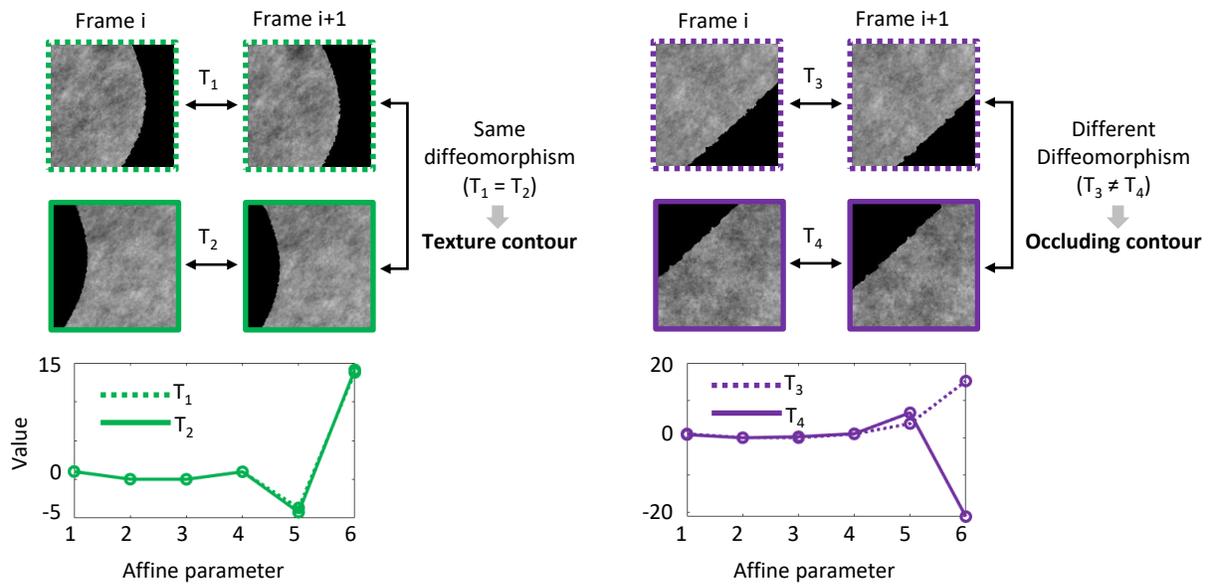
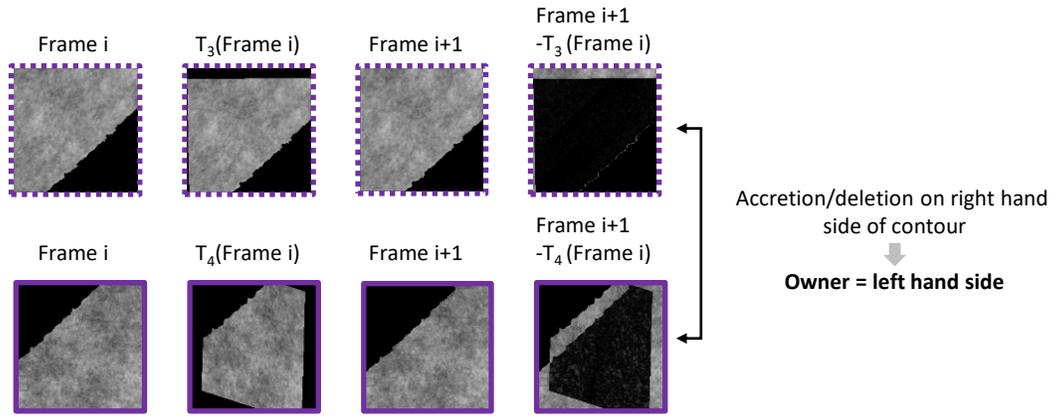

Figure 5

**Figure 5.** Computing diffeomorphisms. (**A**) To compute the diffeomorphism between two patches centered at a point, we project two image patches onto a set of Gabor receptive fields $g_i$ ($i = 1\ldots18$ for 6 orientations and 3 spatial frequencies). *Top left:* Due to the geometry of perspective projection and the brightness constancy constraint, the two image patches will be locally related by an affine transform, $A(\vec{p})$, corresponding to the first term in the Taylor series expansion of the full diffeomorphism; this yields the equation for $E_i$ shown. To compute this transform, we solve for parameters $\vec{p}$ such that the $E = 0$. *Top right:* We do this using Newton's method, which requires computing the derivative matrix $\boldsymbol{E'}$. *Bottom:* This in turn requires computing derivatives of the Gabor receptive fields with respect to each parameter of the affine transform, dubbed "Lie germ receptive fields" (30). (**B**) A pair of image frames from a point centered on a texture contour (left, green) and a point centered on an occluding contour (right, purple); these are the same two neighborhoods shown in **Fig. 4**C; here the center of the patch has been shifted to the left (top) or right (middle) in order to provide a sufficient support for affine transform computation. *Left:* The six parameters of the affine transforms $T_1$ and $T_2$ computed between frames i and i+1 for the left and right neighborhoods respectively are plotted on the bottom. They are equal implying that the contour separating the two neighborhoods is a texture contour. *Right:* The same computation at a different edge point yields affine transforms $T_3$ and $T_4$ for the left and right neighborhoods; these are different, implying that the contour separating the two neighborhoods is an occluding contour. Note that computing a difference in diffeomorphism between the two sides is equivalent to computing a diffeomorphism breakdown on one side, but is computationally easier since it does not depend on detecting non-convergence of Newton's method. (**C**) At occluding contours, the foreground side owns the contour. To determine the owner, we apply the affine transforms computed for the left- and right-hand sides ($T_3$, $T_4$, respectively) to the left and right parts of the image patch in frame i (column 2), and compare this to the left and right parts of the image patch in frame i+1 (column 3). For the owner side, these should be identical (top), while for the occluded side, there should be a border of accretion/deletion leading to difference (bottom). Here, this process reveals a border of deletion to the right of the contour (column 4, bottom), implying that the owner is to the left (see Fig. 4B for zoomed out view of the patch). Note that differences in column 4 are projected onto Gabor receptive fields, thus differences at the edges are discounted.

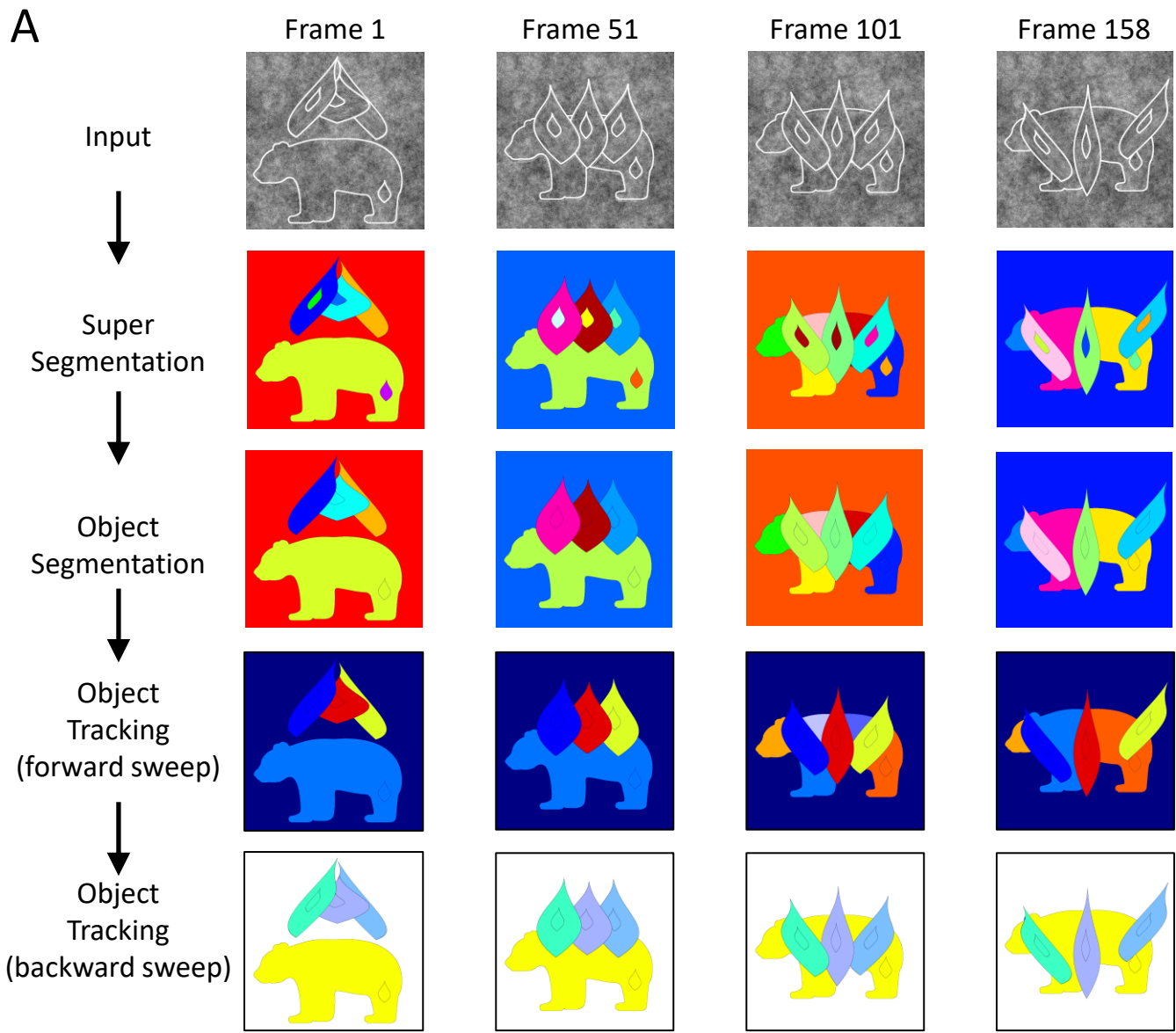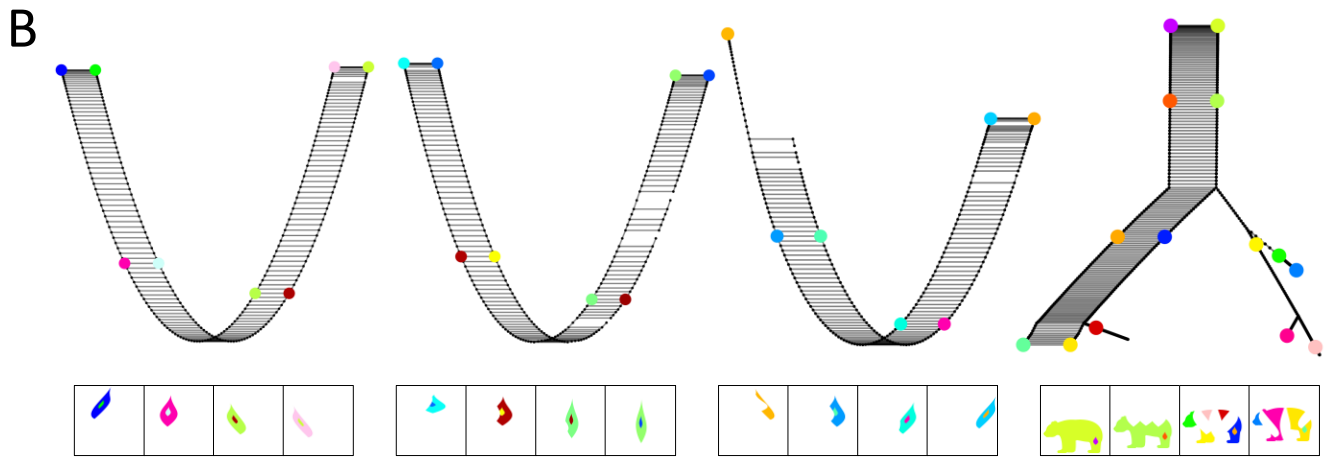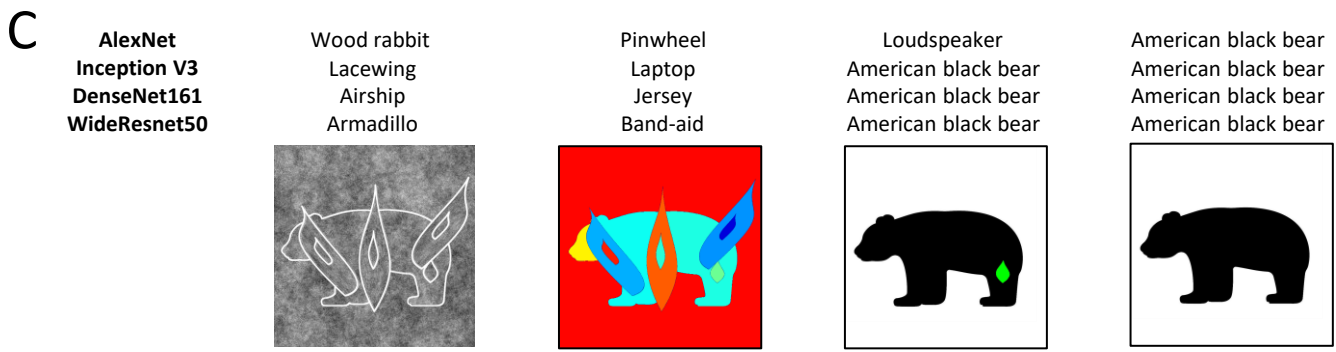

Figure 6

**Figure 6.** Segmenting and tracking objects in a synthetic dataset containing multiple objects despite severe appearance changes due to object deformation, changing perspective, and dynamic occlusion. (**A**) The output of the segmentation and tracking system after each stage of processing (cf. **Fig. 4D**). *Row 1:* Input images at four distinct time points. *Row 2:* Super segmentation maps. *Row 3:* Object segmentation maps. *Row 4:* Object tracking maps. *Row 5:* Revised object tracking maps computed via a backward sweep after computation of invariant object graphs. (**B**) Four connected components of the scene graph computed from this synthetic dataset, corresponding to the three leaves and the bear. Each vertex corresponds to a distinct super segmentation component. Vertices of each graph component corresponding to the frames shown in (A) are indicated by color. The corresponding super segmentation components are reproduced in the frames below each graph component. Note how tracking is robust to severe changes in shape due to object deformation, changing perspective, and dynamic occlusion. (**C**) Four images from a single frame, taken from different processing stages in the topological segmentation and tracking workflow: (from left to right) visual input, super segmentation map, tracked surface component with texture patch distinguished, tracked surface component with texture patch removed. The corresponding classification of each image by four different deep networks is indicated above. Through topological segmentation and tracking, the cluttered input image (left) can be transformed/linked to an un-occluded representation of an isolated surface.

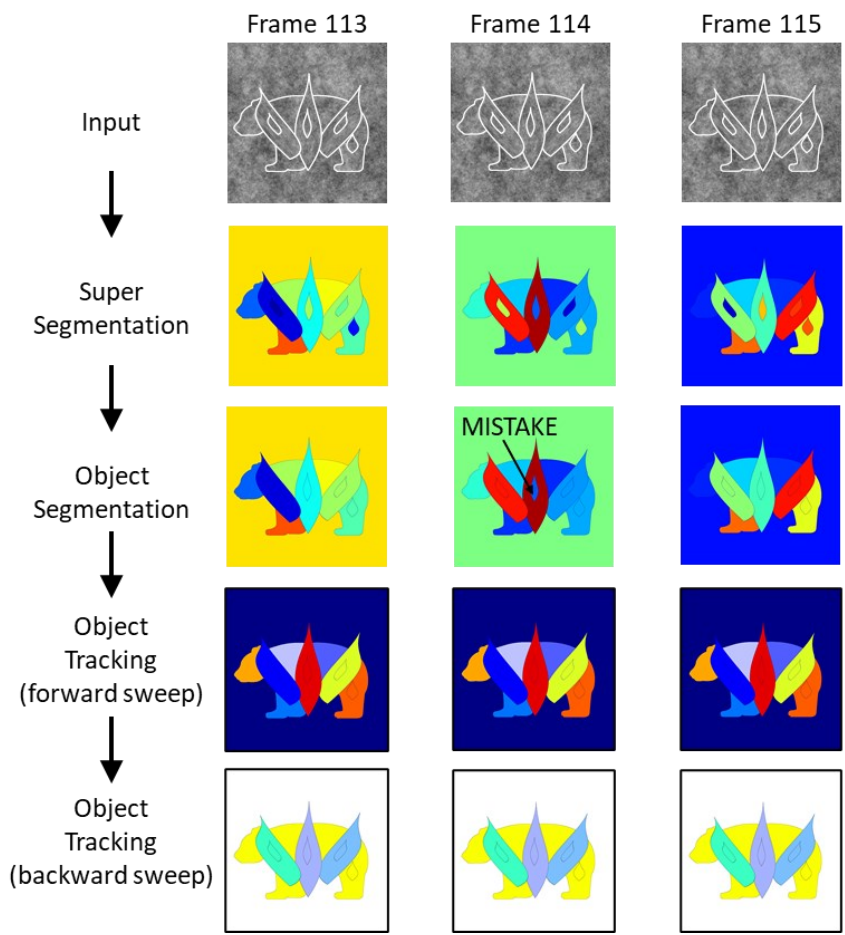
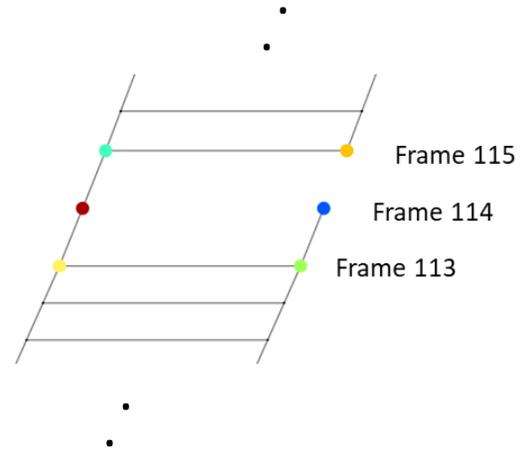
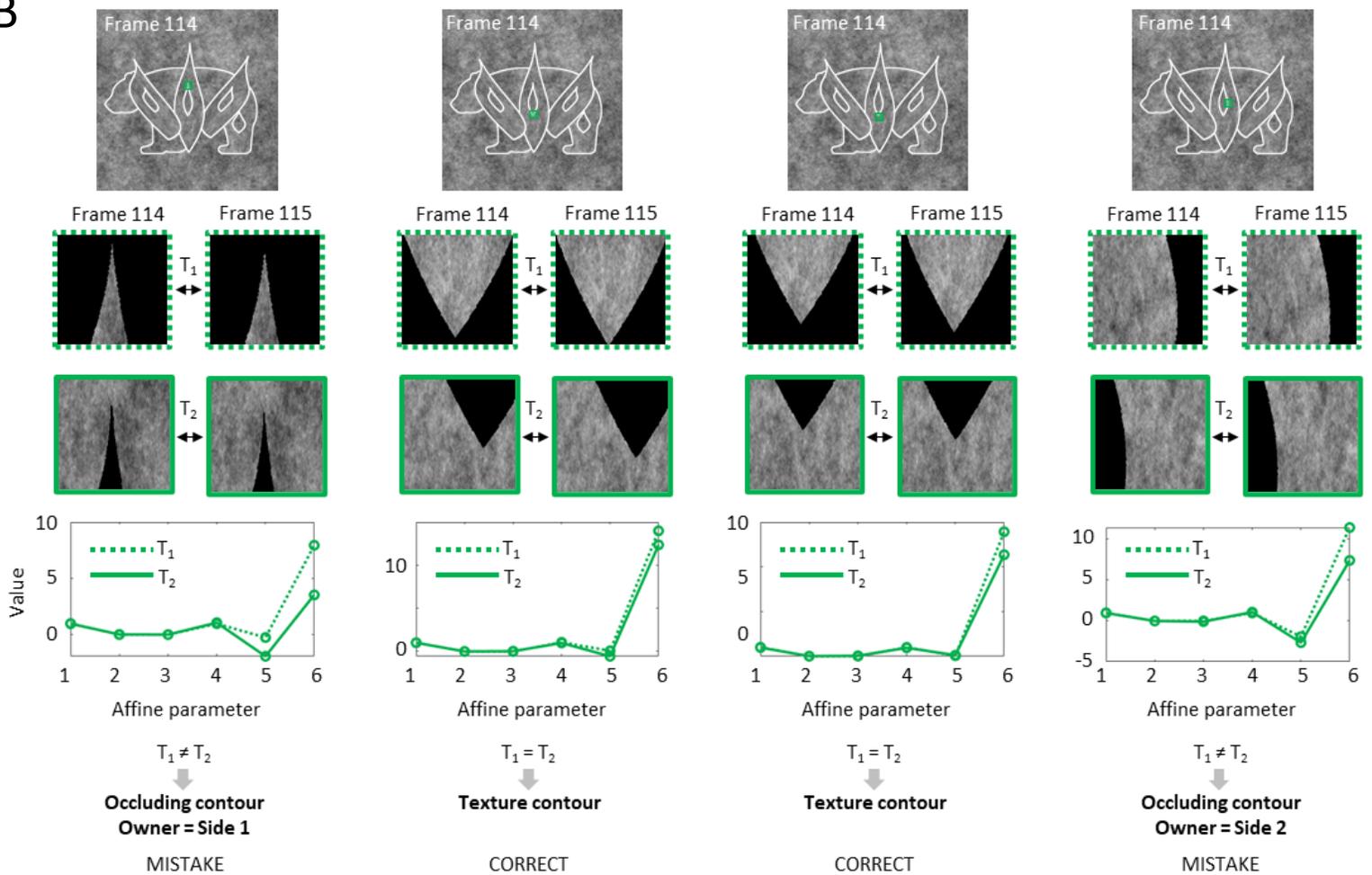

Figure S1

**Figure S1.** The segmentation and tracking process is highly robust due to redundancy of information for surface representation. (**A**) A frame in which a mistake is made in object segmentation, together with two adjacent frames. Conventions as in **Fig. 6A**. (**B**) Pinpointing the source of error: two-sided diffeomorphism detection was performed for neighborhoods of four edge points (top row) along the contour separating the inner from outer component of the middle leaf. Conventions as in **Fig. 5B**. Two of the four neighborhoods are correctly classified as containing texture contours (columns 2, 3), while two are mistakenly classified as containing occluding contours (columns 1, 4). (**C**) Portion of object graph corresponding to the segmentation mistake. The blue dot in frame 114 should be connected to the red dot in frame 114, since they are separated by a texture contour. However, due to mistakes in local diffeomorphism detection, the contour is misclassified in this frame as an edge contour, and thus the connection between between the red and blue dots in the object graph is missing. Nevertheless, due to a connection with the previous frame representing surface persistency (green -> blue), the super segmentation component represented by the blue dot remains connected to the correct global object graph, and the mistake is absent from the object tracking map for the same frame. This example illustrates how redundancy in information for surface representation leads to robustness.

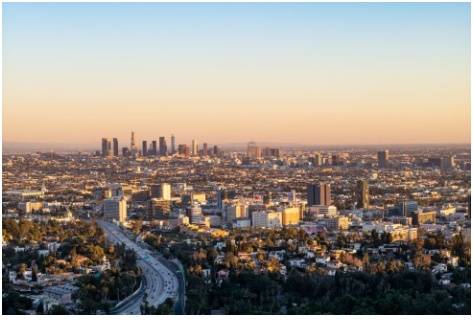 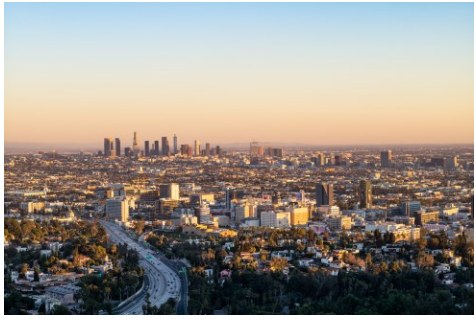 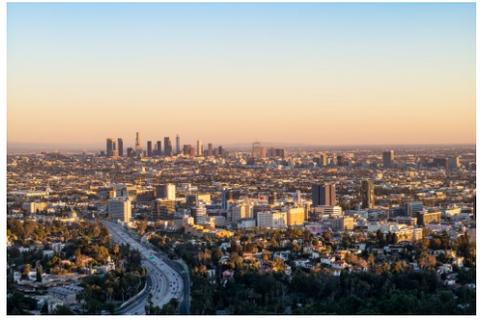

Figure S2

**Figure S2.** The primacy of perspective transformations for determining perception of physical reality. A stereogram of a bear is constructed from a texture pattern consisting of LA skyline. Tracking-by-detection approaches would attempt to segment the buildings, freeway, and other objects in the two frames and then track these and would thus completely miss the bear that pops out only through stereo.

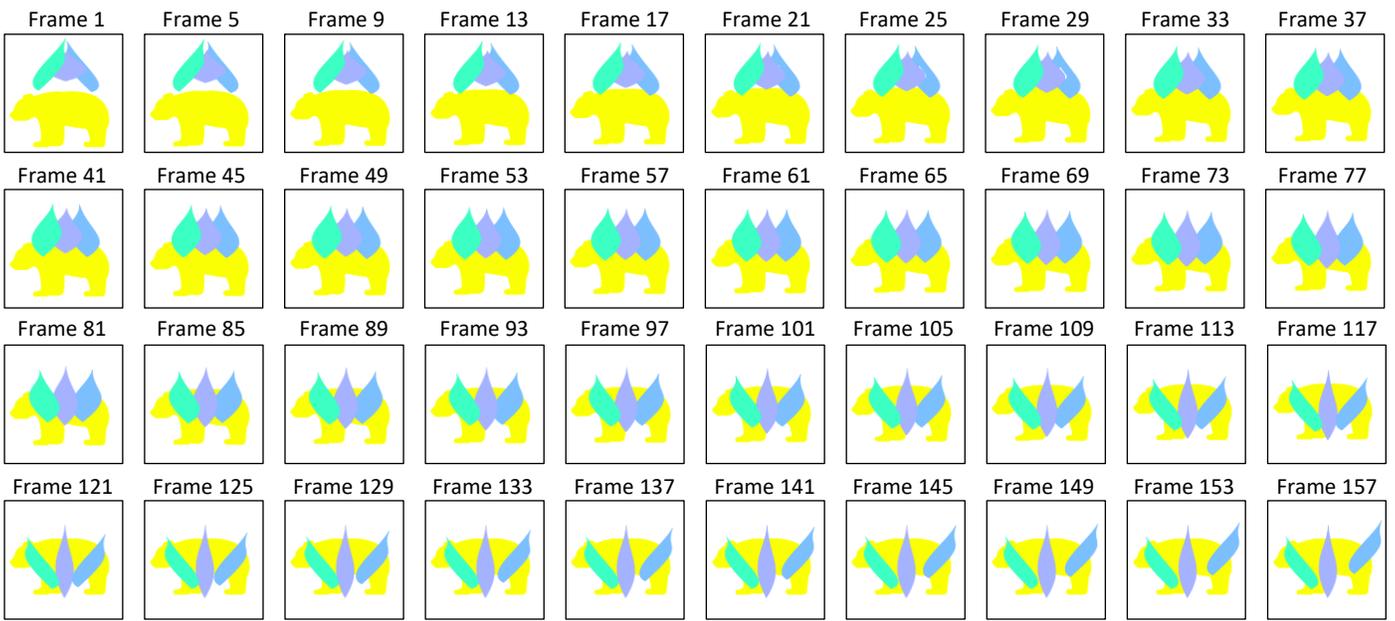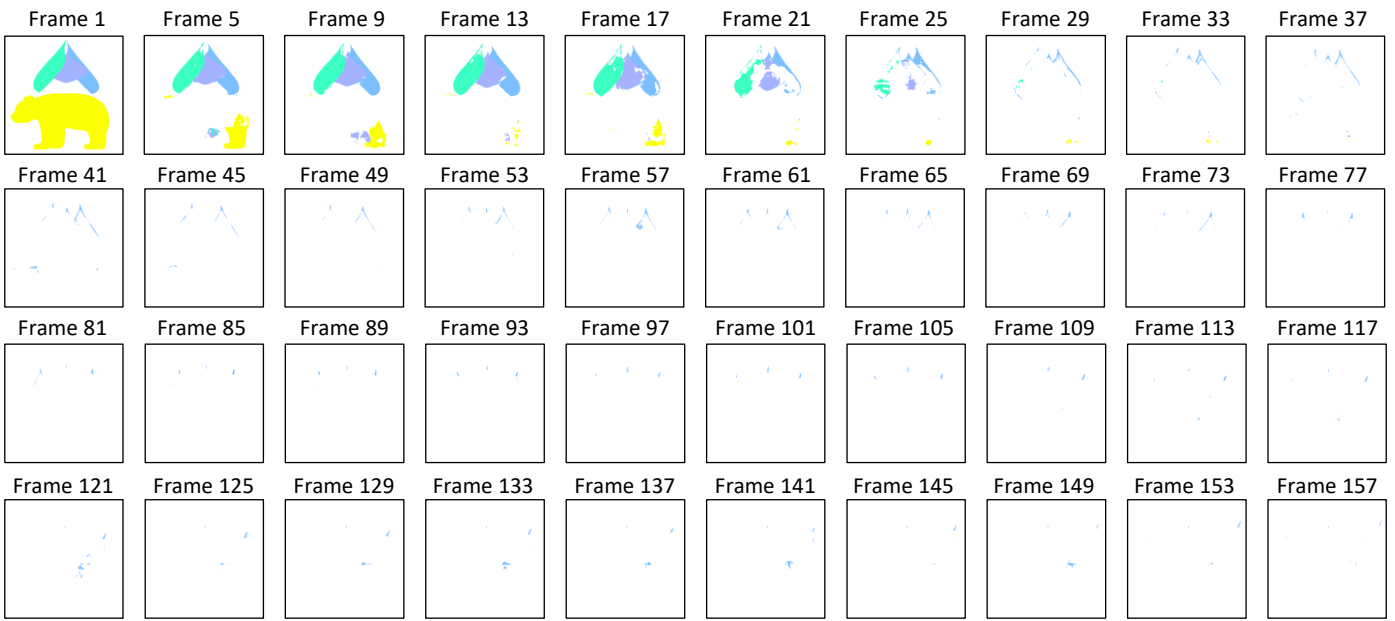

Figure S3

**Figure S3.** Comparison of segmentation and tracking results using our method and a recent multi-object tracking algorithm (69). (**A**) Tracking results for every fourth frame using our method. (**B**) Tracking results on the same synthetic video using a semi-supervised video tracking algorithm that uses a space time memory network. Tracking fails very early, likely because the network relies on interpolation within the training distribution instead of building a cue-invariant surface representation.

# Supplementary material:
# A topological solution to object segmentation and tracking

*Thomas Tsao and Doris Y. Tsao*

## Table of Contents

*"My explanation of vision was then based on the retinal image, whereas it is now based on what I call the ambient optic array."*

-J.J. Gibson

**1. Introduction**

A fundamental quest in vision research is to understand how we perceive the invariant environment from the rapidly changing stimulus sensed during locomotion. What data in the changing visual stimulus specifies the persistent surfaces of invariant 3D objects? In this Supplementary Material, we present the full mathematical theory of ecological optics. We demonstrate that the topological and geometrical structure of the environment is specified by certain structures of light accessible to a moving observer. For the sake of clarity, we have organized this exposition to be self-contained, including additional background context as well as discussion of implications. This necessitates some redundancy with the main paper.

1.1. Seeing objects in 3D space

The eyes of terrestrial animals have evolved to see their environment: the dangers and resources in their surroundings, and the paths and obstructions that require proper actions to avoid the dangers and access the resources. Environmental objects are first of all seen as things outside of the viewer that occupy a part of space that the animal could otherwise move into. The spatial presence of objects in visual awareness sets a general condition for and prior to the rest of visual experience: shape, color, texture, distance, movement, etc.

The problem of how we see things in space was a subject of philosophical debate between empiricism and innatism. For example, Helmholtz asserted that the way the human vision system perceives distances to objects in space from motion parallax is by unconscious inference based upon past experiences (1). In Kant's philosophical statement, the form of "outer intuition" is an *a priori* given condition for any visual experience; space as the form for object representation is innate in the human mind. Despite their differences, both agreed upon one thing: the sense of objects and their surfaces in space cannot be only based upon the images of objects projected to the eyes. The 3D representation of objects and their surfaces needs additional support, either from *past experiences* or from an *a priori* construction of the human mind.

Both theories make an implicit assumption: visual information is conveyed through the retinal images in the form of local color, intensity, texture, contours, etc. In this article, we demonstrate that light rays converging on points of observation in the environment have much richer structures than those commonly associated with retinal images, and a vision system is capable of picking up these structures and extracting invariants from them to generate a 3D surface representation of the environment.

1.2. Visual information for invariant surface perception

Instead of appealing to past experiences or *a priori* mindsets, Gibson offered a different answer to the puzzle of how the brain sees invariant surfaces in space (2). He expanded the origin of visual information from the narrowly focused *retinal images* to a much richer source: the structure of ambient lights, the *optic arrays*, and asserted that the ambient lights surrounding the viewer fully specify the 3D structure of environment surfaces.

The conceptual expansion from vision based upon retinal images (more precisely, structures commonly associated with retinal images such as local color, intensity, texture, contours, etc.) to vision based upon ambient optic arrays leads to a surface-based vision theory. To understand how the ambient lights may carry more information than that already included in the form of retinal images, we need to look closely at the overall structure of ambient lights. In explaining the origin of the notion of 3D space in vision, Poincarè introduced a new concept into vision: *perspective transition* during eye movement, and pointed out that the perception of geometrical space of three dimensions is derived from the isomorphism between composition laws of perspective transitions and that of movements of invariable objects.

> "The images of external objects are painted on the retina, which is a plane of two dimensions; these are perspectives. But as eye and objects are movable, we see in succession different perspectives of the same body taken from different points of view. We find at the same time that the transition from one perspective to another is often accompanied by muscular sensations. … Then when we study the laws according to which these operations are accompanied, we see that they form a group, which has the same structure as that of the invariable bodies. Now we have seen it is from the properties of this group that we derive the idea of geometrical space and that of three dimensions. We thus understand how these perspectives gave rise to the conception of three dimensions, although each perspective is of only two dimensions, - *because they succeed each other according to certain laws.*" (pages 68-69, (3)).

Poincare described two layers of structure in ambient lights. One is recorded in the retinal images, where ambient lights are structured according to their colors and intensities into a two dimensional sheet. Poincare called this layer of structure "perspectives". The other layer of structure is recorded in the perspective transitions, or changes between perspectives. As we will show, these changes include perspective transformations of correspondent areas and accretion/deletion of non-correspondent areas. This *inter-perspective* structure of ambient lights is relatively independent of the perspective structures involved in the transition: e.g., the same inter-perspective structure can apply to two random dot images or two pictures of objects. Poincare pointed out that the source of visual information for perceiving things in 3D space is the inter-perspective structure of ambient lights.

In Section 2, we introduce the concept of perspective transition groupoid to model the inter-perspective structure of ambient lights, and show that this groupoid completely specifies 3D surfaces. Then, in Sections 3 and 4, we describe key elements of a computational approach for picking up this inter-perspective structure. Finally, we summarize the results and discuss some interesting, important, and surprising conclusions with less restriction.

## 2. The representation of invariant surfaces in perspective transitions

In this section, we prove that perspective transitions fully specify 3D surfaces in the environment. **Fig. SM1** show two sets of light rays cast by the visual environment through two different focal points. A perspective transition is simply an ordered pair of two sets of light rays.

In Section 2, we define a mathematical model of the environment. Then, we first show, through the *3D embedding theorem,* that 3D space is homeomorphically mapped through biprojection to a 3D submanifold within the product of a pair of 2D ray spaces. To understand this better, think of 3D space as filled by points like points A and B in **Fig. SM1**, and consider projection through two observation foci. Each of these points maps to a pair of rays, and these ray pairs form a subset of the 4D space of ray pairs. The 3D embedding theorem shows that the set of points in the 3D world and the subset of ray pairs under biprojection have the same topological structure. In other words, biprojection preserves the topological structure of the 3D world within the pair of target ray spaces. This is important because it means the animal, through its two eyes or the movement of one eye, can directly access the topological structure of the 3D world.

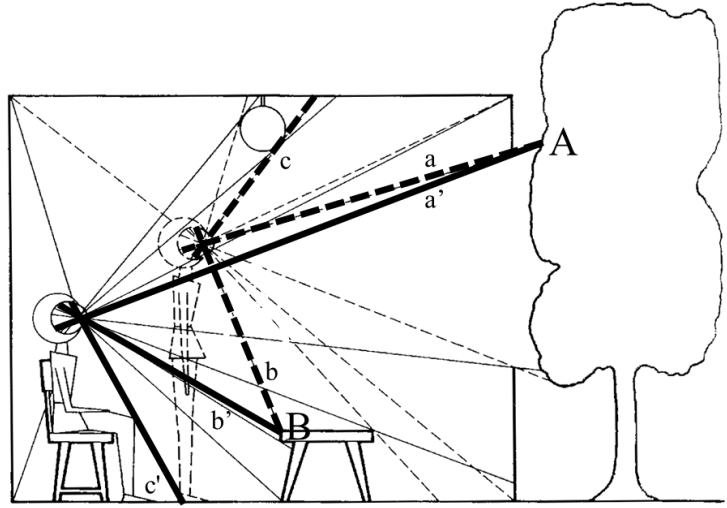

**Figure SM1.** The visual environment casts two sets of light rays through two different focal points ("biprojection"). At each focal point, the light rays span a 2D manifold, called the ray space. The pairs of rays {a,a'}, {b,b'}, and {c,c'} are each elements within the 4D space of pairs of rays. The pairs {a,a'} and {b,b'} are both images of points in 3D space under biprojection. The pair {c,c'} is not.

The visual environment is not a dense set of points in 3D space but is composed of a set of 2D surfaces. The 3D embedding theorem already implies that 2D surfaces in the world are topologically mapped into the product of a pair of rays spaces. However, it is not sufficient for an animal to simply know that the surfaces in the 3D world are mapped *somehow* to a submanifold within a pair of ray spaces. The animal needs to perceive the actual surface, namely the surface with specifics of its embedding in the 3D world. We next show, through the *surfaciness theorem*, that the topology of surfaces is coded by the perspective transformations between pairs of ray spaces. The surfaciness theorem shows that given a pair of ray spaces, a mapping from a domain in one ray space to another ray space fully specifies a surface in the environment. Given multiple pairs of ray spaces, the mappings provide multiple surface representations. We derive through the *fixed vertex theorem* that these different representations form an equivalence class that codes an invariant surface. Next, through the *shape from perspective mapping* theorem, we show that the surface geometry (not just topology) is also coded by the perspective mappings between ray spaces.

Surfaces in the world are always only partially visible from any single perspective due to self-occlusion and mutual occlusion (e.g., the front always occludes the back). How can an animal, which accesses only partial sensory samples of a surface at a time, perceive the same global surface all the time during its locomotion? In the last part of Section 2, we derive the mathematical conditions under which partial samples can be associated to a single, invariant surface. In Section 3, we show how the transition vectors, namely the movements of the eye, are mathematically related to the perspective transitions.

2.1. A mathematical model of the visual environment

In this subsection, we first present a model of a 3D environment composed of rigid surfaces. We define the concept of natural perspective, which models how light rays are structured by the 3D environment. Then we describe a model of locomotion around the environment. Finally, we define the concepts of perspective transition and the groupoid of perspective transitions. This will allow us, in subsection 2.2, to demonstrate how 3D space can be represented in terms of perspective transitions.

Our model of the visual world is a 3D Euclidean space partially occupied by compact objects and the ground. The objects constitute a countable set. The objects and the ground are bounded by smooth, compact, and orientable 2D manifolds, i.e., surfaces. The rationale for the surface smoothness assumption is that rough surfaces can be approximated by smooth ones, and with smoothness we can apply techniques of mathematical analysis.

We also assume a surface called the *sky,* a sphere composed of "ideal points" which have infinite distance to any point in the Euclidean space. All objects in the environment including the ground are inside this sphere. The object surfaces and the ground surface in the Euclidean space are called *ordinary surfaces*. Together with the sky they constitute a *visual environment*.

Given a coordinate system, the Euclidean space is a topological space with the metric topology from the coordinates of its points. An ordinary surface is also a topological space with the relative topology of its ambient Euclidean space. The sky has the topology of the unit sphere at the origin of the coordinate system. The visual environment is the topological sum of the sky and the ordinary surfaces. A *connected open set* in a topological space is called a *domain*. The domain outside of the objects and the ground under the sky is called the *observation space*. We assume each ordinary surface is a connected component, which means there is no contact between different ordinary surfaces. It is a simplification of the real world.

The "connected components" are the basic units of the visual environment. The definitions of connected set (4) and components (5) are quoted as follows:

**Definition (Connected Set)** A topological space $X$ is said to be *connected* if it is not the union of two non-void disjoint closed sets. A set $U$ in a topological space is *connected* if it is not the union of two nonvoid disjoint sets, both open in the relative topology of $U$.

**Definition (Component)** Let $X$ be a topological space and let $x \in X$. The *component* of a point $x$ in $X$ is $C(x)$, the union of all connected sets containing $x$. It is a connected closed set. If $A \subset X$, the *components* of $A$ are the components of the points of the subspace $A$.

The geometric concept of *ray* plays a special role in our mathematical description of vision. A ray is a half line: an ordered set of points with a minimum element called the starting point of the ray and no maximum element. A ray r is said to pass a point $P$ if $P$ is a non-starting point on *r*. We call a ray with an assigned brightness a light ray. We assume that in the environment, only those rays with starting point on some environment surface and directed towards the air (rather than the surface interior) are light rays. We call the environment illuminated if every observation point has light rays passing in every direction.

The points and other mathematical objects, such as surfaces in the environment, are described by coordinates in a 3D Euclidean space: $E^3(O) = (\mathbf{i}, \mathbf{j}, \mathbf{k}; O)$. At a given observation point $P$, from a given direction, an infinite number of different rays could pass the observation point, varying in their starting point. For convenience in describing these equivalence classes, we set a *reference frame* by translating the origin to $P$: $E^3(P) = (\mathbf{i}, \mathbf{j}, \mathbf{k}; P)$. A ray with starting point $P$ may then be given by the coordinates of points along the ray. By using a spherical polar coordinate system:
$ray(\theta, \phi) =$
$\{(\rho, \theta, \phi) = ((x^2+y^2+z^2)^{1/2}, \cos^{-1}(x/(x^2+y^2)^{1/2}), \cos^{-1}(z/(x^2+y^2+z^2)^{1/2})) | \rho \geq 0\}$ the ray can also be represented simply by its two coordinate $(\theta, \phi)$ when there should be no confusion.

The $(\theta, \phi)$ description for a ray passing point $P$ leaves the starting point unspecified. We call $(\theta, \phi)$ the *direction* of a ray passing point $P$. For a given illuminated environment and a point of observation, along each direction there is one and only one light ray passing that point. Thus for a given illuminated environment, we can define the brightness of a direction as the brightness of the unique light ray along that direction; we call a direction with assigned brightness an *illuminated direction*.

The set $S(O)$ of rays starting at a point $O$ is called the *ray space* based at $O$. Point $O$ is the base point of $S(O)$. The rays in $S(O)$ are one-to-one correspondent to the points on the *unit sphere* around $O$. We equip the set $S(O)$ with the metric topology of the unit sphere, and often make no distinction between a ray in the ray space and a point on the unit sphere surrounding the base point of the ray space. The base point of a ray space is also called the center of the ray space. The set $A(O)$ of illuminated directions is called the ambient optic array based at $O$. The ambient optic array is a ray space in which each direction has an assigned brightness.

To give coordinates to all incoming directions to a point in an observation domain, we define the mathematical structure $VS(\Omega) = S \times \Omega$. This is a trivial *fiber bundle*, with bundle projection map

$\pi: VS(\Omega) \to \Omega$,

$\Omega$ is the observation domain, $S = \{(\theta, \phi) | \theta \in [0, 2\pi) \text{ and } \phi \in [0, \pi]\}$ is the *ray space*. $\pi^{-1}(O)$, the ray space at the point of observation, is denoted $S(O)$. We call $VS(\Omega)$ the *visual space* on the *observation domain* $\Omega$.

**Definition (Visible Domain, Stably Visible Domain)** A point of the visual environment is *visible from a point* of observation if the line segment connecting these two points does not intersect any other points on an ordinary surface. A domain of the visual environment is *visible* from a point of observation if every point in the domain is visible from the point of observation. A point of the visual environment is stably visible from a point of observation if it is also visible from a neighborhood of the point of observation. A visible domain is *stably visible* from a point of observation if it is also visible from a neighborhood of the point of observation. A visible domain is *locally stably visible* if each point of the domain has a neighborhood that is stably visible.

**Definition (Point Projection)** Let $S(O)$ be a ray space, and $P$ a point in the Euclidean space, $P \neq O$, $r \in S(O)$ and $P$ on $r$, the function from point $P$ to the ray space $S(O)$ that takes $r$ as its value is called a *point projection*. The open line segment on the *target ray* from the *source point* to the center of the ray space is called the *path* of a point projection.

**Fig. SM2** shows a point $P$ in the Euclidean space with polar coordinate $(\theta, \phi, \rho)$ is projected into the ray space with value $(\theta, \phi)$.

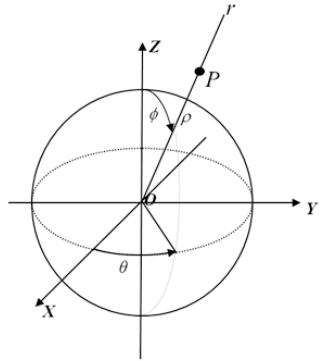

**Figure SM2.** Point $P$ projects to ray space $S(O)$ with value ray $r = (\theta, \varphi)$ in polar coordinate system.

**Definition (Natural Perspective, Perspective Transition)** A *perspective projection* from an ordinary surface in the environment to a ray space is a map from the visible part of the surface to the ray space taking each point on the visible part of the surface as a source and taking its value under point projection to the ray space as the target. The *ambient perspective projection* to a ray space is a map from the visible part of the visual environment to the ray space through point projections. An ambient perspective projection is also called a *natural perspective*, or a *perspective*. When the environment is illuminated, the image of the natural perspective at *P* is a set of illuminated directions, called the *ambient optic array*. An *ambient perspective projection* to a *visual space* is a map from the visual environment to the visual space through ambient perspective projection to each of its ray spaces. An ordered pair of perspectives in a visual space is called a *perspective transition*. For brevity it is denoted a *P-transition*. Base points of the ray spaces in perspectives or in *P*-transitions are called base points of perspectives or *P*-transitions.

Some concepts need to be further clarified. (1) A perspective is a projection mapping from visible environment surfaces to a ray space. Through the source of the mapping, the visible environment surfaces, two perspectives in a perspective transition are related. (2) An ambient optic array is a set of illuminated directions at an observation point, i.e., rays with starting point at the observation point and an assigned brightness. Gibson carefully distinguished between the ambient optic array and the 2D "retinal image." One can make the set of illuminated directions look like a "retinal image" by borrowing the topology and coordinate system from the unit sphere around the point of observation (or in the case of a pinhole camera, the image plane behind the point of observation). However, the biological vision system cannot forget that the nature of an illuminated direction is a ray, not a point on the retina. The unit sphere represents a ray space, which includes an observation point hiding outside the 2D sheet, and as we shall see, this point plays a critically important function in 3D vision.

The observation space is not homogeneous for 3D movements: at each point, the allowed movements are limited by environment surfaces. An observer cannot move into the ground or inside an object. Instead of the whole 3D translation group, the set of allowed movements at each point in the observation space is a subset of the 3D translation group, different from place to place. Movements in the observation space can be represented by line segments each with specified start and end points, which we call *steps*. Unlike the situation with vectors in a vector space, the composition law of steps can only be partially defined at best: In a convex subspace of the observation space, if the end point of one step is the start point of a second step, the steps can be added together to yield a third step. If the space in not convex, even this partial composition can fail.

Our intuitive notion of locomotion in the observation space contains two concepts: (1) steps, and (2) paths, where a movement from one place to another place with a sequence of steps is called a path. Two consecutive paths can always be added together to form a longer path. The set of all paths thus formed in the observation space equipped with the partially-defined composition law we have just described has a group-like algebraic structure, including the trivial (zero) path, the inverse path of a path, and the association laws of compositions of paths. The algebraic structure of this path model of animal locomotion in its environment is a *groupoid*. The definition of groupoid was first given by Brandt ((6), also see (5, 7)).

**Definition (Groupoid)** A *groupoid* with base ***B*** (or "over ***B***") is a set ***G*** with mappings $\alpha$ and $\beta$ from ***G*** onto ***B*** and a partially defined binary operation $(g, h) \mapsto gh$, satisfying the following four conditions:

1. $gh$ is defined whenever $\beta(g) = \alpha(h)$, $\alpha(gh) = \alpha(g)$, and $\beta(gh) = \beta(h)$.
2. Associativity: If either $(gh)k$ or $g(hk)$ is defined, then both are defined and they are equivalent.
3. For every $g \in G$, there are left- and right-identity elements $\lambda_g$ and $\rho_g$ respectively, satisfying $\lambda_g g = g = g \rho_g$ (identity elements are defined for each element in ***G***).
4. Every $g \in G$ has an inverse $g^{-1}$ satisfying $g^{-1}g = \rho_g$ and $gg^{-1} = \lambda_g$.

**Definition (Path of Locomotion, Leap, Locomotion Groupoid, Motion Groupoid)** Two steps are called *consecutive* if the end of the first step is the start of the second step. An ordered sequence of finite consecutive steps is called a *path*, the start of the first step is called the start of the path, the end of the last step is called the end of the path. The start point of a path, the ordered sequence of start points of intermediate steps, and the end point of a path are together called the *footprints of the path*. Two paths with common start and end points are called equivalent paths. Each equivalence class of paths is called a *leap*. The set of paths in the observation space with the groupoid algebra structure is called the *locomotion groupoid* on the observation space. The base of the locomotion groupoid is the observation space. The groupoid generated by the set of leaps is called the *motion groupoid*.

The *perspective on an observation domain* $\Omega$ defines a function on $\Omega \times \Omega$ that takes P-transitions as its values: $\mathcal{P} \times \mathcal{P}$: $(P,Q) \mapsto (\mathcal{P}(P), \mathcal{P}(Q))$, where $\mathcal{P}(P), \mathcal{P}(Q)$ are the perspectives at $P, Q$, respectively. Let $(\mathcal{P}(P), \mathcal{P}(Q))$ and $(\mathcal{P}(Q), \mathcal{P}(R))$ be two P-transitions, $P, Q, R$ are points in $\Omega$; with the composition law of perspective transitions $(\mathcal{P}(P), \mathcal{P}(Q)) \bullet (\mathcal{P}(Q), \mathcal{P}(R)) = (\mathcal{P}(P), \mathcal{P}(R))$, the perspective transition groupoid is isomorphic to the motion groupoid, since they have the same base space, one-to-one correspondence of their members, and homomorphism in their compositions. Let $\tau : \overrightarrow{PQ} \mapsto (\mathcal{P}(P), \mathcal{P}(Q))$ be the correspondence of the isomorphism, we may use $\tau(P,Q)$ to denote the P-transition correspondent to the motion vector $\overrightarrow{PQ}$: $\tau(P,Q) = \tau(\overrightarrow{PQ})$.

**Definition (Perspective Transition Groupoid)** The groupoid of P-transitions with base points in a domain of observation space $\Omega$ and with the above given composition law of the P-transitions is called the *perspective transition groupoid* on $\Omega$, denoted as $PTG(\Omega)$. For $\tau \in PTG(\Omega)$, $\alpha(\tau) \in \Omega$ is called the start position of $\tau$, $\beta(\tau) \in \Omega$ is called the end position of $\tau$, $\mathcal{S}(\alpha(\tau))$ is called the *start ray space* of $\tau$, $\mathcal{S}(\beta(\tau))$ is called the *end ray space* of $\tau$.

2.2. Representing 3D space through perspective transitions: 3D Embedding Theorem

**Definition (Stereo and Conjugate Sets of a P-Transition)** Let $(\mathcal{P}(P), \mathcal{P}(Q))$ be a P-transition, $\Sigma(\mathcal{P}(P))$ the set of source points of point projections in $\mathcal{P}(P)$ and $\Sigma(\mathcal{P}(Q))$ the set of source points of point projections in $\mathcal{P}(Q)$. A point projection $f$ in $\mathcal{P}(P)$ or in $\mathcal{P}(Q)$ is called a *stereo point projection* if its source $\sigma(f) \in \Sigma(\mathcal{P}(P)) \cap \Sigma(\mathcal{P}(Q))$. Two rays $r_p \in \mathcal{S}(P)$ and $r_q \in \mathcal{S}(Q)$ are *conjugate* to each other if they have the same source point in natural perspective; both $r_p$ and $r_q$ are called stereo rays. Sets of stereo rays in a ray space of a P-transition are called *stereo sets* of the P-transition. Two stereo sets are *conjugates* of each other if each ray in one set has its conjugate in the other set.

Poincare pointed out an obvious but profound fact: a perspective transition is defined on two ray spaces whose product constitutes a 4D space. This is what makes it possible for the perspective transitions to encode 3D space. These two ray spaces are able to simultaneously record the spatial locations of two rays each passing a perspective center. These joint events occupy a 4D space. A point in Euclidean space can project to two rays passing the two perspective centers. It is a joint event and can have a coordinate in the 4D transition space, just as any other joint events with two rays. But it is a special joint event: the two rays emanating from a point in space must lie in a plane, and the angle between the first and second rays is restricted to be larger than 0 and less than $\pi$ (**Fig. SM3**). A general joint event need not obey these restrictions. We will prove that these *coplanar* rays occupy a 3D submanifold of the 4D transition space, and the further restricted 3D submanifold is homeomorphic to the Euclidean space.

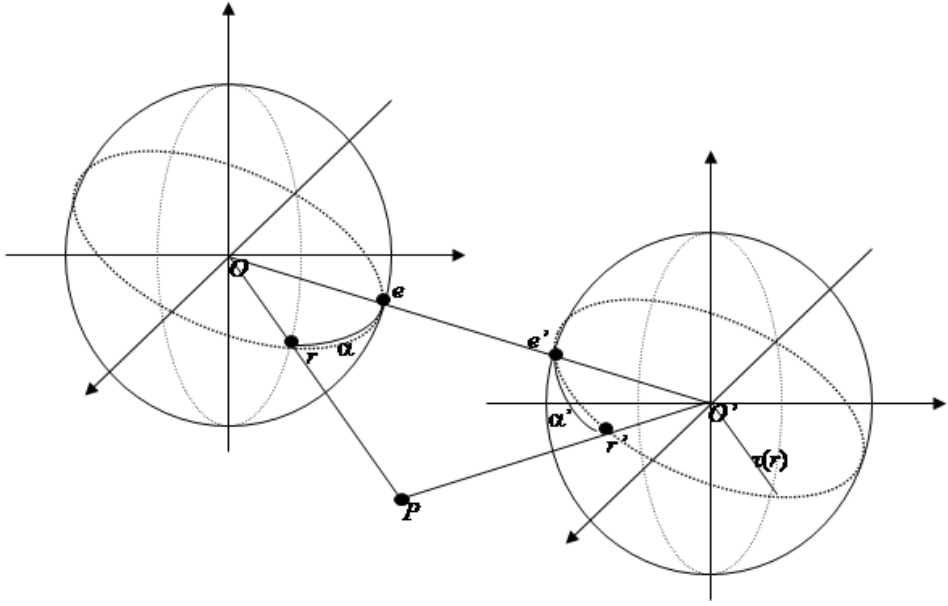

**Figure SM3.** The plane passing points *P, O, O'* is a *transition plane*, *e* and *e'* are two *transition poles*. The minor open arc on the transition circle linking ray *r* to the transition pole *e* is the *transition arc* α of the ray *r*. Ray *τ(r)* is the parallel transport of ray *r*. The *space-like conjugate arc* of ray *r* is the transition arc from *e'* to its parallel transport *τ(r)*.

Before going further, we first define some important geometric objects associated with perspective transitions.

**Definition (Transition Geometry of a Visual Space)** Given a visual space, an element of the motion groupoid on the base space of the visual space from point *O* to point *O'*, denoted as $\overrightarrow{OO'}$, is called a *transition vector*, or simply a *transition*. The Euclidean length of a transition vector is called the *distance of the transition*. The product *S(O)×S(O')* of the two ray spaces is called the *transition space* of the transition $\overrightarrow{OO'}$. The transition space of the null transition vector is a single ray space. The transition space *S(O')× S(O)* is called the *transpose transition space* of *S(O)× S(O')*. The point $(\theta_1, \phi_1, \theta_1', \phi_1') \in S(O)\times S(O')$ is the transpose point of $(\theta_2', \phi_2', \theta_2, \phi_2) \in S(O')\times S(O)$ if $(\theta_1, \phi_1) = (\theta_2, \phi_2)$ and $(\theta_1', \phi_1') = (\theta_2', \phi_2')$, and vice versa. A point in a transition space has two component rays, the source ray and the target ray. Both ray spaces are called ray spaces of the transition. Line $\overline{OO'}$ is called the *transition line*. A plane passing the transition line is called a *transition plane*. In terms of the unit sphere representation of the ray spaces, the transition geometry also includes: two *transition poles* which are the two points of the unit spheres of the two ray spaces that intersect the line segment $\overline{OO'}$, the *transition circles* which are intersections of the two ray spaces and transition planes, and the *transition arcs* which are the minor open arcs of the transition circles linking the rays to the transition poles. The transition plane, transition pole, transition circle, and transition arc of a null transition are null sets.

**Definition (Biprojection)** For each point *Q* in the Euclidean space other than the two points *O* and *O'*, the mapping from *Q* to the transition space *S(O)× S(O')* by way of point projections from *Q* to *S(O)* and *Q* to $S(O')$ is called a *point biprojection* from $Q$ to the 4D space $S(O) \times S(O')$. Let $D$ be a domain in the 3D Euclidean space, $O \in D$ and $O' \in D$, the mapping from $D$ to the 4D space $S(O) \times S(O')$ through point biprojections is called a *biprojection* from $D$ to $S(O) \times S(O')$.

We show in the 3D embedding theorem that the set of points in the 3D world and the subset of ray pairs in the 4D transition space under biprojection have the same topological structure. In other words, biprojection preserves the topological structure of the 3D world in the transition space. We call the image of the 3D Euclidean space in the transition space the stereo space. Each of its elements is called a stereo pair. This is important because it means the animal, through its two eyes or a movement of one eye, can directly access the topological structure of the 3D world if the signals from the two eyes are used in a combinatorial way, e.g., as inputs to binocular cells, rather than in a complementary way, e.g., as inputs to two classes of monocular cells. Whether the two 2D retinas span a 4D space or form a union of two 2D spaces, which is still a 2D topological space, depends on how they are used.

**Fig. SM3** shows the basic geometric objects in the transition space during locomotion.

Now we show that through biprojection, the topological structure of the 3D environment is represented in the transition spaces. The proper mathematical concept for describing "representation" is manifold *embedding*. The following definition is quoted from page 22 of (8).

**Definition** Let $\psi : M \rightarrow N$ be $C^\infty$.
   (a) $\psi$ is an *immersion* if $d\psi_m$ is non-singular for each $m \in M$.
   (b) The pair $(M, \psi)$ is a *submanifold* of $N$ if $\psi$ is a one-to-one immersion.
   (c) $\psi$ is an *embedding* if $\psi$ is a one-to-one immersion which is also a homeomorphism into; that is, $\psi$ is open as a map into $\psi(M)$ with the relative topology.
   (d) $\psi$ is a *diffeomorphism* if $\psi$ maps $M$ onto $N$ and $\psi^{-1}$ is $C^\infty$.

A diffeomorphism is a smooth homeomorphism.

**Definition (Parallel and Opposite Parallel Rays)** Let $S = S(O)$ and $S' = S(O')$ be the two ray spaces of a transition from $O$ to $O'$, $ray(\theta',\phi') \in S'$ is called *parallel* to $ray(\theta, \phi) \in S$ if in the coordinate system representation used in this article $(\theta',\phi') = (\theta,\phi)$, called *opposite parallel* if $(\theta',\phi') = ((\pi+\theta) \bmod 2\pi, \pi-\phi)$.

**Definition (Restricted Euclidean Space)** Given transition $\overrightarrow{OO'}$, the 3D manifold $(E^3 \setminus \overline{OO'})$ is called the *restricted Euclidean Space* of the transition.

**Definition (Stereo Mapping, Stereo Diffeomorphism, Vertex Set, Stereo Space)** Give transition $\overrightarrow{OO'}$, a mapping from an open set in ray space $S(O)$ to ray space $S(O')$ is a *stereo mapping* if every ray in the source set intersects with its image in the target ray space. A stereo mapping is called a *stereo diffeomorphism* if it is diffeomporphic. The intersection point is called a *vertex* of the mapping. A pair of rays is called a *stereo pair* if one is the target of the other under some stereo map. The vertex of a pair of rays with the same direction is a point in the sky. A stereo mapping determines a *vertex set*. The assignment of the vertex of two stereo rays gives the *inverse mapping* of a biprojection. The inverse mapping is called 3D recovery. The set of all stereo pairs in the transition space is called the *stereo subspace* of the transition space, or simply called the *stereo space* of the transition.

**Theorem (3D Embedding)** Given two ray spaces centered at $O$ and $O'$, the biprojection map from the restricted Euclidean Space $M = (E^3 \setminus \overline{OO'})$ to the transition space $N = S(O) \times S(O')$ is a manifold embedding.

**Proof**: First we show the biprojection from $M$ to $S(O) \times S(O')$ is an immersion at every point $(x, y, z) \in E^3 \setminus \overline{OO'}$. Without loss of generality, we assume the z-axis of reference frame $\mathcal{F} = (\mathbf{i}, \mathbf{j}, \mathbf{k}; O)$ passes point $O'$ (**Fig. SM4**). The coordinate of $O'$ in frame $\mathcal{F}(O) = (\mathbf{i}, \mathbf{j}, \mathbf{k}; O)$ is $(0, 0, l)$, $l = \|\overline{OO'}\|$. A point $Q \in M$ with $\mathcal{F}(O)$ coordinate $(x, y, z)$ has its $\mathcal{F}(O') = (\mathbf{i}, \mathbf{j}, \mathbf{k}; O')$ coordinate $(x, y, z-l)$. The biprojection mapping $\psi : (x, y, z) \mapsto (\theta, \phi, \theta', \phi')$ is given by the following formula:

$$\begin{cases} \theta = \cos^{-1}(x/\sqrt{x^2+y^2}) \\ \phi = \cos^{-1}(z/\sqrt{x^2+y^2+z^2}) \end{cases} \text{ and } \begin{cases} \theta' = \cos^{-1}(x/\sqrt{x^2+y^2}) \\ \phi' = \cos^{-1}((z-l)/\sqrt{x^2+y^2+(z-l)^2}) \end{cases}$$

We show at point $Q$, $d\psi_Q$ has rank 3. Now

$$d\psi_Q = \begin{bmatrix} \dfrac{-|y|}{x^2+y^2} & \dfrac{zx}{(x^2+y^2+z^2)\sqrt{x^2+y^2}} & \dfrac{-|y|}{x^2+y^2} & \dfrac{(z+l)x}{(x^2+y^2+(z+l)^2)\sqrt{x^2+y^2}} \\ \dfrac{x(sign(y))}{x^2+y^2} & \dfrac{zy}{(x^2+y^2+z^2)\sqrt{x^2+y^2}} & \dfrac{x(sign(y))}{x^2+y^2} & \dfrac{(z+l)y}{(x^2+y^2+(z+l)^2)\sqrt{x^2+y^2}} \\ 0 & \dfrac{-\sqrt{x^2+y^2}}{(x^2+y^2+z^2)\sqrt{x^2+y^2}} & 0 & \dfrac{-\sqrt{x^2+y^2}}{(x^2+y^2+(z+l)^2)\sqrt{x^2+y^2}} \end{bmatrix}$$

We first subtract the 3$^{rd}$ column by the 1$^{st}$ column. Since $x^2 + y^2 > 0$, either $x \neq 0$ or $y \neq 0$. If $y \neq 0$, we multiply the second row by $-x/y$ and add to the first row, so that we have the matrix

$$\begin{bmatrix} \dfrac{-sign(y)(x^2+y^2)}{(x^2+y^2)y} & 0 & 0 & 0 \\ \dfrac{x(sign(y))}{x^2+y^2} & \dfrac{zy}{(x^2+y^2+z^2)\sqrt{x^2+y^2}} & 0 & \dfrac{(z+l)y}{(x^2+y^2+(z+l)^2)\sqrt{x^2+y^2}} \\ 0 & \dfrac{-\sqrt{x^2+y^2}}{(x^2+y^2+z^2)\sqrt{x^2+y^2}} & 0 & \dfrac{-\sqrt{x^2+y^2}}{(x^2+y^2+(z+l)^2)\sqrt{x^2+y^2}} \end{bmatrix}$$

Now

$$\begin{vmatrix} \dfrac{zy}{(x^2+y^2+z^2)\sqrt{x^2+y^2}} & \dfrac{(z+l)y}{(x^2+y^2+(z+l)^2)\sqrt{x^2+y^2}} \\ \dfrac{-\sqrt{x^2+y^2}}{(x^2+y^2+z^2)\sqrt{x^2+y^2}} & \dfrac{-\sqrt{x^2+y^2}}{(x^2+y^2+(z+l)^2)\sqrt{x^2+y^2}} \end{vmatrix}$$

$$= \frac{ly}{(x^2+y^2+z^2)(x^2+y^2+(z+l)^2)\sqrt{x^2+y^2}} \neq 0.$$

If $x \neq 0$, we multiply the first row by $-y/x$ and add to the second row, so that we have the matrix

$$\begin{bmatrix} \frac{-|y|}{x^2+y^2} & \frac{zx}{(x^2+y^2+z^2)\sqrt{x^2+y^2}} & 0 & \frac{(z+l)x}{(x^2+y^2+(z+l)^2)\sqrt{x^2+y^2}} \\ \frac{(x^2+y^2)(sign(y))}{x(x^2+y^2)} & 0 & 0 & 0 \\ 0 & \frac{-\sqrt{x^2+y^2}}{(x^2+y^2+z^2)\sqrt{x^2+y^2}} & 0 & \frac{-\sqrt{x^2+y^2}}{(x^2+y^2+(z+l)^2)\sqrt{x^2+y^2}} \end{bmatrix}$$

Now $\begin{vmatrix} \frac{zx}{(x^2+y^2+z^2)\sqrt{x^2+y^2}} & \frac{(z+l)x}{(x^2+y^2+(z+l)^2)\sqrt{x^2+y^2}} \\ \frac{-\sqrt{x^2+y^2}}{(x^2+y^2+z^2)\sqrt{x^2+y^2}} & \frac{-\sqrt{x^2+y^2}}{(x^2+y^2+(z+l)^2)\sqrt{x^2+y^2}} \end{vmatrix}$

$$= \frac{lx}{(x^2+y^2+z^2)(x^2+y^2+(z+l)^2)\sqrt{x^2+y^2}} \neq 0.$$

In either case, $d\psi_Q$ has rank 3. Therefore point $(x, y, z)$ is a regular point. If two regular points have the same image in $N$ under biprojection, they must be the same point. This is because if two rays intersect, that have only one intersection point. Thus biprojection is injective. Now we show the inverse mapping of biprojection from $N = \psi(M)$ to $M$, $\psi^{-1} : N \to M$, is continuous when $N$ is given the relative topology of the transition space. Using the same coordinate system as before, let $P_0 = (x_0, y_0, z_0) \in E^3 \setminus \overline{OO'}$ be a point of the restricted Euclidean space, and the point $(\theta_0, \phi_0, \theta_0', \phi_0') \in S(O) \times S(O')$ its image in the transition space under biprojection. In this coordinate representation, we have $\theta_0 = \theta_0'$, and $0 < \phi_0 < \phi_0' < \pi$. In terms of transition space coordinates, the vertex point has the following Euclidean space coordinates:

$$\begin{cases} x = \dfrac{l \sin\phi' \sin\phi \cos\theta}{\sin(\phi'-\phi)} \\ y = \dfrac{l \sin\phi' \sin\phi \sin\theta}{\sin(\phi'-\phi)} \\ z = \dfrac{l \sin\phi' \cos\phi}{\sin(\phi'-\phi)} \end{cases},$$

for all transition space points $(\theta, \phi, \theta', \phi') \in S(O) \times S(O')$, where $\theta = \theta'$, $0 < \phi < \phi' < \pi$. This is a continuous function from the image points under biprojection in the transition space to the preimage points of the restricted Euclidean space. Therefore the biprojection mapping from $M$ to $N$ is a homeomorphism and $\psi$ is an embedding. *QED*.

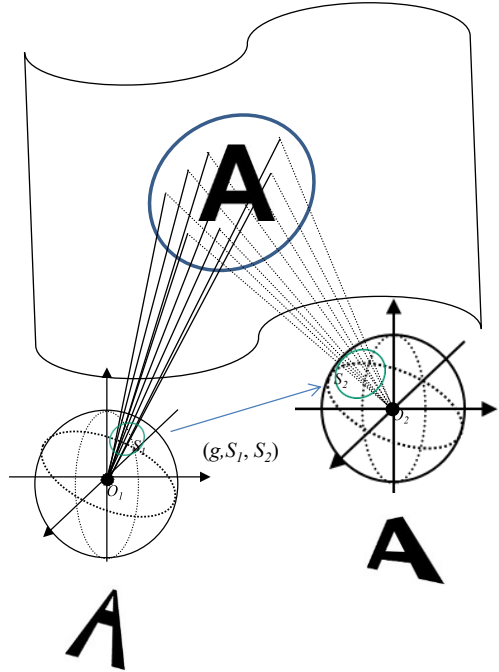

**Figure SM4.** The surface with label A is mapped into the stereo space of the transition $\overrightarrow{O_1O_2}$ as the image of a diffeomorphism $(g, S_1, S_2)$, which is a 2-D manifold in the transition space $S(O_1) \times S(O_2)$. There are infinite number of such copies of the surface with label A in the visual space.

2.3. Surface representations in transition spaces

2.3a. Coding the topology of 3D surfaces: Surfaciness Theorem

How can we see surfaces from ambient lights? First of all, we need to clarify the meaning of seeing surfaces. A surface is not just a *point set*. It is a topological structure. Gibson created the term *surfaciness* to indicate something with *spatial presence* and with *continuous extension* like a sheet. Many vision researchers assume the topological quality of being a surface is imposed by the human mind (a 3D surface *"interpretation"* of the retinal image) based upon some discrete depth maps obtained from feature point correspondences between binocular (or motion) stereo pairs. In Gibson's view, surfaciness is *directly seen* from the ambient lights and is not the result of a subsequent process of data interpolation. He suggested the information for surfaciness can be found in the *transformations* between optic arrays: "the available information in the optic array for *continuity* could be described as the *preservation of adjacent order*."

We have shown that biprojection from the 3D Euclidean space to the 4D transition space is a topological embedding. Surfaces are 2D submanifolds in 3D Euclidean space. The surface representation problem then is a part of the 3D embedding problem: during 3D embedding, a 2D submanifold of the 3D Euclidean space is also embedded in the 4D transition space, in the 3D stereo subspace. The image points in the transition space resulting from the 3D embedding are constrained by stereo diffeomorphism. The points in the transition space from the 3D embedding of a surface are characterized by an additional constraint: each ray has a *unique* conjugate ray, and the correspondence mapping between conjugate rays is a homeomorphism; we prove this in the *stereo mapping theorem*. The sets in the transition space constrained by a stereo map specify the surface; we prove this in the *surfaciness theorem*.

**Theorem (Surfaciness)** Let $O$ and $O'$ be two points in observation space $\Omega$, $g$ a diffeomorphism: $g: S_o \to S_{o'}$ from a domain in the ray space $S(O)$ to a domain in the ray space $S(O')$: $S_o \subset S(O)$, $S_{o'} \subset S(O')$. The set $M = \{(\theta,\phi,\theta',\phi') | (\theta,\phi) \in S_o, (\theta',\phi') \in S_{o'}, \text{ and } (\theta',\phi') = g(\theta,\phi)\}$ is a 2D manifold in the transition space. If $g$ is a stereo mapping and $M$ is a stereo set of $S(O) \times S(O')$, the biprojection $\psi$ from space $E^3 \setminus \overrightarrow{OO'}$ to the transition space of the transition $\overrightarrow{OO'}$ has inverse on $M$, and $\psi^{-1}(M)$ is a surface.

**Proof:** A *diffeomorphism* : $g: S_o \to S_{o'}$ can be written as follows:
$$\begin{cases} \theta' = g_1(\theta,\phi) \\ \phi' = g_2(\theta,\phi) \end{cases}, \text{ where } (\theta,\phi) \in S_o, (\theta',\phi') \in S_{o'}.$$

Let $\begin{cases} F_1(\theta,\phi,\theta',\phi') = \theta' - g_1(\theta,\phi) \\ F_2(\theta,\phi,\theta',\phi') = \phi' - g_2(\theta,\phi) \end{cases}$, $F(\theta,\phi,\theta',\phi') = \begin{pmatrix} F_1(\theta,\phi,\theta',\phi') \\ F_2(\theta,\phi,\theta',\phi') \end{pmatrix}$ is a mapping from the 4D submanifold $N = S_o \times S_{o'}$ of the transition space into a 2D Euclidean space $R^2$.

$DF = \begin{pmatrix} -\partial g_1/\partial \theta & -\partial g_1/\partial \phi & 1 & 0 \\ -\partial g_2/\partial \theta & -\partial g_2/\partial \phi & 0 & 1 \end{pmatrix}$ has rank of 2. From the preimage theorem, $F = \begin{pmatrix} 0 \\ 0 \end{pmatrix}$ cuts a 2D submanifold $M = \{(\theta,\phi,\theta',\phi') | (\theta,\phi) \in S_o, (\theta',\phi') \in S_{o'}, \text{ and } (\theta',\phi') = g(\theta,\phi)\}$ in the 4D manifold $N = S_o \times S_{o'}$. When $g$ is a stereo diffeomorphism, $M$ is a subset in the 3D stereo space. The vertex set from 3D recovery is homeomorpic to $M$, and therefore a 2D manifold in the Euclidean space, a surface. *QED*.

The theorem shows the environment surfaces are locally specified by the diffeomorphisms between the ray spaces in a perspective transition. In the proof of the surfaciness theorem, we have shown that a *triple* ($g$, $S1$, $S2$) specifies *a 2D submanifold* of the 4D manifold $S1 \times S2$. Thus a surface is not specified by a set of points. *A surface is specified by a function, a stereo diffeomorphism which maps each source ray to its stereo conjugate.*

Note that in real life, the coplanar diffeomorphism $g$ need not be specified exactly. A visual representation is always an approximation, and an approximation to $g$ can be given by a finite order Taylor approximation. This method of surface approximation is very different from the conventional approach to approximate surfaces by *depth maps*. The triple representation directly specifies the topological quality of being a 2D manifold (*surfaciness*) as well as the geometric shape and location of the surface. These three pieces of information are what the vision system needs for surface representation. Neither shape nor topology can be conveyed by a depth map: a set of discrete 3D points possesses no 2D manifold topological structure nor any shape.

2.3b. Surfaces are invariants of perspective transformations: Fixed Vertex Theorem

We have just shown that in a given *P*-transition, a stereo diffeomorphism between domains in two ray spaces provides a code for a surface in the Euclidean space. Now we further demonstrate that any environment surface can be locally represented by stereo diffeomorphisms. Furthermore, for a local surface, the set of all the different diffeomorphisms of a local surface as taken from different *P*-transitions form an equivalence class and represent the same local surface.

We now define a topology for the visual space with which the mappings representing the surface are defined. The simplest way to make such a topological space is to take the topological sum of all the ray spaces.

**Definition (Ray Space Topology of Visual Space)** Let $VS(\Omega) = S \times \Omega$ be the visual space over observation

domain $\Omega$, and $\mathcal{T}(P)$ the topology (family of all open sets) of ray space $\mathcal{S}(P)$. $\mathcal{B}(\Omega) = \bigcup_{P\in\Omega} \mathcal{T}(P)$ is called the *bundle of the ray space topologies* $\mathcal{S}(P)$. For $N\in\bigcup_{P\in\Omega}\mathcal{T}(P)$ the point $P\in\Omega$ such that $N\in\mathcal{T}(P)$ is called the *base point* of $N$, and denoted as $\pi_\Omega(N)$. The topological sum $\mathcal{T}(\Omega) = \coprod_{P\in\Omega}\mathcal{T}(P)$ is called the *ray space topology of the visual space* $\mathcal{VS}(\Omega)$; it is larger than $\mathcal{B}(\Omega)$, including the union of neighborhoods in different ray spaces.

**Definition (Mapping Triples, Subtriples of a Mapping Triple)** Let $S_1$ and $S_2$ be two open sets in the ray space topology of $\mathcal{VS}(\Omega)$, $g: S_1 \to S_2$ a diffeomorphism from $S_1$ to $S_2$, the 2D submanifold of $S_1 \times S_2$ constrained by mapping $g$ according to the surfaciness theorem is called a *mapping triple* and denoted by $\mu\tau(g, S_1, S_2)$. It is also called a *triple*. A mapping triple $\mu\tau(h_{PQ}, T_1, T_2)$ is called a *subtriple* of the mapping triple $\mu\tau(g_{PQ}, S_1, S_2)$ if $(h_{PQ}, T_1, T_2)$ is a *restriction* of $(g_{PQ}, S_1, S_2)$, i.e., $T_1 \subseteq S_1$, $T_2 \subseteq S_2$ and $h_{PQ} = g_{PQ}|_{T_1}$.

Can an animal's vision system respond selectively only to the triples from the stereo space? If so, then the vision system will be selective for surfaces in space. The brain with such a neural wiring would have a Kantian mind possessing the *a priori* form of seeing surfaces in space.

While the surfaciness theorem shows that stereo diffeomorphisms determine surfaces, it is possible that for a local surface there may not be a stereo diffeomorphism in the visual space serving as its representation. The next theorem shows that surfaces do actually generate stereo diffeomorphisms; therefore all visible surfaces are coded in the form of triples. A basic mathematical concept used here is the *regular mappings* (diffeomorphisms) between different ray spaces.

**Theorem (Existence of Stereo Diffeomorphism)** If a domain of the visual environment is stably visible from the two base points of a transition, then there is a stereo diffeomorphism between the ray spaces with the visible domain as its vertex set.

**Proof:** Let $\mathcal{VA}(P_1, P_2)$ be a domain of the visual environment, $f_1: \mathcal{VA}(P_1, P_2) \to \mathcal{S}(P_1)$ and $f_2: \mathcal{VA}(P_1, P_2) \to \mathcal{S}(P_2)$ perspective projections. Suppose there is some singular point of the projections $f_1$ or $f_2$, then with arbitrarily small perturbation, the singularity will degenerate into folds possibly accompanied by cusps (Arnold 1981, Whitney's theorem on plane-to-plane mappings), and no matter how small the perturbation, the surface area must include some part not visible after the perturbation because of the existence of a fold. Thus the domain is not stably visible, contradicting our hypothesis. This means the domain must be regular at every point on $\mathcal{VA}(P_1, P_2)$. Now for each regular point $p$, there is a neighborhood $N(p) \subseteq \mathcal{VA}(P_1, P_2)$ on which $f_1$ and $f_2$ are diffeomorphisms to neighborhoods in the range of $f_1$, $\mathcal{RG}(f_1)$, and neighborhoods in the range of $f_2$, $\mathcal{RG}(f_2)$, respectively. Therefore the mapping from ray space $\mathcal{S}(P_1)$ to ray space $\mathcal{S}(P_2)$ given by

$$g = f_2 \circ f_1^{-1}: \mathcal{RG}(f_1) \to \mathcal{RG}(f_2)$$

is a diffeomorphism. $\mathcal{RG}(f_1)$ and $\mathcal{RG}(f_2)$ are a pair of stereo domains. *QED*.

**Definition (Perspective Mapping, Perspective Surface)** Given a *P*-transition, the stereo diffeomorphism $g$ from a stereo domain to its stereo conjugate as described in the above theorem is called a *natural perspective mapping* or *perspective transformation*. The stably visible surface area is called the *perspective surface* of the perspective mapping.

Actually, from the surfaciness theorem, all the stereo diffeomorphisms are perspective mappings associated with the perspective surface they determined. Later on we do not distinguish the concepts of *perspective mapping* and *stereo diffeomorphism*. The counterpart of the concept of perspective mapping in classic projective geometry is also called *perspective transformation*. They are related but different concepts. In **Fig. SM5a**, correspondences of lines in different pencils in a *projective space* are made by joint incidence to points on a plane called the perspective plane; in **Fig. SM5b**, correspondences of rays in ray spaces based at different points of observation are made by joint incidence to points on a non-flat surface, the perspective surface. In classic projective geometry, a perspective plane can be determined by three pairs of correspondent lines with intersection points not on one line. Determining a perspective surface requires a *stereo diffeomorphism*.

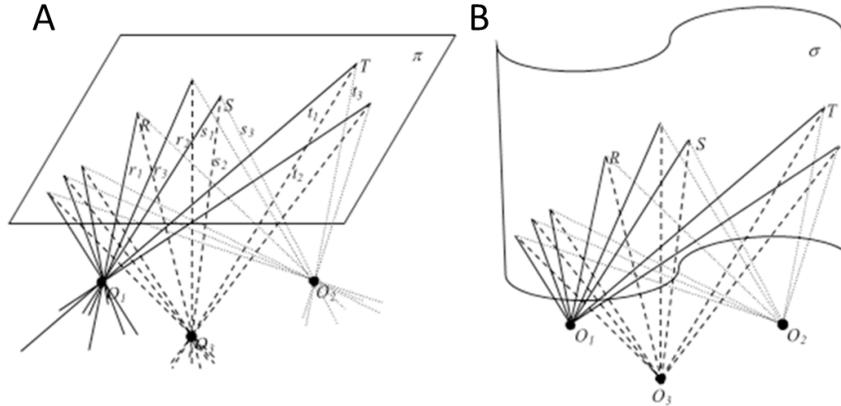

**Figure SM5.** (**A**) A classic perspective transformation between a pair of pencils of lines given by a perspective plane $\pi$ maps lines $r_1, s_1, t_1$ of the line pencil at $O_1$ to lines $r_2, s_2, t_2$ of the line pencil at $O_2$ through incidence to points $R, S, T$ on the perspective plane $\pi$. The plane determines a perspective transformation between any pair of line pencils in the 3D projective space, such as at points $O_1$ and $O_2$. Three pairs of correspondent lines with their intersection points not on one line determine the perspective plane, thereby the perspective transformation. (**B**) A natural perspective transformation between rays in different ray spaces is given by a surface through incidence to common points on the surface $\sigma$. The surface determines a perspective transformation between ray spaces based in the observation area.

Having shown that environment surfaces can be represented by perspective mappings, we now show the surface coded by a sequence of perspective mappings is fixed in the Euclidean space, i.e., a visible surface is an invariant of its representations in the form of perspective mappings. To prove this, we first introduce the concept of an algebraic structure of regular mappings in the visual space with the ray space topology. We have already used the concepts of *regular* and *critical* points. A formal definition is quoted below.

**Definition 1.11** (from pg 33-34, (9)). Let $X$ and $Y$ be differentiable manifolds and $f: X \rightarrow Y$ a $C^1$-mapping. Then
   (1) corank $(df)_P = \min(\dim X, \dim Y) - \text{rank}(df)_P$.
   (2) a point $p \in X$ is a critical point of $f$ if corank $(df)_P > 0$.
   (3) the set of critical points of $f$ is denoted by $C[f]$.
   (4) a point $q \in Y$ is a critical value of $f$ if $q \in f(C[f])$.
   (5) a point $p \in X$ is a regular point of $f$ if $p \notin C[f]$.
   (6) a point $p \in X$ is a regular value of $f$ if it is not a critical value of $f$. So, in particular, a point not in Image($f$) is a regular value.

The perspective transformations are mappings between different open sets in the ray space topology of the visual space. With the ray space topology of visual space, we can now define the totality of *perspective transformations* for the visual space as a special groupoid called a *pseudogroup* (9, 10). The definition of a pseudogroup is quoted here from Kobayashi and Nomizu (10):

**Definition (Pseudogroup of Transformations)** A *pseudogroup of transformations* on a topological space $S$ is a set $\Gamma$ of transformations satisfying the following axioms:
(1) Each $f \in \Gamma$ is a homeomorphism of an open set (called the domain of $f$) of $S$ onto another open set (called the range of $f$) of $S$;
(2) If $f \in \Gamma$, then the restriction of $f$ to an arbitrary open subset of the domain of $f$ is in $\Gamma$;
(3) Let $U = \bigcup_i U_i$ where each $U_i$ is an open set of $S$. A homeomorphism $f$ of $U$ of an open set of $S$ belongs to $\Gamma$, if the restriction of $f$ to $U_i$ is in $\Gamma$ for every $I$;
(4) For every open set $U$ of $S$, the identity transformation of $U$ is in $\Gamma$;
(5) If $f \in \Gamma$, then $f^{-1} \in \Gamma$;
(6) If $f \in \Gamma$ is a homeomorphism of $U$ onto $V$ and $f' \in \Gamma$ is a homeomorphism of $U'$ onto $V'$ and $V \cap U'$ is non-empty, then the homeomorphism $f' \circ f$ of $f^{-1}(V \cap U')$ onto $f'(V \cap U')$ is in $\Gamma$.

**Definition (Global Transition Mapping Function, Transition Mappings of Visual Space)** Let $\mathcal{VS}(\Omega) = S \times \Omega$ be the visual space over observation domain $\Omega$, $\tau\mu$ a function defined on $\Omega \times \Omega$ by assigning each $(P,Q)$, $P \neq Q$, a *mapping triple* of the visual space: $\tau\mu(P,Q) = \mu\tau(g_{PQ}, S1, S2)$, where $S1, S2 \in \mathcal{B}(\Omega)$, $\pi_\Omega(S1) = P$, $\pi_\Omega(S2) = Q$, and $g_{PQ}: S1 \to S2$ a diffeomorphism, with the requirement that $\tau\mu(Q,P) = \tau\mu^{-1}(P,Q)$, the inverse mapping of $\tau\mu(P,Q)$, and $\tau\mu(P,P) = 1_{S(P)}$, the identity (or trivial) mapping for all $P \in \Omega$. This function $\tau\mu$ is called a global *transition mapping function* on the visual space. Each of its values is called a *global transition mapping*, denoted as a *GT-mapping*. A restriction of a global transition mapping is called a *transition mapping* with respect to the *GT*-mapping function of the visual space, or simply a *T-mapping*. The transition vector $\overrightarrow{PQ}$ is called the *transition* of the *T*-mapping. We can think of the global transition mapping function as a function defined on the set of transitions on the observation space.

**Definition (Consecutive Transition Mappings and their Composition)** Two *T*-mappings are consecutive if the set of targets of the first mapping is the set of sources of the second mapping. Given two consecutive *T*-mappings $h_{PQ}: SD \to SM$ and $h_{QR}: SM \to ST$, where $SD$, $SM$, and $ST$ are open sets in ray spaces, $\pi_\Omega(SD) = P$, $\pi_\Omega(SM) = Q$, and $\pi_\Omega(ST) = R$, the mapping $h_{PQR} = h_{QR} \circ h_{PQ}: SD \to ST$ is called the composition of the two consecutive *T*-mappings.

Although with two consecutive mappings we can always properly define their composition from one ray space to another ray space, there is no guarantee their composition is also a *T*- mapping with respect to the same *GT*-mapping function.

**Definition (Motion Consistent Global Transition Mapping Function)** Let $\tau\mu$ be a *GT*- mapping function. If for any three points $P, Q, R$ in the base space $\Omega$ of the visual space, and three transition mappings with respect to $\tau\mu$, $h_{PQ}: SD \to SM$, $h_{QR}: SM \to ST$, $h_{PR}: SD \to ST$, where $h_{PQ}$, $h_{QR}$ are consecutive, the following diagram commutes, we call the global transition mapping function *motion consistent*.

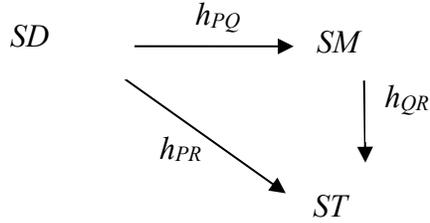

The set of *T*-mappings with respect to a *GT*-mapping function in the visual space satisfies the axioms of pseudogroup (1), (2), (4), and (5). The requirement that compositions of *T*-mappings be *motion consistent* guarantees axiom (6) is observed. The *axioms of topology* require the union of a family of open sets be included in the topology. The union of open sets across ray spaces is included in our *ray space topology*. As a consequence, maps from a source open set distributed in different ray spaces to a target open set distributed in different ray spaces are included in the mapping structure, therefore satisfying axiom 3. Such mappings are not *T*-mappings, which happen between two ray spaces. However, their existence in our pseudogroup will not cause any inconsistency with our statements about *T*-mappings.

**Definition (A Mapping Structure of the Visual Space)** The pseudogroup generated by the value set of a motion consistent *GT*-mapping function of the visual space with the ray space topology is called the *mapping structure* of the visual space imposed by the *GT*-mapping function, denoted by $MS(\Omega, f)$, where $\Omega$ is the observation space and *f* is the *GT*-mapping function. A mapping structure $MS(\Omega, f)$ is called a *stereo mapping structure* if each of its *T*-mappings is a stereo mapping.

**Definition** (**Perspective Mapping Structure of a Visual Space**) The mapping structure of the visual space generated by the set of natural perspective mappings on the visual space is called the *perspective mapping structure* of the visual space. The perspective mapping structure is the pseudogroup of all stereo diffeomorphisms in the visual space imposed by environment surfaces.

Now we will show the surface recovered from a stereo diffeomorphisms is invariant within the equivalence class defined by the perspective mapping structure of the visual space.

**Definition (Equivalence Class with respect to a Mapping Structure)** Given a mapping structure of visual space, two members of $\mathcal{B}(\Omega)$ are *equivalent* with respect to the mapping structure if there is a *T*-mapping mapping one member to the other. The equivalence classes thus formed are called *equivalence classes* with respect to the mapping structure.

**Definition (Stable Open Set under a Mapping Structure, Locally Stable Mapping Structure)** Let *MS* be a mapping structure in a visual space, *S* a member of $\mathcal{B}(\Omega)$. *S* is *stable* if the *base points* of members of the equivalence class of *S* with respect to the mapping structure contain a neighborhood of the base point of *S*. *S* is called *locally stable* if each ray in *S* is contained in a stable member of $\mathcal{B}(\Omega)$. A mapping structure is *locally stable*, if each of its members has a locally stable source.

**Theorem (Fixed Vertex Set)** Given a locally stable space-like coplanar mapping structure of a visual space, each equivalence class determined by this mapping structure determines a fixed vertex set.

**Proof:** Within each equivalence class, the compositions of transition mappings are motion consistent (by definition of mapping structure). Consider a ray *a* in ray space $S(O_1)$ that has its image ray *b* in ray space $S(O_2)$, which in turn has its image ray *c* in ray space $S(O_3)$. Suppose $O_1, O_2,$ and $O_3$ are not along one line. We can assume this due to local stability. The maps are motion consistent, therefore *a* has its image ray *c* in ray space S($O_3$). This means rays *a* and *b* are in a plane $m_1$, rays *b* and *c* are in a plane $m_2$, and rays *c* and *a* are in a plane $m_3$. Because the mapping structure is space-like, the intersection point $p_1$ of *a* and *b* exists and is on $m_1$, $m_2$, and $m_3$. Similarly the intersection point $p_2$ of *b* and *c* exists and is on $m_1$, $m_2$, and $m_3$, and the intersection point $p_3$ of *a* and *c* exists and also is on $m_1$, $m_2$, and $m_3$. Since $p_1$, $p_2$, $p_3$ are each on $m_1$, $m_2$, and $m_3$, we must have $p_1 = p_2 = p_3$, since three planes intersect in at most one point. Therefore the trajectory of a domain in a ray space has an invariant attached vertex set. *QED.*

We now show that the set of regular values of ambient perspective projection is locally stable under the perspective mapping structure, thus completing our demonstration that the *P*-mapping structure provides codes for local environment surfaces, and surfaces are invariant to the coding variations through *P*-transitions.

**Theorem (Local Stability of the Regular Value Set)** The set of regular values of a natural perspective projection is locally stable under the perspective mapping structure.

**Proof:** Let ray *x* be a regular value of a perspective projection from a surface to the ray space $\mathcal{S}(O)$, *f*(. ; *O*): $x = f(P)$, *P* the preimage of *x* on the surface. Since *x* is a regular value, there is a neighborhood $N(P)$ of the point *P* on the surface, and a neighborhood $N(x)$ in the ray space such that the map

$$f(.;O): N(P) \to N(x)$$

is a diffeomorphism. Now that $N(P)$ is a regular domain of the perspective projection to *O*, the determinant of *Df* does not vanish on $N(P)$. Let $N'(P)$ be a surface extension of $N(P)$ containing the closure of $N(P)$. We may assume $N'(P)$ is also a regular domain with compact closure, and $\det(Df(X;O)) \geq \varepsilon > 0$ for any point $X \in N'(P)$, where $\det(Df(X;O))$ is the determinant of the derivative of the perspective projection at surface point *X*. Referring to **Fig. SM6**, *C* is a cone with vertex *O* and base surface $\overline{N}(P)$, the closure of $N(P)$. We can cover each point of $\overline{N}(P)$ with an open set in 3D Euclidean space such that it contains no points of environment surfaces other than those of $N'(P)$, since $\overline{N}(P)$ is visible from *O*. and each point of the cone surface of *C* with an open set in the observation space. By first choosing a finite open cover *Co* for the cone surface *C* we can enlarge cone *C* to a cone *C'* within the scope of *Co* with same base $\overline{N}(P)$ and containing point *O* inside. Unlike cone *C*, cone *C'* may contain points of $N'(P)$ inside, because it is made of the open covers for the base surface $\overline{N}(P)$. Let $N(O)$ be a neighborhood of *O* inside the cone *C'*. We claim there is a neighborhood ball $N_\delta(O)$ centered at *O* with radius $\delta$ in the observation space, $N_\delta(O) \subseteq N(O)$, such that $N(P)$ is visible from any point in $N_\delta(O)$. If this is true, we reach the conclusion of the theorem. Suppose the contrary is true, then for any $\delta(i) = \delta/2^i$, there is some point $O_i \in N_{\delta(i)}(O)$ such that a point $P_i \in N(P)$ is occluded by some surface point $Q_i \in N'(P)$ inside the cone *C'*. Linking points $P_i$ and $Q_i$ by a path in $N'(P)$, there must be a critical point $T_i \in N'(P)$ of the perspective projection $f(.,O_i)$, and $\det(Df(T_i;O_i)) = 0$. Let $i \to \infty$, we can choose a subsequence of $\{T_i\}$ converging to a point $T_0 \in N'(P)$. The sequence of (3) converges to *O*. The function $\det(Df(X;Y))$ is a continuous function with respect to *Y* and *X*. Therefore $\det(Df(T_0, O)) = 0$. A contradiction shows that such a $N_\delta(O) \subseteq N(O)$ must exist. *QED.*

From Sard's theorem, which states that the image of the set of critical points of a smooth function *f* from one manifold to another has measure 0, it is easy to deduce that almost all points in a ray space are regular values under natural perspective projection. This local stability theorem further demonstrates for each regular value there is a stable neighborhood. Therefore, at almost every ray in a ray space, there is a stable neighborhood. The perspective mapping structure is locally stable.

The vertex set computation described in the proof of the Fixed Vertex Set Theorem is a fundamental computation for 3D vision. It is what makes possible explicit perception of the distance of points on environment surfaces. It is only possible through explicit representation of *visual rays*, direction vectors linking an observation point behind the image plane to points on the image plane. It is the rays that reach the third dimension from the 2D retina. This is what Gibson meant when he emphasized that the eyes sense optic arrays, not retinal images; even when looking at a 2D picture, the eyes sense the optic array generated by lights starting from the 2D picture.

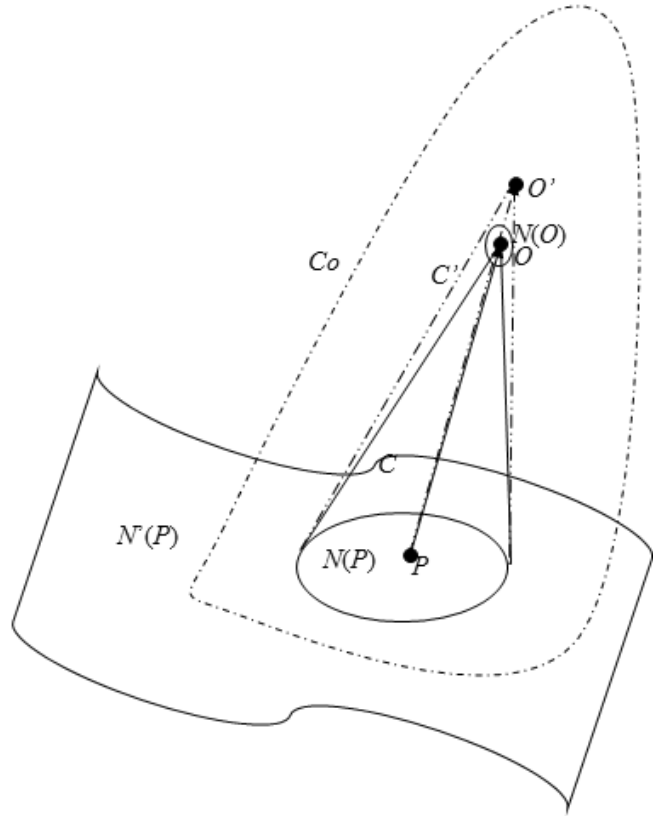

**Figure SM6**. The cone $C$ with base $\overline{N}(P)$, the closure of a neighborhood of $P$ on the surface $S$, as the base and the point $O$ as the vertex. Since $N(P)$ is visible from $O$, the interior of $C$ contains no points of environment surfaces. Cone $C$ has a finite open cover $Co$ with a boundary of some minimum positive distance away from cone $C$. Cone $C'$ enlarges $C$ within the scope of $Co$ containing $O$ inside. $C'$ may contain part of surface $N'(P)$ inside.

2.3c. Rigid shape from perspective mappings

So far we have shown that the *triples* of the *perspective mapping structure* in the same equivalence class specify a surface patch fixed in 3D space, and thus provide a representation for the topology of the surface. A rigid surface also has a *geometric shape*, i.e., a specific numeric characterization of the geometric construction, the curving, of the surface. Now we further show that the geometric shape of a surface is also specified by the perspective mapping structure.

In this section, we use a simple "pinhole" camera model for representing rays in a ray space, and a translation of the observation point along the Y-axis by one unit ("binocular setting") as the coordinate representation of the $P$-transition. **Fig. SM7** represents our pinhole camera model and **Fig. SM8** represents the binocular setting (ambient optic arrays at two observation points) with *convergent point* located at infinity.

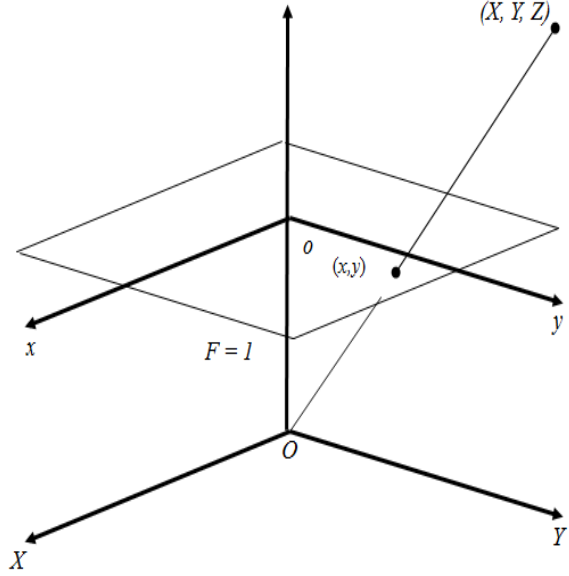

**Figure SM7.** An environment-associated coordinate system *O-XYZ* and an image plane coordinate system *o-xy* for a pinhole camera model are attached in a fixed geometric relation.

Using the image plane coordinate system and referring to **Fig. SM8**, a perspective mapping from domain $S_h(O)$ (the image plane for the half ray space centered at *O*) to domain $S_h(O')$ (the image plane for the half ray space centered at *O'*):

$$g: p \mapsto p' = g(p); p \in U, p' \in U'$$

can be written in terms of image coordinates: $\begin{pmatrix} x' \\ y' \end{pmatrix} = \begin{pmatrix} g_1(x,y) \\ g_2(x,y) \end{pmatrix}$,

Here $\begin{pmatrix} x \\ y \end{pmatrix}$ is the image plane coordinate representation of ray *p*, and $\begin{pmatrix} x' \\ y' \end{pmatrix}$ is the image plane coordinate representation of ray *p'*, and $\begin{pmatrix} g_1 \\ g_2 \end{pmatrix}$ are two components of the perspective mapping *g*.

The first order Taylor expansion of *g* at point $p_0$ with image plane coordinate $(x_0, y_0)$ has image plane coordinate representation:

$$\begin{pmatrix} x' \\ y' \end{pmatrix} = \begin{pmatrix} g_1(x,y) \\ g_2(x,y) \end{pmatrix} = \begin{pmatrix} g_1(x_0,y_0) \\ g_2(x_0,y_0) \end{pmatrix} + \begin{bmatrix} \frac{\partial x'}{\partial x} & \frac{\partial x'}{\partial y} \\ \frac{\partial y'}{\partial x} & \frac{\partial y'}{\partial y} \end{bmatrix} \begin{pmatrix} \Delta x \\ \Delta y \end{pmatrix} + o(\|\Delta\|^2),$$

where $\Delta = \begin{pmatrix} \Delta x \\ \Delta y \end{pmatrix} = \begin{pmatrix} x - x_0 \\ y - y_0 \end{pmatrix}$.

The translation vector $\begin{pmatrix} \delta_x \\ \delta_y \end{pmatrix} = \begin{pmatrix} g_1(x_0, y_0) - x_0 \\ g_2(x_0, y_0) - y_0 \end{pmatrix}$ is the "disparity" between corresponding centers of the two respective neighborhoods, the matrix $\begin{pmatrix} a_{11} & a_{12} \\ a_{21} & a_{22} \end{pmatrix} = \begin{bmatrix} \frac{\partial x'}{\partial x} & \frac{\partial x'}{\partial y} \\ \frac{\partial y'}{\partial x} & \frac{\partial y'}{\partial y} \end{bmatrix}_{(x_0, y_0)}$ is the derivative $dg: S_h(O)_{p_0} \to S_h(O')_{g(p_0)}$, representing a mapping from the tangent space at $p_0$ in area $U$ to the tangent space at $g(p_0)$ in $U'$.

Using the coordinate system of **Fig. SM8**, the perspective mapping $g$ has only three parameters in the first order Taylor expansion: three parameters representing the *coplanar constraint* are fixed: $\delta_x \equiv 0$, $\frac{\partial x}{\partial x'} \equiv 1$, and $\frac{\partial x'}{\partial y} \equiv 0$.

$$dg = \begin{bmatrix} a_{11} & a_{12} \\ a_{21} & a_{22} \end{bmatrix} = \begin{bmatrix} 1 & 0 \\ \frac{\partial y'}{\partial x} & \frac{\partial y'}{\partial y} \end{bmatrix}.$$

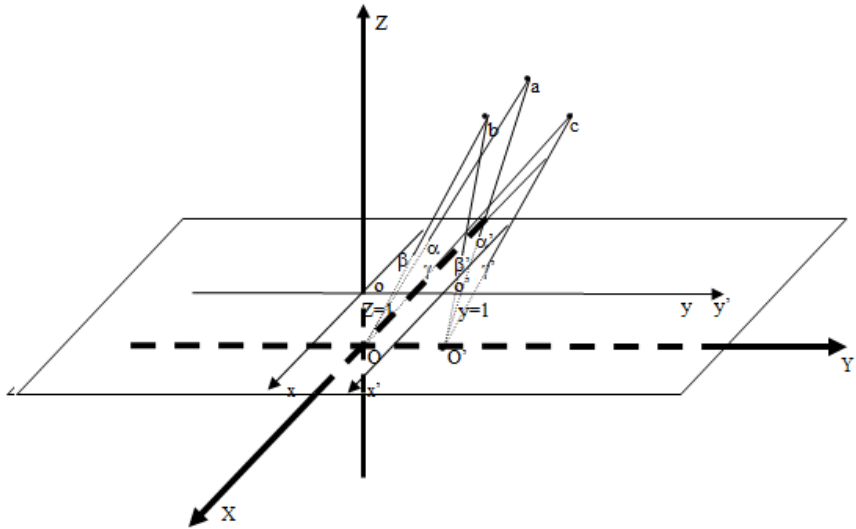

**Figure SM8.** Two cameras located parallel each other with one unit shift along the Y-axis. Both image planes are located on the Z=1 plane. The first center of projection is at the origin of the O-XYZ coordinate system and the second center of projection moves to O' by shifting one unit along the Y axis. The origin of the coordinate system of the second image plane o' is at y=1. Three space points *a, b, c* project to α, β, γ, and α', β', γ' in the first and second camera systems, respectively.

If $\frac{\partial y'}{\partial y} \neq 0$, point $p_0$ is regular and there is a neighborhood $U$ of $p_0$ such that $g: U \to g(U)$ is a diffeomorphism onto the open set $g(U)$ in $S_h(O')$, Namely, there is a perspective mapping with $U$ as its domain. The shape of the rigid surface is specified by the perspective mapping:

**Theorem (Shape from Perspective Mapping)** The 2D manifold in 3D space recovered from a perspective mapping $g$ has a rigid *shape* determined by up to second order derivatives of this mapping.

**Proof:** From the fundamental theorem of surface in differential geometry (11) the rigid shape of a surface is determined by the *first type* and *second type* of *fundamental forms* of the surface. Now we show these fundamental forms are completely determined by the perspective mapping. First we introduce a parameter representation for the surface patch recovered from the perspective mapping $g$. With this parameterization, we then calculate its first type fundamental form and second type fundamental form, both from $g$ and its derivatives. Thus we conclude that with $g$, the Euclidean geometry shape of the 2D manifold is completely specified.

The surfaces recovered from the *triples* can naturally be parameterized by the *ray space coordinates* of the domains of the triples. Using the image plane coordinate system as our ray space coordinate representation, the parameter representation of the surface patch $f$ is now given by

$$f(x,y) = \begin{pmatrix} X(x,y) \\ Y(x,y) \\ Z(x,y) \end{pmatrix} = \begin{pmatrix} xZ(x,y) \\ yZ(x,y) \\ Z(x,y) \end{pmatrix}, \quad (x,y) \in U.$$

With reference to the coordinate system of **Fig. SM8**, we have

$$Z(x,y) = -1/(g_2(x,y) - y)$$

and the surface $f$ can be expressed in terms of $g$ as follows:

$$f(x,y) = \begin{pmatrix} x/(y - g_2(x,y)) \\ y/(y - g_2(x,y)) \\ 1/(y - g_2(x,y)) \end{pmatrix}, \quad (x,y) \in U.$$

From the above expression, it is easy to calculate the *first fundamental form* for the surface $f$: $df_u \bullet df_u$, in terms of $u \in U$ i.e., the parameters $x, y$ taken on the image plane, $g(x, y)$, and $dg(x,y)$, the first derivative of $g(x, y)$.

To compute the second fundamental form, we first look at the normal vector at a point $\begin{pmatrix} X_0 \\ Y_0 \\ Z_0 \end{pmatrix} = \begin{pmatrix} X(x_0, y_0) \\ Y(x_0, y_0) \\ Z(x_0, y_0) \end{pmatrix}$ on the surface. It is along the direction of the cross product of two tangent vectors at that point:

$$((\partial X/\partial x, \ \partial Y/\partial x, \ \partial Z/\partial x) \times (\partial X/\partial y, \ \partial Y/\partial y, \ \partial Z/\partial y))|_{(x_0, y_0)} =$$

$$(Z(-\partial Z/\partial x, \ -\partial Y/\partial y, \ Z + x\partial Z/\partial x + y\partial Z/\partial y))|_{(x_0, y_0)}.$$

From this we have the normalized normal vector:

$$n_0 = (a_{21}/D, \ (a_{22}-1)/D, \ (g_2(x,y) - ya_{22} - xa_{21})/D)|_{(x_0, y_0)}.$$

where $D = \sqrt{a_{21}^2 + (a_{22}-1)^2 + (ya_{22} + xa_{21} - g_2(x,y))^2}$ is the normalization factor, and

$$a_{21} = \left.\frac{\partial g_2(x,y)}{\partial x}\right|_{(x_0, y_0)} \quad \text{and} \quad a_{22} = \left.\frac{\partial g_2(x,y)}{\partial y}\right|_{(x_0, y_0)}.$$

We see that at each point on the surface, the normalized normal vector can be expressed by the perspective mapping $g$ and its first derivative, and it is easy to calculate the *second fundamental form* for the surface $f$: $dn_u \bullet df_u$, in terms of $u \in U$, i.e., the parameters $x, y$ taken on the image plane, $g(x,y)$, $dg(x,y)$, the first derivative and $d^2g(x,y)$, the second derivative of $g(x, y)$. QED.

Thus far, we have shown that the topology and geometric shape of an environmental surface is represented by the *inter-perspective* structure imposed on the visual space by the environment: the perspective mapping structure. And we have shown that despite the variations of this code, the surface is invariant with respect to the perspective mapping structure.

2.4. Persistent surfaces under changing visibility

The visual system perceives environment surfaces invariant to changes of perspective. The term invariant surface in vision has two different meanings: when part of an environment surface is visible from different points of observation, it is seen as the same and fixed in the environment. When different parts of a surface are seen from different points of observation, they are seen as *of the same surface*.

In Section 2.3b, we showed that surfaces are invariant under variations of perspective mappings. A surface is represented by perspective mappings: these perspective mappings form an equivalence relation, and the surface is the invariant with respect to the equivalence relation of perspective mappings.

We formulate the problem of the perception of invariant surfaces under sampling variation in three steps: (1) we show that for each perspective, the visual system derives an *ad hoc* representation of surfaces with visible samples; (2) we show that an equivalence relation can be established on the different visible samples; and (3) we show that from each equivalence class of visible samples an invariant global surface can be derived.

In this section, we first introduce a general criteria for surface connectedness in terms of local surfaces and surface extensions, and define the concept of equivalence classes of visible samples of a surface in terms of surface extensions. We then describe a condition for the equivalence of visible samples obtained along a locomotion path, and demonstrate that the global surfaces are invariants along such a path.

2.4a. Occluding contours separate visible surfaces: One Side Projection Theorem, Local Stability Criteria

of the Regular Value Set Theorem, One Owner Theorem, Separated Surface Continuation Theorem

**Definition (Local Surface, Surface Continuation)** A connected open set of an environment surface is called a *local surface*. Let $\bar{S}$ be the closure of a local surface $S$, and $C(\bar{S}) = \{U_i \mid i \in I\}$ an open cover of $\bar{S}$, where $I$ is a set of indices, $U_i$ connected open sets of the environment surface, each with nonempty intersection with S, and $\bar{S} \subset \bigcup_{i \in I} U_i$. The set $\bigcup_{i \in I} U_i$ is called an *open continuation of the surface S*, or simply a *continuation of surface S*.

The union of two local surfaces is a connected 2D space if they have non-empty overlap ((5), 3.2.5). In general, the union of two overlapping local surfaces is not necessarily a surface: it may form a branched union (**Fig. SM9**). We call those that do result in a surface the surface extension to each local surface.

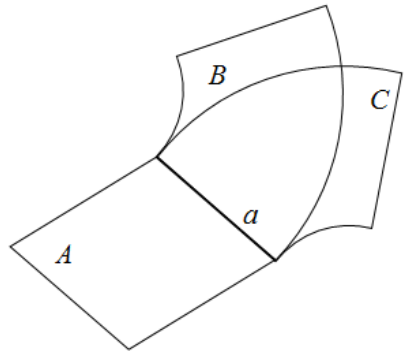

**Figure SM9.** Surface $A$ has two separated extensions $B$ and $C$ from its boundary $a$. Both $A \cup B$ and $A \cup C$ are surfaces. But $A \cup B \cup C$ is not a surface. Surfaces with partial overlap form a connected 2D space but do not necessarily form a surface.

**Definition (Surface Extension, Extension Chain, *E*-chain Surface, *E*-chain Connected Surfaces)** The union set of two local surfaces with nonempty overlap is called a *surface extension* to each of the two local surfaces, if it is a surface. Let $\sigma = \{S_i \mid i = 1, ..., n\}$ be a sequence of local surfaces, it is called an *extension chain*, or *E-chain*, of local surfaces, if for $i = 1, ..., n-1$, $S_i \cap S_{i+1} \neq \varnothing$, and $S_i \cup S_{i+1}$ is a surface. The surface $S = \bigcup_{i=1}^{n} S_i$ is called the *E*-chain surface of $\sigma$, and $S_1$ and $S_n$ are called *E-chain connected* local surfaces, and denoted as $S_1 \sim_c S_n$. It is easy to see that E-chain connectedness is an equivalence relation of the local surfaces of the visual environment: For any local surfaces $\sigma_i$, $\sigma_j$, and $\sigma_k$, (1) $\sigma_i \sim_c \sigma_i$; (2) $(\sigma_i \sim_c \sigma_j) \rightarrow (\sigma_j \sim_c \sigma_i)$; and (3) $(\sigma_i \sim_c \sigma_j) \& (\sigma_j \sim_c \sigma_k) \rightarrow (\sigma_i \sim_c \sigma_k)$.

Obviously, a surface extension of a local surface is a connected surface. By induction, *E*-chain surfaces are connected surfaces. We have the statement:

**Proposition** The *E*-chain surface of an *E*-chain of local surfaces is a connected surface.

Taking the connected surface as the intermediate "local surface" any two simple surfaces in a connected surface are *E*-chain connected.

**Theorem (*E*-chain Connected Criteria)** A surface is a connected surface if and only if any of its two local surfaces are *E*-chain connected.

Proof: We only need to show the "if" part. By definition, if the surface is not connected, then it is the union of two open sets $\sigma_1$ and $\sigma_2$ of the environment surfaces with no intersection. Surfaces are locally connected. Pick from each of the two open sets a local surface. They are not *E*-chain connected. For if they are, then the *E*-chain surface connecting them, $\sigma_e$, is a connected open set. Both $\sigma_1 \cap \sigma_e$ and $\sigma_2 \cap \sigma_e$ are open sets of the visual environment. Their intersection is the null set. Therefore the set $(\sigma_1 \cap \sigma_e) \cup (\sigma_2 \cap \sigma_e) = \sigma_e$ is not a connected set. The contradiction shows if any two local surfaces of a surface are *E*-chain connected, the surface is a connected surface. *QED*.

The concepts of *E*-chain surface and *E*-chain connected surfaces are pertinent to vision for the following reason. At each point of observation, due to self-occlusion, the global surface of an object, even if it is not occluded by other objects, is only partially visible. Therefore from each perspective what the vision system can visibly access are local surfaces. On the other hand, connected surface components are units of ecological significance. They represent spatially coherent objects of the environment. It is through the *E*-chain connected surface, that the vision system is able to derive a surface component from its visible local samples.

The boundary of a 2D domain is called a *contour*. Two local surfaces are called neighboring surfaces if they are non overlapping but their boundaries have non empty intersection. The boundary between two neighboring local surfaces or two neighboring ray space domains is called a *border*.

**Definition (Sandwich Domain, Regular Domain, Side of a Contour)** A *sandwich domain* of a contour *C* in a 2D manifold is a connected open set *S* in the 2D manifold such that $C \subseteq S$, and $S \backslash C = S_1 \cup S_2$, $S_1$ and $S_2$ non-empty non-overlapping domains. *C* is called the border between $S_1$ and $S_2$. $S_1$ and $S_2$ are called *side domains* of *C* within *S*. Given a family of sandwich domains of a contour *C*, side domains of *C* of this family form equivalence classes according to the intersection relation: $S_i \sim S_j$ iff $S_i \cap S_j \neq \emptyset$. The equivalence class of connected side domains is called a *side class of C*. With a given perspective, a domain in the ray space is called a *regular domain* if all its points are *regular values* of the perspective. If both $S_1$ and $S_2$ are *regular domains* of the perspective, *S* is called a *regular sandwich domain of C*. If all members of a side class of a contour are regular domains, the side class is called a *side of the contour*.

**Definition (Spatially Separated Surfaces, Contour of Surface Separation)** Two local surfaces are called *spatially separated* if the intersection of their closures is a null set. A contour in a ray space is called a *contour of surface separation* if it has a regular sandwich domain such that the preimage surfaces of its two side domains are spatially separated.

A surface is a 2D manifold in the 3D Euclidean space. The perspective image of a visible surface is in the 2D ray space. Connectivity is an invariant topological property under homeomorphisms. The set of regular values in a ray space and its preimage, the set of regular points of visible surfaces, are homeomorphic under natural perspective projection. The sets of regular values in the ray space provide a local representation of sets of regular points of environment surfaces. The following theorem shows that the same holds true of surface connectivity.

**Theorem (Pre-image Surface Connectivity)** The preimage of a *connected* open set of regular values of a perspective is a connected open set of an environment surface.

**Proof:** Let *S* be a connected open set of regular values of a perspective. Consider two local surfaces of the preimage surface $f^{-1}(S)$ with images $D_i$ and $D_{i+k}$. We show they are chain connected. Since *S* is connected, it is *E*-chain connected. For any two domains $D_i$ and $D_{i+k}$ in the covering there is a chain of domains $D_i, D_{i+1}, \ldots D_{i+k}$ such that each consecutive pair has nonempty intersection: $D_l \cap D_{l+1} \neq \emptyset$, where $i \leq l < k$. Then for any two domains $f^1(D_i)$ and $f^1(D_{i+k})$ in the covering of the preimage surface $f^{-1}(S) = \bigcup_i f^{-1}(D_i)$, the chain of local surfaces $f^1(D_i), f^1(D_{i+1}), \ldots f^1(D_{i+k})$ has also nonempty intersections: $f^1(D_l) \cap f^1(D_{l+1}) \supseteq f^1(D_l \cap D_j) \neq \emptyset$, where $i \leq l < k$, for each consecutive pair. Therefore $f^{-1}(S)$ is a connected surface. *QED*.

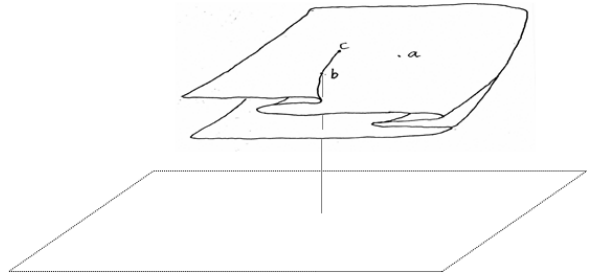

**Figure SM10**. In the projection from a surface to the plane below, point *b* is a fold point, *c* is a cusp point, and *a* is a regular point of this projection.

In a given perspective, the topological division of the set of regular points of the visible surfaces into connected components is represented by the components of regular values in the ray space. The visible surfaces also contain critical points and the image of a natural perspective contains critical values. The next question is: what information of the surface connectivity is carried by the set of critical values?

The structure of singularities of surface projection was described in a theorem proved by Whitney: Every singularity of a smooth mapping of a plane to a plane after an *appropriate small perturbation* splits into *folds* and *cusps* (12, 13). Knowing that surface occlusion is related to the *fold singularity* of perspective projection, the problem of surface extension across the occlusion contours can be reduced to and resolved by the topology of the surface *fold*. **Fig. SM10** shows the critical points in a projection to a plane consist of regular points, folds, and cusps.

The occluding contours are commonly described as an *intensity gradient contour* caused by *depth discontinuity* of nearby visible surfaces (14) and treated as a subset of "real contours," contours of *intensity discontinuities* in images. To define the *occluding contour* with its original meaning of *occlusion* requires distinguishing two sides of the contour. From the pre-image connectedness theorem of regular domains, we can easily deduce the following statement relating the sides of a contour in the ray space to a definite surface in the environment.

**Proposition (Preimage of a Side)** The preimage of a regular side domain of a contour is a local surface. In terms of connectedness, the preimage surfaces of all members of a side of the contour form an equivalence class of surfaces.

It is important to note that in the vicinity of a fold contour, two sides of the fold contour project to one side of the image contour of the fold. Only one side of the surface can be visible. The surface at the other side of the fold is occluded.

**Theorem (One Side Projection)** Every fold contour of an environment surface has a sandwich domain such that the images of its two side domains under point projection to the ray space are located on one side of the image of the fold contour.

**Proof:** Let $p$ be a fold point of the environment surface. In a neighborhood of $p$, there is a parameter representation for the surface:
$$\begin{cases} \rho = \rho(\xi,\eta) \\ \theta = \theta(\xi,\eta) \\ \phi = \phi(\xi,\eta) \end{cases}$$, and the projection to the ray space can be expressed as the mapping $f:(\xi,\eta) \mapsto (\theta(\xi,\eta),\phi(\xi,\eta))$. From the *normal form theorem* ((9), Theorem 4.5), there is a coordinate system $u$ for the $\xi\eta$-plane at the fold point $p$ and a coordinate system $v$ for the $\theta\phi$-plane at the occluding contour point $f(p)$ such that in this coordinate system the projection mapping has the form: $f:(u_1,u_2) \mapsto (u_1, u_2^2)$, namely $(v_1,v_2) = f(u_1,u_2) = (u_1, u_2^2)$. The sandwich domain is divided by the fold contour $u_2 = 0$ into two side domains $\{(u_1,u_2) \mid u_2 > 0\}$ and $\{(u_1,u_2) \mid u_2 < 0\}$. The image of the fold contour in the ray space is the contour $\{(v_1,v_2) \mid v_2 = 0\}$. And through the mapping $f$, which assigns $v_2 = u_2^2$, both the two side domains $\{(u_1,u_2) \mid u_2 > 0\}$ and $\{(u_1,u_2) \mid u_2 < 0\}$ are projected onto one side of the contour $\{(v_1,v_2) \mid v_2 = 0\}$, the domain $\{(v_1,v_2) \mid v_2 > 0\}$. *QED*.

The occluding contour is the image of the surface fold. The fact that the surface at the vicinity of its fold projects to one side of the occluding contour singles out this side as the "owner" of the occluding contour. The occluding contour as the image of a fold has no orientation, but we can assign an occluding contour an orientation so that the owner is on the left side. This way a border has two possible orientations, each representing a different owner side.

**Definition (Occluding Contour, *T*-junction, Owner Side)** In a given perspective, an oriented contour is called an *occluding contour* if it is the image of a fold contour of a visible surface, and the preimage of its left side is a side of the fold of the visible surface. The intersection point of two occluding contours is called a *T-junction*. The left side of the occluding contour is also called the *owner* side.

While the concept of occluding contour thus defined is directly related to the phenomenon of *occlusion* itself, it has a "drawback" compared with intensity edges and figure/ground segregations. These concepts are supported by "proximal" stimuli: the retinal image contains intensity contours and figure qualities such as symmetry and convexity. Our concept of occluding contour explicitly refers to the *environment surfaces*, which are "distal stimuli." How can the visual system directly pick up the information about occluding contours, particularly their *owners*, from the ambient lights reaching the eyes?

The points on occluding contours are critical values of perspective projection. We already know that a regular value is characterized by having some stable neighborhood. Can a critical value be characterized by some sort of unstable behavior under small perturbations of the point of observation? Indeed, as we show next, the theorem of local stability of a regular value set has an inverse: A locally stable set is a set of regular values.

**Theorem (Local Stability Criteria of the Regular Value Set)** A set is locally stable under the perspective mapping structure if and only if it is a set of regular values of natural perspective.

**Proof:** We only need prove the "only if" part to show all points in a locally stable open set are regular values of natural perspective projection. The set is locally stable. Therefore each point in this set has a stable neighborhood under the perspective mapping structure. Each perspective mapping from this stable neighborhood to some other ray space defines a 2D manifold in the respective transition space which corresponds to an invariant environment surface. The stable neighborhood is a set of regular values of perspective projection from the environment surface to the ray space because the *det (Dh) = det(Dg∘Df$^{-1}$)*, where *h* is a perspective mapping of the stable neighborhood between two ray spaces, and *f, g* are perspective projections to two ray spaces. This is true for any point in the locally stable set. The set is a set of regular values. *QED*.

This theorem says no neighborhood of an occluding contour is stable under perspective mapping. Actually, the occluding contours can be more specifically characterized in terms of accretion: Given a sandwich domain of an occluding contour, if a transition mapping maps its two side domains into another ray space, this is always accompanied by an accretion. The extraction of occluding contours from accretion between two perspectives is called da Vinci stereopsis (**Fig. SM11**).

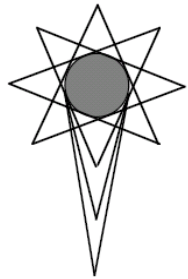

**Figure SM11.** Diagram from Leonardo's notebook illustrating the fact that the light rays leaving an object's surface may be considered to form a collection of cones (which Leonardo calls "pyramids"), each cone constituting an image that would be taken by a pinhole camera at a given location. The diagram shows that different cones cover different parts of an object.

The non-symmetric relation of an occluding contour to its two sides was noticed and studied in Gestalt psychology. Gestalt psychologists call the side of the surface with fold the figure side and noticed its assignment affects our perception of an image. **Fig. SM12** shows the same contour represents two different surface shapes depending on which side owns it.

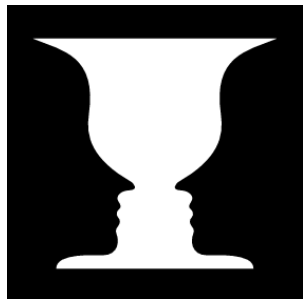

**Figure SM12**: Different assignments of the owner of the occluding contour affect our perception of the shape of the contour as of human face or of a vase.

We have included the *owner* assignment as part of the definition of an occluding contour. With owner assignment, the occluding contours are able to specify the surface connectivity between what is currently visible and what is occluded. As the famous Kanizsa triangle illusion shows, an occluding contour can be perveived without an accompanying intensity gradient contour, and the defining characteristic of an occluding contour is border ownership, not the intensity gradient.

The one-sided projection justifies the "owner" concept for the occluding contour. There would be a problem if the same contour had two owners, two surfaces which each had a fold contour projecting in a perspective to the same occluding contour. This would result in a border dispute! Fortunately, in general this does not happen. To show this we first define a surface created for the first owner. We show generically there would be no second owner.

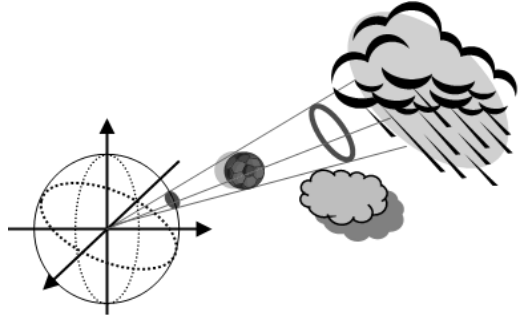

**Figure SM13.** A fold cone surface is the cone of lines passing fold points, which are also the rays on the occluding contour. The upper cloud surface is in a generic position with this cone, while the lower cloud surface is not.

**Definition (Fold Cone Surface)** Given an environment surface with fold in a perspective, the surface formed by lines passing the center of the perspective projection and points on the fold contour is called a *fold cone surface* for the environment surface.

The law governing interrelations between smooth surfaces in 3D Euclidean space and contours on a 2D manifold is *transversality*. In Guillemin and Pollack (15), it is defined:

**Definition (Transversality)** Let $i$ be the inclusion map of a submanifold $X \subset Y$, and $Z \subset Y$ be another submanifold of $Y$. Map $i$ is transversal to $Z$ if and only if for every $x \in X \cap Z$

$$T_x(X) + T_x(Z) = T_x(Y),$$

where $T_x(X)$ is the tangent space of submanifold $X$ at $x$, $T_x(Z)$ is the tangent space of submanifold $Z$ at $x$, and $T_x(Y)$ is the tangent space of the manifold $Y$ at $x$. Notice that this equation is symmetric in $X$ and $Z$. When it holds, we shall say that the two submanifolds are *transversal*.

It can be proven that two surfaces in 3D space are generically transverse, i.e., they are not in general (generically) tangent to another surface if they are not so made, and similarly contours in 2D space are generically transverse, i.e., a contour in a 2D manifold is not in general (generically) tangent to a different contour if it is not so made. Surfaces and curves either keep away from each other or intersect with a non-zero angle.

As shown in **Fig. SM13**, a fold contour is always accompanied by a fold cone surface. Intersections of the fold cone surface with environment surfaces other than the one containing the fold contour are transversal.

**Theorem (One Owner)** An occluding contour in a perspective has one and only one owner.
**Proof:** From the one-side projection theorem, one of two sides of an occluding contour is its owner, and we assign the orientation of the occluding contour to make the owner at the left. Now we show that there is a surface extension from the right side which has no singularity on the preimage of the occluding contour. In general we can assume the occluding contour contains no *T*-junction points. Otherwise, we can separate the occluding contour to several open segments without the *T*-junctions and prove the result for each segment. For an occluding contour containing no *T*-junctions, there always is a regular sandwich domain in the ray space. With the fold contour of the preimage surface of the left side regular domain, we can construct a *fold cone surface* $S_c$. Let $S_r$ be a surface continuation of the preimage surface of the right side regular domain. Two surfaces $S_r$ and $S_c$ intersect transversally. Because all points on the surface $S_c$ project to the occluding contour, the intersection points also project to the occluding contour. We claim these intersection points are *regular points* of $S_r$ for the projection to the ray space. Denote the unit sphere around the perspective center $O$ by $S_o$. The derivative of the projection map $\varphi : S_r \to S_o$ at an intersection point $x$ is the natural linear map between the corresponding tangent planes: $d\varphi_x : T_x S_r \to T_{\varphi(x)} S_o$. This linear map is degenerate only when the normal vector to the plane $T_{\varphi(x)} S_o$, which is along a line of the fold cone passing intersection point $x$, is on the tangent plane $T_x S_r$. Due to the transversality condition of the two surfaces, this cannot happen: The tangent vector of the curve of intersection of $S_r$ and $S_c$ is located in both the tangent planes $T_x S_c$ and $T_x S_r$. The line $\overline{Ox}$ is on the tangent plane $T_x S_c$. If it is also on the tangent plane $T_x S_r$, the two tangent planes are the same. $T_x S_r + T_x S_c = T_x S_r \neq T_x E^3$, the 3D tangent space of the 3D Euclidean space $E^3$ at point x. Therefore $d\varphi_x$ does not degenerate and the point $x$ is not a fold point of the right side preimage surface. Therefore the right side of the contour is not the owner of the occluding contour. QED.

**Theorem (Separated Surface Continuation)** The preimage surfaces of the side domains of the sandwich domain of an occluding contour have separated continuations.

**Proof:** If an open interval of an occluding contour contains *T*-junctions, we can divide the interval into open intervals without *T*-junction. Assuming an occluding contour contains no *T*-junction point, it has a regular sandwich domain. Denote the left and right regular side domains as $D_l$ and $D_r$ and their preimages $S_l$ and $S_r$. The natural perspective to a ray space is a one-to-one mapping from the visible part of environment to the ray space. Therefore, because $D_l$ and $D_r$ do not intersect, the two local surfaces $S_l$ and $S_r$ have no intersection. If $\overline{S_l} \cap \overline{S_r} = \emptyset$, the occluding contour will be a contour of surface separation: since their closures are compact, there is a minimum Euclidean space distance δ between $\overline{S_l}$ and $\overline{S_r}$. An open covering of both $\overline{S_l}$ and $\overline{S_r}$ by open sets of environment surfaces with diameter smaller than δ will give a surface continuation of both $S_l$ and $S_r$ with no intersections. If $\overline{S_l} \cap \overline{S_r} \neq \emptyset$, the projection image of an intersection point in $\overline{S_l} \cap \overline{S_r}$ must be on $\overline{D_l} \cap \overline{D_r}$, the occluding contour. From the one-owner theorem, each intersection point is a regular point of $\overline{S_r}$, but singular point of $\overline{S_l}$. A point of a surface cannot be both a critical point and a regular point. Therefore $\overline{S_l} \cap \overline{S_r} = \emptyset$. QED.

The preimage surface of the owner side has a continuation across the fold that becomes occluded behind the preimage surface of the owner side. The preimage surface of the other side has a continuation across the contour of its intersection with the fold cone surface and also becomes occluded by the preimage surface of the owner side.

2.4b. Ad hoc surface representation and the extraction of occluding contours: Border of Accretion Criteria for Occluding Contours Theorem

The above theorem means that through occluding contours, the vision system can see surfaces extended into the *occluded* area. The vision system can see an occluded surface that has no image on the retina! Gibson called this the most radical discovery of the ecological approach to vision. The perception of an occluded surface near an occluding contour is not inferred (a guess with high probability) based on past experiences. It is directly picked up from the structure of ambient lights. The topological construction of the environment is directly coded in the structure of ambient lights.

Through extraction of occluding contours the vision system makes a topological division of the set of regular values in the ray space such that (1) the components delineated by occluding contours are homeomorphic to their preimage surfaces; (2) the assigned owner side specifies how the preimage surfaces are connected to the preimage of the occluding contour; and (3) the occluding contours further specify how visible surfaces are continuously extended to the occluded surfaces, and thereby connected to the global surfaces. The occluding contours therefore fully specify the topology of a visible surface, including its connection to a global surface which is never fully visible. It is quite clear that within the scope of binocular stereopisis, a main mechanism for surface representation, this computational goal is achieved by the da Vinci stereopisis, not Wheatstone stereopsis.

**Definition** (**Global Surface**) A connected component of the visual environment is called a *global surface*.

The initial surface representation takes the form of "image segmentation" by occluding contours. The segmentation turns a perspective (image) into a countable collection of components, a symbolic representation of the environment surfaces. Being tied to a particular point of observation, it is an ad hoc surface representation.

**Definition** (**Regular Component, ad hoc Surface Representation, Owner of an ad hoc Representation**) In a given perspective, a maximum connected open set of regular values with respect to the partial order of $\subseteq$ is called a *regular component*. A tuple $T = (C,B)$ of a regular component $C$ and its surrounding occluding contours $B$ is called an *ad hoc* representation of a global surface, or an *ad hoc surface representation*. The global surface containing the preimage of the regular component is called the *owner* of the *ad hoc* representation.

**Definition** (**Topological Segmentation of Perspective**) Given a perspective on a ray space S, let each $T_i = (C_i, B_i)$, $i \in I$, be an ad hoc surface representation of the perspective, $I$ a set of indices, the sequence $\{(C_i, B_i) | i \in I\}$ such that $\bigcup_{i \in I} C_i \cup B_i = S$ is called the *topological segmentation* of the perspective.

The topological segmentation of a perspective is a representation of components of visible surfaces. Each component of a visible surface can be further divided into areas, based upon color, texture, and other meaningful environmental invariants.

So far we have deduced properties of occluding contours based upon the definition that they are images of fold singularities. These properties explain why a vision system can "see" occluded parts of a surface--at least see their existence and their connection to the currently visible part. Now we deduce characteristic properties of the occluding contours based upon the perspective mapping structure. These properties explain how the vision system can *extract* the "remotely" determined occluding contours from directly accessible ambient lights.

**Definition (Regular Point with respect to a Mapping Structure, *M*-regular)** A ray in a ray space is called a *regular point with respect to a mapping structure* if it is in the source domain of a nontrivial transition mapping of this mapping structure. The regular points with respect to a mapping structure are called *M-regular points*. Since an *M*-regular point is also an *M*-regular value, and vice versa, we do not distinguish them.

**Definition (*M*-Regular Component and *M*-border)** A maximum element of the set of connected open sets of *M*-regular points in a ray space with the partial order of the subset relation "⊆" is called an *M-regular component*. The boundary of an *M*-regular component is called the *M-border* of the M-component.

We already deduced from Sard's Theorem that almost all points in a ray space are regular points with respect to the perspective mapping structure. This leads to an important property: the boundary of an *M*-regular set is also a border between two *M*-regular sets, i.e., it separates two *M*-regular sets.

Ray spaces are compact 2D manifolds with the metric topology of the unit sphere. When a sequence of inner points in the source domain approaches (Cauchy converges to) a boundary point, the sequence of their image points also approaches a boundary point of the target domain. Therefore a transition mapping always has a continuous extension to the boundary of its source domain.

**Definition: (Boundary Image of Transition Mapping, Side Image of an *M*-Border)** The image of the boundary of the source domain of a transition mapping under its continuous extension is called the *boundary image* of the transition mapping. If a side domain of an *M*-border is the source domain of some transition mapping, the boundary image of the border is called a *side image of the M-border* under the transition mapping.

Now we characterize the *M*-borders in terms of their side images by transition mappings from one ray space to another. *M*-regular components are 2D open sets. Their boundaries are contours.

**Definition (Border of Accretion of a Transition, Border of Infinitesimal Accretion)** Given a mapping structure and an *M*-border in a ray space surrounded by a sandwich domain, if a transition mapping in the mapping structure maps both side domains of the sandwich domain to another ray space, such that the two side images of the *M*-border are *M-regular and have no intersection point*, we call it a *border of accretion of the transition*. The transition is called an *accretion transition* for the border. If for any positive real number ε, there is an accretion transition for the border with distance less than ε, we call the border a *border of infinitesimal accretion*.

**Definition (Border of Accretion of a Mapping Structure)** Given a mapping structure in the visual space, if for every point $p$ on a border $B$ in a ray space there is an open neighborhood segment $S_p$ of $p$, $S_p \subset B$ and $p \in S_p$, that is a border of accretion of some transition, the border $B$ is called a *border of accretion of the mapping structure*. If at every point on a border, there is an open segment neighborhood of the point that is a border of infinitesimal accretion, the border is called a *border of infinitesimal accretion of the mapping structure*.

**Definition (Deletion)** A *P*-transition is said to have a *deletion* towards a border in the target ray space, if its inverse *P*-transition is an accretion transition for the border.

**Definition (Accretion Bordered Mapping Structure)** A mapping structure of the visual space in which every *M*-border is a border of *accretion* of the mapping structure is called an *accretion bordered mapping structure*. It is called an *infinitesimal accretion bordered mapping structure* if every *M*-border is an infinitesimal accretion border.

**Theorem (Border of Accretion Criteria for Occluding Contours)** A contour in a perspective is an occluding contour if and only if it is a border of infinitesimal accretion of the perspective mapping structure. The transition pole of each accretion transition for the occluding contour is at the right side of the occluding contour, namely the opposite side of the owner of the occluding contour.

**Proof:** We first show that an *occluding contour* is locally a *border of infinitesimal accretion*. We do this by first showing that any *point* on an occluding contour has two side images after some transition with arbitrarily small distance which are *M*-regular and different. Since the mappings to side images are continuous along a border of two regular domains, the point has a neighborhood segment on the occluding contour such that every point within the neighborhood also has two different *M*-regular side images in the target ray space. Thus the border is an infinitesimal accretion border.

Let $b$ be a point (ray) on the occluding contour in the ray space centered at $O$, and take a small segment $\sigma$ of the occluding contour containing $b$, and a regular sandwich domain of this contour segment. This always can be done by avoiding the *T*-junctions along the occluding contour. Move the observation point $O$ to a nearby position $O'$ of distance less than some positive number $\varepsilon$. The transition plane determined by points $b, O, O'$ intersects with the environment surfaces with curves $AB$ and $CD$ (**Fig. SM14**) where $A$ and $D$ are on rays of the boundaries of two regular side domains. We assume $C$ is a regular value of projection to the ray space with base point at $O$, although $C$ is occluded by point $B$ (this is true generically). $B$ and $C$ are on the same ray $b$. $B$ is the fold point of the surface. The occluding contours are contours of surface separation, so that $B \neq C$.

The tangent line $\overline{BC}$ of curve $AB$ at point $B$ divides the transition plane into two half planes, right half plane $H_r(b)$ and left half plane $H_l(b)$. The owner side is in the left half plane $H_l(b)$. We now show that when the transition pole $e$ is on the opposite side of the owner in the right half plane $H_r(b)$, a small transition to perspective center $O'$ will map ray $b$ of the ray space $O$ to two side images, ray $O'B$ and ray $O'C$, in the ray space based at $O'$. And both are *M*-regular rays. This will finish the proof that an occluding contour is an infinitesimal accretion border.

We note that near the fold point $B$, the intersection curve of the environment surface with the transition plane has positive curvature (Koenderink & van Doorn 1976 describes local geometry around fold points). Without loss of generality, we assume between points $A$ and $P$ as shown in **Fig. SM14**, the curve has positive curvature. Therefore, we can limit the transition vector to a neighborhood of $O$, such that the tangent lines from the new perspective center to the curve $AP$ has tangent points located within the curve $AP$. This is possible because the location of the tangent point depends continuously on the position of $O'$, and $B$ is inside the curve $AP$.

Both curve $AB$ and curve $CD$ are visible from $O$. Therefore there is no environment surface inside the triangle-like area $OAB$ bounded by line $OA$ and line $OB$ with base curve $AB$, and the triangle-like area $OCD$ bounded by line $OC$ and $OD$ with base curve $CD$. $O'$ is inside the area. Point $B$ is visible from $O'$. Also because $O'$ is not on the line segment $OB$, $C$ is visible.

$B$ is not a fold point in projection to $O'$, for otherwise line $O'B$ would be equal to line $OB$. Using tangent line $t_A$ of the curve $AB$ at $A$ as the base line, because of the positive curvature of the curve $BB'$, line $O'B'$ is tangent to the curve $AP$, its angle with $t_A$ is larger than that of ray $b$. Because the curve has positive curvature, the tangent point $B'$ must be further away from the point $A$ along the curve $AP$ and located in the occluded area of perspective $O$. $B$ is visible from $O'$ and regular. This means ray $O'B$ is a regular value of perspective projection. Because C is a regular value of projection to O, by local stability, it is also a regular value of projection $O'$. Both $B$ and $C$ are regular visible points in the perspective to $O'$. Ray $O'C$ is regular value of perspective projection. Both side images of $b$ are now located in the *M*-regular domain in the perspective centered at $O'$. Thus we proved the occluding contour is an accretion border of the perspective mapping structure. The above proof holds for any such $O'$ in the $\varepsilon$ neighborhood of $O$, where choice of $\varepsilon$ is made to guarantee the tangent lines from $O'$ to the curve $AP$ is inside of the curve $B$ $P$. Therefore an occluding contour is locally a border of infinitesimal accretion.

Now we show a border of accretion in some *perspective transition* is an occluding contour. We first show every point in a border of accretion is a critical value of the perspective projection. Suppose the contrary is true, then there is a point on the border of accretion that is regular value of the perspective projection. The set of regular values are open. There must be a connected neighborhood of this regular value consisting of regular values. Any transition mapping that maps this neighborhood into another ray space is a homeomorphism, and the two side images of the border segment within this regular domain are the same. This shows a border containing a regular value cannot be an accretion border of a *P*-transition. The singularities of a projection from a surface to the ray space are either fold points or cusp points. Cusp points are at the ends of the fold contours. The points of an open segment of a contour of critical values must be the values of the folds. Therefore, the border of accretion is an *occluding contour*.

The transition pole of the start ray space in an accretion transition of the occluding contour is always located on the right side of the occluding contour, namely the opposite side of the owner of the occluding contour. Suppose in a transition, the transition pole of the start ray space is located on the side of the owner. Referring to **Fig. SM14**, $b$ is a point on the occlusion contour in the ray space centered at $O$. Let $\overrightarrow{OO''}$ be a transition vector of a transition in which the transition pole of the start ray space is located on the owner side of $b$. In order for the transition pole, namely the ray from $O$ passing $O''$, to be on the owner side, the left side of $b$, the point $O''$ of the transition vector must be in the left half plane $H_l(b)$. We claim such an transition cannot be an accretion transition of an occlusion contour containing $b$. Suppose this is not true, and a transition with transition vector ending in the left half plane $H_l(b)$ is an accretion transition of the occluding contour. Then there must be a small neighborhood of $B$ on the surface that is regular in the perspective projection to $O''$. Therefore we can pick a point $P$ on the intersection curve of the surface and the transition plane from $O$ to $O''$ containing $b$ such that $P$ is occluded in the perspective to $O$ but visible to $O''$, and all points on the curve $AP$ on the surface are regular in the perspective at $O''$. $P$ is beyond the curve $AB$ because it is not visible from $O$. We show this is impossible.

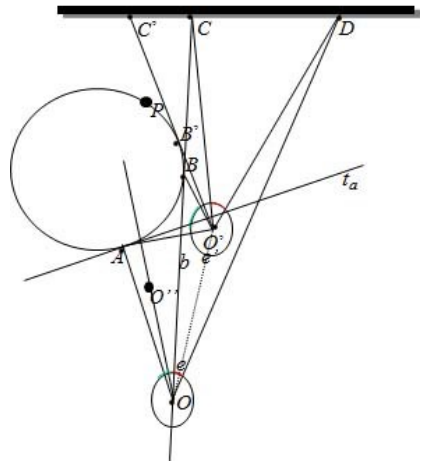

**Figure SM14.** Occluding contours are borders of accretion. $B$ is a fold point for the perspective at $O$, but is regular for the perspective at $O'$.

All points on the curve $AP$ on the surface are regular in the perspective at $O''$. Now the tangent line of curve $AP$ passing $O''$ has a smaller angle with $t_a$ than that of $b$, and the curve $AB$ has positive curvature, thus the tangent line from $O''$ must be tangent to the curve $AB$ at some point between $A$ and $B$. Points $B$ and $P$ are behind this tangent point and are occluded in the perspective at $O''$. The contradiction shows $O''$ cannot be the center of an accretion transition for $b$. Therefore for any accretion transition of an occluding contour, the transition pole of the start ray space must be located on the side opposite the owner. *QED*.

## 2.4c. Specifying a persistent global surface: Global Surface from the Union Surface of a Complete Sample Theorem, Fixed Owner Theorem

Now we demonstrate the global surfaces of the visual environment can be specified by continuations of visible surfaces through accretion transitions. This means the distal global surfaces can be perceptually accessed through sequences of local surfaces coded in the form of perspective mapping triples.

**Definition (Set of Regular Visible Surfaces)**: Let $\tau(P, Q)$ be the *P*-transition from a point *P* to another point *Q*, $\Sigma(\tau(P,Q))$ the set of *perspective surfaces* recovered from the correspondent domains of $\tau(P,Q)$. We call $\Sigma(\Omega) = \bigcup_{(P,Q)\in\Omega\times\Omega} \Sigma(\tau(P,Q))$ the set of *perspective surfaces* of the observation domain $\Omega$. When $\Omega$ is the observation space, $P\in\Omega$, $\Sigma(P, \Omega) = \bigcup_{Q\in\Omega} \Sigma(\tau(P,Q))$ is called the set of *regular visible surfaces* at *P*. It consists of open sets of regular points on the environment surfaces visible from the perspective center *P*. In addition to these points, the fold points are also visible.

During locomotion or head/eye movement, the sampled local surface of a global surface is changing. Now we show that from the changing local surfaces, the vision system can see the invariant global surfaces of the environment.

**Definition (the Complete Sample of a Surface)** The equivalence class $C(\sigma_0) = \{\sigma \mid \sigma \sim_c \sigma_0, \sigma\in\Sigma(\Omega)\}$ is called a *complete sample of surface* $\sigma_0$ in $\Sigma(\Omega)$, or simply a complete sample in $\Sigma(\Omega)$. Each element of a complete sample is called a *sample* surface.

Given a perspective transition $\tau(P,Q)$, the set $\Sigma(P,Q) = \Sigma(P)\cap\Sigma(Q)$ $(= \Sigma(\tau(P,Q)))$ contains the surfaces visible from both *P* and *Q*. We call $D\Sigma(\tau(P,Q)) = \Sigma(P)\backslash\Sigma(\tau(P,Q))$ the set of *out-of-sight surfaces* of $\tau(P,Q)$, and $A\Sigma(\tau(P,Q)) = \Sigma(Q)\backslash\Sigma(\tau(P,Q))$ the set of *into-sight surfaces* of $\tau(P,Q)$. Obviously, from $\tau^1(P,Q) = \tau(Q,P)$, and $\Sigma(\tau(P,Q)) = \Sigma(\tau(Q,P))$, we have $A\Sigma(\tau^1(P,Q)) = D\Sigma(\tau(P,Q))$, and $D\Sigma(\tau^1(P,Q)) = A\Sigma(\tau(P,Q))$.

**Proposition** Let $\sigma_0$ be a perspective surface, there is only one complete sample of the surface of $\sigma_0$ in $\Sigma(\Omega)$.

**Proof:** Suppose $C'(\sigma_0)$ is another complete sample of $\sigma_0$, any element $\sigma\in C'(\sigma_0)$ implies $\sigma \sim_c \sigma_0$, hence $\sigma\in C(\sigma_0)$. Therefore $C'(\sigma_0) \subseteq C(\sigma_0)$. Similarly, we have $C(\sigma_0) \subseteq C'(\sigma_0)$. Therefore $C'(\sigma_0) = C(\sigma_0)$. *QED*.

Thus the *complete sample of a surface* defines a partition of $\Sigma(\Omega)$, i.e., is an equivalence class.

**Theorem (Global Surface from the Union Surface of a Complete Sample)** Let $\Omega$ be the observation space. The surface of the union of samples in a complete sample in $\Sigma(\Omega)$ is a global surface.

**Proof:** A global surface is a compact connected component of a 2D manifold embedded in the 3D Euclidean space. The proof is in two steps.  We first show that the maximum element is a global surface.  Then we show that the union of complete samples is a maximum element.  Therfore the union of complete samples must be a global surface. Given a surface $\sigma_0$, let $\sigma_m$ be its maximum continuation, we claim $\sigma_m$ is a global surface. Certainly $\sigma_m$ is a an open, connected surface because it is a union of an open, connected cover of $\sigma_0$. It is also closed. For if it is not then its closure must be different from it, and further continuation can be made. This contradicts the fact that it is a maximum element with respect to surface continuation. The maximum element of such a set is unique. For if there are two such maximum elements, they have non-empty intersection σ₀, so their union will be a continuation of σ₀ containing both as subsets. This contradicts the fact that they are maximum elements.

Now we show that the union surface from a complete sample is also a maximum element with respect to the partial order of surface continuation. The union surface from a complete sample must be an open subset of the maximum element. We show it is also closed. If it is not closed, its closure must contain a point that is on the maximum element but not in the union of the members of the complete sample. Let *p* be a point of the closure of a perspective syrface σ of the sample that is not in the union of members of the complete sample. We now prove such a point *p* cannot exist. Point *p* must be a boundary point of σ, for otherwise it is already a point of the union of the complete sample. If *p* is a regular point, its image has an *M*-regular domain *s*. The preimage of *s* is a neighborhood of p. The union of the preimage of *s* with σ is a new sample because σ and *s* have non-empty intersection. This contradicts the assumption that the complete surface includes all samples within the equivalence class. If *p* a singular point, its image under perspective projection must be on an occluding contour. An occluding contour is a border of infinitesimal accretion. The transition that makes the side images of the image of point *p* into *M*-regular domains will reduce this situation to the previous one: either side image of such a *P*-transition will turn the singular preimage point into an *M*-regular point. Therefore, such a point *p* cannot exist. This shows the union of the complete sample of a surface is a global surface. *QED*.

**Definition (Perspective Inspection Path and Perspective Inspection Groupoid)** A perspective inspection path is defined recursively. A perspective transition is a perspective inspection path with two ordered footprints: the start point and the end point. Two perspective inspection paths are consecutive if the end point of one is the start point of the other. In this case, we call the first one the prior and the second one the posterior. The composition of two consecutive perspective inspection paths is a path with footprints the concatenation of the footprints of the two consecutive perspective inspection paths such that the start point of the prior is the start point and the end point of the posterior is the end point. The groupoid formed from perspective inspection paths is called the perspective inspection groupoid, or *PI*-groupoid. An element of the *PI*-groupoid is called a *PI*-path.

**Definition (Persistent *ad hoc* representations through *PI*-paths, *PI*-path equivalent *ad hoc* Representations)** If the preimage surface of a regular component *Δ* of a perspective is not in the out-of-sight part of a *P*-transition, there is a correspondent visible component in the target perspective of the *P*-transition such that their pre-image surfaces overlap. The two pre-image surfaces belong to the same complete sample. We call the first pre-image surface persistent in the *P*-transition and the second pre-image surface a persistence of the first one. In this case, there is a mapping triple in the perspective mapping structure that maps a subset of *Δ* to a subset of a regular component *Δ'* in the second perspective. The *ad hoc* surface representation of *Δ'* and its borders in the second perspective is called a persistence of the *ad hoc* representation of *Δ* and its borders in the first perspective. Let $\{\tau(P_i, P_{i+1})| P_1, P_{i+1} \in \Omega, i = 1, ..., n\}$ be a *PI*-path starting from position $P_1$ and ending at position $P_n$. Let $\sigma(\Delta_i) = (\Delta_i, B_i)$ be the *ad hoc* representation of a surface, where $\Delta_i \in S(P_i)$ a regular component, $B_i$ its border, $i = 1, ..., n, n+1$. If $\sigma(\Delta_{i+1})$ is a persistence of $\sigma(\Delta_i)$, $i = 1, ..., n$, we call the *ad hoc* representations of surface $\sigma(\Delta_i)$, $i = 1, ..., n$, persistent through the *PI*-path. Persistence through *PI*-paths is an equivalence relation of *ad hoc* representations of a surface. Two *ad hoc* rperesentations are called *PI-path equivalent* if there is a *PI*-path such that one is a persistence of the other through the *PI*-path.

**Theorem (Fixed Owner)** The owner of the *ad hoc* representations persistent through a *PI*-path is fixed.

Proof: The preimage surface of the regular component of the persistence of an *ad hoc* representation has non-empty overlap with the preimage surface of the regular component of its prior *ad hoc* representation. They belong to one equivalence class with respect to the relation $\sim_c$. The global surface of the union of the members of the complete sample is fixed. *QED*.

Persistence of the *ad hoc* representations of a surface through *P*-transitions maintains *invariance* of the owner under variations of the samples. Gibson made a careful distinction between the perception of *persistent surfaces* and the *persistent perception* of surfaces. Persistent perception of a surface implies some kind of memory mechanism to link different samples. Perception of a persistent surface is the extraction of the fixed owner from its persistent *ad hoc* representation through various *PI*-paths.

## 3. Extracting transition vectors from perspective transitions

In perspective transition, vision system does not see any change of the environment. Instead, it generates a sense of its own movement. *Visual kinesthesis* is "a fact of psychology" that is ignored in the egocentric depth map formulation of distance perception. There is a critical difference between perceiving an object moving toward you, and perceiving that you are moving toward an object. The latter is what is actually perceived during ego motion through an invariant 3D environment. Information about the movement of the eye in space underlies all significant visual processes for extraction of invariant surfaces. Perspectives are structured through *P*-transitions in a way isomorphic to eye movements. It is from the structured perspectives, the groupoid of perspective transitions, that the invariant environment surfaces can be extracted. Without the groupoid structure, a sequence of perspectives is only a pile of highly redundant 2D images, from which no invariant surface can be extracted. The transition vector is what makes two perspectives a *P*-transition.

How does the vision system extract information about the movement of the eye when there is a *P*-transition? Poincarè thought motion information is derived from the motor control signal accompanying the body, head, and eye movements. Gibson pointed out a visual channel of ego motion information must exist for the simple reason that a person sitting in a vehicle still has ego motion information.

The mathematical relation between transition vector and the perspective transition can be demonstrated with a pinhole camera model and an image plane coordinate system associated with the camera (**Fig. SM7**). In such a system, camera movement includes a translation from one point in space to another as well as rotation around the point of projection.

Suppose a surface point *P*: (*X, Y, Z*) is projected to two correspondent points *p, p'* on the image plane in two views and measurements are made: *p*: (*x, y*) and *p'*: (*x' y'*). Let $R(\alpha,\beta,\gamma)$ be the rotation operator in terms of the three angles of rotation (yaw, pitch, and roll) around the *z*-axis, *y*-axis and *x*-axis, and ($t_x$, $t_y$, $t_z$) the translation of the camera. The *coplanar constraint* for the correspondent points on the image plane in the two views can be written as:

$$F_f(\alpha,\beta,\gamma,t_x,t_y,t_z) = ((x,y,f) \times (t_x,t_y,t_z)) \bullet ((R(\alpha,\beta,\gamma) \circ (x',y',f)) = 0$$

where *f* is the focal length of the camera.

The operation of $(t_x,t_y,t_z) \times \bar{v} = \bar{w}$ provided by the vector $(t_x,t_y,t_z)$ using the cross product is a linear transformation on the 3D vector space and can be expressed by a 3 by 3 matrix:

$$T(t_x,t_y,t_z) = \begin{bmatrix} 0 & -t_z & t_y \\ t_z & 0 & -t_x \\ -t_y & t_x & 0 \end{bmatrix}.$$

And the coplanarity condition can be expressed as:

$$F_f(\alpha,\beta,\gamma,t_x,t_y,t_z) = (T(t_x,t_y,t_z) \circ (x,y,f)) \bullet (R(\theta,\varphi,\psi) \circ (x',y',f)) = 0.$$

The coplanarity constraint is a system of homogeneous equations and consequently its solution carries an undetermined *scale* in motion vector. This can also be seen from the "motion-and-structure from image motion" system of equations:

$$x_i' = \frac{(r_{11}(\rho_i x_i - t_x) + r_{12}(\rho_i y_i - t_y) + r_{13}(\rho_i f - t_z))}{(r_{31}(\rho_i x_i - t_x) + r_{32}(\rho_i y_i - t_y) + r_{33}(\rho_i f - t_z))}, \text{ and}$$

$$y_i' = \frac{(r_{21}(\rho_i x_i - t_x) + r_{22}(\rho_i y_i - t_y) + r_{23}(\rho_i f - t_z))}{(r_{31}(\rho_i x_i - t_x) + r_{32}(\rho_i y_i - t_y) + r_{33}(\rho_i f - t_z))},$$

where $\rho_i$ is the distance of the *i*-th point $P_i$ from the origin $O$, representing the 3D geometric structure of the rigid environment, and $r_{kl} = r_{kl}(\alpha,\beta,\gamma)$ is the *k*-th row *l*-th column element of the rotation matrix $R = R(\alpha,\beta,\gamma)$. The system of equations will not change if $(t_x,t_y,t_z)$ is multiplied by a scale factor $s$, provided that all $\rho_i$ in this system of equations are also multiplied by the same scale factor.

From a *P*-transition, there is an equivalence class of 3D representations of the transition vector and the vectors from the point of observation to the points of the environment surface points. The ratio between the scale of transition vector and the scale of distances between points in the rigid environment is invarfiant, not the *absolute scale* of the transition vector or the distance between points of the rigid environment, define a *rigid structure* of the environment. That is to say, the invariant "length" of the eye movement is the ratio of measurements of the magnitude of th transition vector and the distance of a fixed pair of ground points, or some rigid body of the environment, such as a standard "meter" or a ruler, or any two fixed points on the ground,etc.

By taking the transition vector as the unit for measuring distances between points on rigid bodies of the environment, the transition vector can be represented in a polar coordinate system with two angles: $(t_x,t_y,t_z) = (\sin(\theta)\cos(\phi), \sin(\theta)\sin(\phi), \cos(\theta))$, and the matrix

$$T(\theta,\phi) = \begin{bmatrix} 0 & \cos(\theta) & \sin(\theta)\sin(\phi) \\ -\cos(\theta) & 0 & -\sin(\theta)\cos(\phi) \\ -\sin(\theta)\sin(\phi) & \sin(\theta)\cos(\phi) & 0 \end{bmatrix}$$

has two variables. The coplanar constraint equation has five parameters to be determined:

$$F_f(\alpha,\beta,\gamma,\theta,\phi) = (T(\theta,\phi) \circ (x,y,f)) \bullet (R(\theta,\varphi,\psi) \circ (x',y',f)) = 0.$$

There needs five correspondent points between two consecutive frames to fully constrain the *orientation* of the transition vector and the 3D rotation of the camera.

Although the *scale* of movement is not an *environment invariant*, the *direction* is. If *T* is part of the solution of the coplanarity equation, -*T* also is. However, only one is true. There needs a check on whether two *coplanar rays* intersect to determine the *direction* of movement.

In above formulation, the rigid motion of a pinhole camera is specified by five parameters: two parameters for the direction of camera translation, and three parameters for camera rotation around its focal point. The equation is non-linear. The computer and robotics vision studies of egomotion are mostly based upon two popular concepts: essential matrix (16) and fundamental matrix (17). In the essential matrix the five rigid motion parameters are mixed into eight substantial parameters. By imposing a minimization problem within a 5D manifold as a minimization within a 8D manifold, the essential matrix method effectively gave up the most essential geometric constraint of rigid motion, seems unlikely to be able to achieve the very goal of computing rigid motion.

From a *P*-transition, there is an equivalence class of 3D representations of the transition vector and the vectors from the point of observation to the points of the environment surface points. What is *invariant* is the ratio between the scale of transition vector and the scale of distances between points in the rigid environment, not the *absolute scale* of the transition vector or the distance between points of the rigid environment, define a *rigid structure* of the environment.

### 4. Extracting perspective mappings from perspective transitions

Having shown that in vision surface information is represented by *mapping triples* of the visual space (section 2.3), we now address the problem of how to pick up these *triples* from the structure of ambient lights, in particular the perspective transitions. A detailed description of these ideas is given in (18, 19).

The images are the samples taken from optical arrays. They form a subset of the Hilbert space of real-valued functions defined on 2D domains. Our goal is to pick up the information of *perspective mappings* in the form of the *transformations* between stereo pairs of images. By using the term "information" we mean (1) the *qualitative* information about *whether there* is a space-like coplanar transformation between image patches of different perspective; and (2) *numerical descriptions for* such transformations.

Our computational approach to *perspective transformation* is through *Taylor approximation*. The derivatives are a sufficient tool to judge if there is a diffeomorphism between small domains around a pair of correspondent points. Guillemin and Pollack made the following comment on the role of the determinant of the derivative of a mapping: "a truly remarkable and valuable fact. …… tells us that the seemingly quite subtle question of whether *f* maps a neighborhood of *x* diffeomorphically onto a neighborhood of *y* reduces to a trivial matter of checking if a single number – the determinant of $df_x$ – is nonzero!"

A perspective mapping is a mapping between two ray spaces by way of coincidences on environment surfaces. An exact numeric description of a diffeomorphism by a Taylor expansion requires infinitely many real numbers. However, we can make the following compromises: (1) working on only small domains of neighborhoods of a set of grid points, and (2) tolerating certain levels of error. What we gain from this compromise are: (1) by using the first order Taylor expansion we need *only three numbers* to provide a local description of the perspective mapping; (2) the approximation provided by the three numbers covers a truly 2D domain nearby the grid point. Therefore although the number of the grid points may be large, it is a *finite set* in each bounded 2D area. In practical situations, the approximations based upon these grid points may be sufficiently accurate.

An image can be regarded as a linear summation of Hilbert space components. Even from a small image patch, we can pick up plenty of different Hilbert space components. By selecting a set of *linear independent* *basis elements*, we can obtain a set of *coefficients* of these basis elements through linear space projections. The perspective transformations act on images as linear operators on a Hilbert space, acting on each component of an image in the same way as the image. The role that perspective transformations play to change images into different images is different but related to the original role of perspective transformations for mapping an open set to another open set as originally defined. The transformations as linear operators on the Hilbert space are *representations* of the original perspective transformations. In vision, we extract the perspective transformations from their representations as linear operators on images (for discussion of the representation of transformation groups, see (20)).

Suppose $A(S(O))$ is an ambient optic array defined on the ray space $S(O)$, and $U \subset S(O)$ is a domain in the ray space. The part of the optic array on $U$ is denoted by $A_o(U)$. Similarly, we have $A_{o'}(U')$, $U' \subset S(O')$. The action of the transformation $g$ can be expressed by the equation

$$A_o(U) = A_{o'}(g(U)). \tag{1}$$

Sampled into images from two locations $O$ and $O'$, image matching between "image patches" $f_o(U)$ and $f_{o'}(U')$ gives:

$$f_o(U) = f_{o'}(g(U)). \tag{2}$$

This gives the transformation $g$ the proper representation in the Hilbert space of images.

The neurons in the primary visual cortex of monkeys have receptive fields. The response signal to an image patch $f(U)$ can be treated as the inner product of the receptive field functions (*rf*-functions) $F_i(U)$, $i = 1, 2, ..., n$ of the neuron on the patch with $f(U)$:

$$\gamma^i(U) = <F_i(U), f(U)>, \ i = 1, \ldots, n.$$

Equation (2) implies

$$<F_i(U), f_o(U)> = <F_i(U), f_{o'}(g(U))>, \ i = 1, \ldots, n. \tag{3}$$

The system of equations (3) is a "weak" form of relation (2) as it checks only $n$ "response" signals, while the single equation (2), the *true equation* for images, represents infinitely many of these component equations. Our original formulation of picking up transformation from *image pairs* now turns into the formulation of picking up transformation from some *Hilbert space components* of the sampled image pairs. The change raises two questions: (1) Will this equation be mathematically solvable; and (2) will the solution of this "weak" form equation still yield the transformation we are looking for?

With reference to **Figs. SM7 and SM8**, when the image plane makes a parallel shift along the *y*-axis, at each point the first order Taylor expansion of the coplanar diffeomorphism $g$ is an affine transformation with three parameters. At point $p_0$, we use $A(\frac{\partial y'}{\partial x}\big|_{p_0}, \frac{\partial y'}{\partial y}\big|_{p_0}, \delta_y(p_0))$ as an approximate representation of $g$ and measure the *degree of matching* within the scope $U$. Using $a_{21} = \frac{\partial y'}{\partial x}\big|_{p_0}$, $a_{22} = \frac{\partial y'}{\partial y}\big|_{p_0}$ to simplify notation, and letting $t = \delta_y(p_0)$ represent the "disparity" of center $p_0$ when $O$ moves to $O'$, we have the following system of equations:

$$< F_i(U), f_o(U) >\ =\ < F_i(U), f_{o'}(A(a_{21}, a_{22}, t) \circ U) >,\ i = 1, \ldots, n.$$

The equations for the three variables $a_{21}, a_{22}$, and $t$ form a nonlinear system of equations. We can use a gradient method to solve for the three unknowns. However, to compute the gradients requires taking *Lie derivatives* on the components against the parameters of transformation. Because the images are assemblages of measurements of the intensity of incident lights impinging on the sensor units, they have no analytical form that we can compute derivatives from.

From the Hilbert space representation concept, we can view the affine transformation on $U$ as a linear transformation on the function, denoted by the same symbol, so that we have:

$$< F_i(U), f_o(U) >\ =\ < F_i(U), A(a_{21}, a_{22}, t) \circ f_{o'}(U) >,\ i = 1, \ldots, n.$$

Let $A^*(a_{21}, a_{22}, t)$ be the conjugate transformation of $A(a_{21}, a_{22}, t)$ with respect to the inner product. Finally we have derived the system of equations that we can solve:

$$< F_i(U), f_o(U) >\ =\ < A^*(a_{21}, a_{22}, t) \circ F_i(U), f_{o'}(U) >,\ i = 1, \ldots, n. \qquad (4)$$

The process for picking up the transformation can now be formulated as minimizing the differences between the components of the image and its transformation. The objective function for minimization, the energy $E$ with three undetermined parameters $t$, $a_{21}$, and $a_{22}$, is the following:

$$E(a_{21}, a_{22}; t, U, o, o') = \sum_{i=1}^{n} \left\| < F_i(U), f_o(U) > - < A^*(a_{21}, a_{22}, t) \circ F_i(U), f_{o'}(U) > \right\|^2.$$

Equation (4) has important operational consequences. We are seeking a mechanism for the visual system to pick up the 2D to 2D transformations, the *numeric description* of the perspective mapping. The input from ambient optic arrays is in the form of a pair of similar images. The transformation itself is implicit in the data. It is up to the vision system, with its processing mechanisms, to pick this up. In terms of inner products and conjugate transformations, and with a set of *analytical rf-functions*, a gradient dynamical system (21) for determining $dg(p_0)$ in terms of $a_{21}$ and $a_{22}$, and the shift $t$ is fully specified, in which the Lie differentiations against the parameters $t$, $a_{21}$, and $a_{22}$ are represented by a type of *linear operator* derived from *rf-functions* known as *Lie germs*.

In Equation (3), the perspective transformation represents a physical fact resulting from different perspective projections. In Equation (4), however, the transformation becomes the "duty" of receptive fields of neurons in the vision system to make the Hilbert space components match! Nature has made a mapping resulting in a changed image when the eye moves from one point of observation to another, it is up to the brain to respond to this mapping with its own transformation. Equation (4) means the *responsibility is* given to the receptive fields of neurons. By so responding, the perspective mapping is picked up by the neurons, in the form of the *conjugate* transformation. It is conceivable that since nature supplies transformations that can be approximated with only three real parameters, the dynamical receptive fields of the neuron may also controlled by three real parameters.

Does the weak form of equation (1) discard useful information and lead to a distorted version of the true perspective mapping? Lights emitted or reflected from the same source or surface point may not have the exact same optical characteristics when seen at different points of observation. Furthermore, they must pass the medium and lens to be sensed. Thus noise, disturbances, and geometric distortion are inevitable. The visual system model, like any meaningful physical system model, must assume certain stability criteria: percepts (system prediction) should be stable under perturbations on measurements (22).

In equation (3), many aspects of the ambient lights $f_o(U)$ and $f_o(g(U))$ must be "ignored" and others emphasized by properly choosing $F_i(U)$ ($i= 1, 2, …, n$), in terms of frequency, scale, and orientation, such as band pass 2D Gabor functions, so that both *high frequency random noise* and DC components (constant intensity shifts) will not affect the measurement results. Instead, the measurements will specifically respond to the stable characteristics of the ambient light, such as orientation specific contrasts associated with textures of the surfaces.

Picking up *perspective transformation* is equivalent to picking up the *surface patches*. Detecting *the accretion border of the perspective transformation* is equivalent to picking up the *occluding contours*. The computation of perspective transformation is the very *core* of *surface perception*. A detailed description of the algorithm can also be found in (19).

## 5. Summary

Unlike taste and touch, the function of vision is to allow an animal to know the environment without immediate body contact. Psychologists call objects distant from the body of an animal the "distal stimulus," and those immediately contacting the sensory system the "proximal stimulus." In vision, the link between the distal and proximal stimulus is given by lights reflected from environment surfaces. Gibson called the laws governing the light rays reflected from environment surfaces the "ecological optics."

In this article, we first introduced two topological spaces: one for describing the 3D objects in the environment, and one for describing the light rays reflected from the environment objects. The first space is the familiar 3D Euclidean space. The second space is the set of ray spaces assigned to each observation point in the medium, and the ambient optic array is the particular pattern of light rays residing in this space.

The concept of ambient optic array is the key to understanding how visual surface perception works. Eyes are often modeled as pinhole cameras. The ambient optic array converging to the potentional pinholes constitutes the physical reality within which the vision system evolved.

With the proper spatial concepts, we show how the 3D objects and their surfaces, contours, domain-connected components, etc., are able to impose structures of rays in the visual space through perspective projections and create the two most important structures in ecological optics, the *perspective mapping structure*, and the *occluding contours* at each perspective. The first has the form of a *pseudogroup*, and the second is a very peculiar mathematical structure, the critical values of perspective projections in each ray space. With these concepts we developed a new type of local surface representation, *ad hoc surface representation*, which encodes all the contiguous surfaces and their overlap relationships at a single perspective, and a new type of global surface representation, *persistent surface representation*, which encodes the globally invariant surfaces in the environment through overlaps between different ad hoc surface surface representations.

A major puzzle in vision research is how the objects and their surfaces are perceived as invariant in outer space, from changing perspective images (e.g., see (23)). We reformulate this problem as one of *how an equivalence class* of topological objects in the visual space represents a unique object in the 3D Euclidean space. The persistent surface representation is specified by the equivalence relation of *chain connectedness*.

The mathematical theory presented in Sections 2 and 3 concerns only the air and the surfaces in the environment. It has nothing to do with the vision system. It describes the topologic and geometric structure present in the animal's optical environment. Because we are dealing with optics, everything can be addressed in a pure and exact sense. Now, if the light carries the information about the topologic and geometric structure, and the eye is able to see this information in the lights, this makes vision possible. Here approximation and ambiguity may occur. In Section 4, we describe one possible algorithm by which the visual system may extract perspective mappings computationally in order to perceive invariant surfaces.

The concept of topological information processing as a major component of vision has far reaching consequences for our understanding of how biological vision functions and how computational vision algorithms should be designed.